\documentclass[12pt,a4paper]{article}
\usepackage[utf8]{inputenc}
\usepackage{LobsterTwo}
\usepackage[T1]{fontenc}
\usepackage{setspace}
\usepackage[
backend=biber,
style=numeric-comp,
citestyle = chem-acs,
sorting=ynt
]{biblatex}
\usepackage{mathptmx}
\usepackage{amsmath}
\usepackage{amsfonts}
\usepackage{amssymb}
\usepackage{amsthm}
\theoremstyle{definition}
\newtheorem{definition}{Definition}[section]
\usepackage[inline]{enumitem}
\usepackage{array}
\usepackage{tabularx}
\usepackage[table]{xcolor}
\usepackage{xy}
\usepackage{soul}
\usepackage{ragged2e}
\usepackage{times}
\usepackage{csquotes}
\usepackage{graphics}
\usepackage{graphicx}
\usepackage{caption, subcaption}
\usepackage{booktabs}
\graphicspath{{./images/}}
\usepackage{copyrightbox}
\usepackage{float}
\usepackage{epsfig}
\usepackage{threeparttable}
\usepackage{multirow}
\usepackage{multicol}
\usepackage{epstopdf}
\usepackage{url}
\usepackage{color}
\usepackage[ruled,vlined]{algorithm2e}%
\usepackage{pslatex}
\usepackage[colorlinks]{hyperref}
\hypersetup{
	colorlinks=true,
	linkcolor=blue,
	filecolor=magenta,      
	urlcolor=cyan,
	pdftitle={Aditya MS-Project-report},
	linktocpage = false,
	citecolor=green,
}
\usepackage[left=1.in, right=1.00in, top=1.00in, bottom=1.00in]{geometry}
\usepackage{fancyhdr}
\usepackage{datetime}
\usepackage{titlesec}
\titleformat{\paragraph}
{\normalfont\normalsize\bfseries}{\theparagraph}{1em}{}
\titlespacing*{\paragraph}
{0pt}{3.25ex plus 1ex minus .2ex}{1.5ex plus .2ex}
\usepackage{tocloft}

\allowdisplaybreaks
\begin{document}
	\thispagestyle{empty}
	\begin{center}
		\hrule
		\vspace*{0.65cm}
		\begin{center}
			\textbf{\Huge Multi-Label Classification on\\
					\vspace*{0.5cm}Remote-Sensing Images}
		\end{center}
		\vspace*{0.6cm}
		\hrule
		\vspace*{0.6cm}
		{\large A project report submitted in partial fulfillment of the requirements for the degree of}
		\\
		\vspace*{0.6cm}
		\textbf{\Large Master of Science (MS)}
		\\
		\vspace*{0.6cm}
		by
		\\
		\vspace*{0.6cm}
		\textcolor[rgb]{0.00,0.00,0.63}{\textbf{\Large Aditya Kumar Singh}}
		\\
		\vspace*{0.45cm}
		{\Large Roll No: 16MS023}
		\\
		\vspace*{0.6cm}
		to the
		\\
		\vspace*{0.6cm}
		{\uppercase{\large Department of Mathematics and Statistics}}
	\end{center}
	
	\begin{figure}[h]
		\centering
		\includegraphics[width=5.2cm, height=5.2cm]{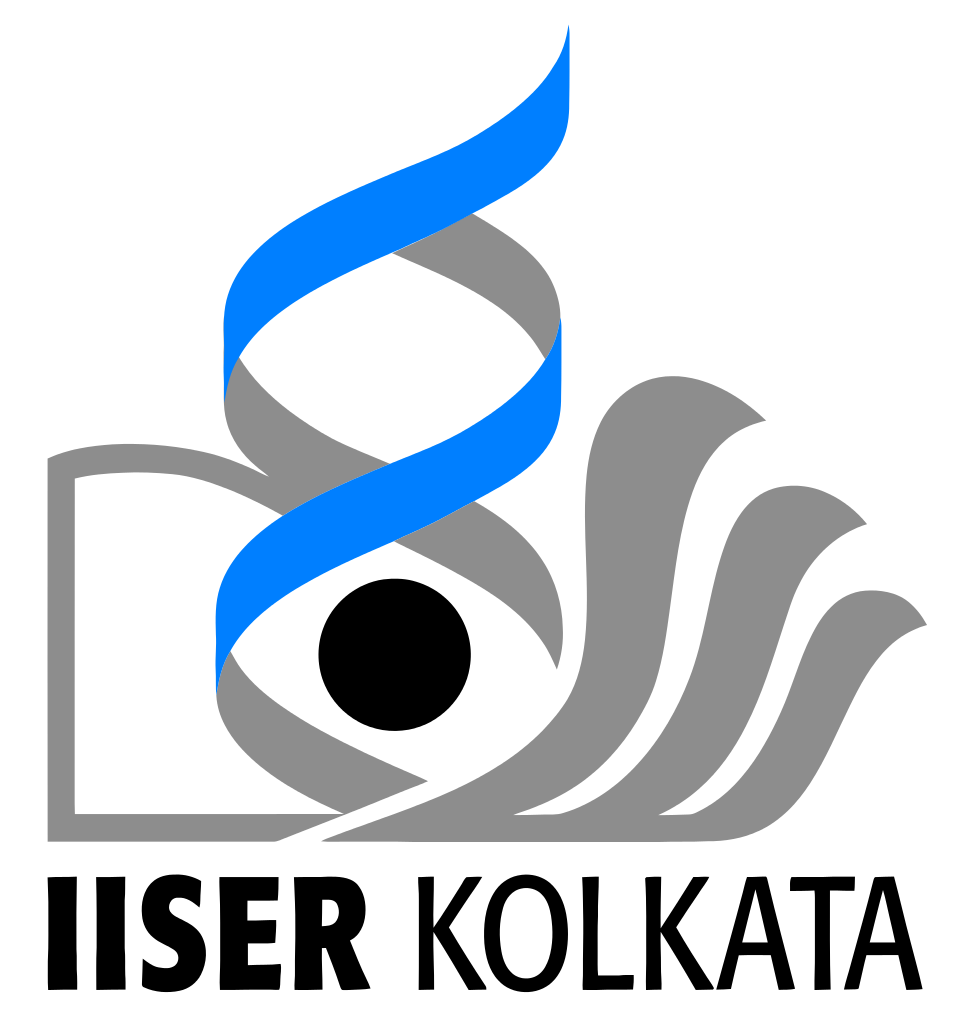}
	\end{figure}
	
	\begin{center}
		{\uppercase{\large Indian Institute of Science Education and Research, Kolkata}}
		\\
		\vspace*{0.65cm}
		Supervised by:
		\\
		\vspace*{0.45cm}
		\textcolor[rgb]{0.00,0.00,0.63}{\textbf{\Large Dr. B. Uma Shankar}}
		\\
		\vspace*{0.35cm}
		{\large Associate Professor - Machine Intelligence Unit - ISI Kolkata}
		\\
		\vspace*{0.45cm}
		{\Large July 2021}
	\end{center}
	\pagenumbering{roman}
	\newpage
	\thispagestyle{plain}
	\leftline{\textbf{\Large Declaration of Authorship}}
	\vspace*{1.25cm}
	\rightline{\today}
	\vspace*{0.35cm}
	\begin{doublespace}
		I, Mr. Aditya Kumar Singh, Registration No. $16 \mathrm{MS} 023$, a student of the Department of Mathematics and Statistics of the 5 years integrated BS-MS Dual Degree Programme of IISER Kolkata, hereby declare that the project report titled \enquote{\textbf{Multi-Label Classification on Remote-Sensing Images}} is my own work and, to the best of my knowledge, it neither contains materials written by any other person, nor it has been submitted for any degree/diploma or any other academic award anywhere before, except where due references have been made.
	\end{doublespace}
	\vspace*{2cm}
	\begin{flushleft}
		\rule{0.4\textwidth}{0.55pt}\\
		\begin{doublespace}
			Aditya Kumar Singh\\
			Department of Mathematics and Statistics\\
			Indian Institute of Science Education and Research, Kolkata\\
			Mohanpur, 741246, West Bengal, India.
		\end{doublespace}
	\end{flushleft}
	\newpage
	\thispagestyle{plain}
	\leftline{\textbf{\Large Certificate from the Supervisor and the Co-Supervisor}}
	\vspace*{1.25cm}
	\rightline{\today}
	\vspace*{0.35cm}
	\begin{doublespace}
		This is to certify that the project report entitled \enquote{\textbf{Multi-Label Classification on Remote-Sensing Images}} submitted by Mr. Aditya Kumar Singh, Registration No. $ 16\mathrm{MS}023 $, a student of Department of Mathematics and Statistics of the 5-year integrated BS-MS Dual Degree Programme of IISER Kolkata, is based upon his own work under my supervision. This is also to certify that either the project report or any part of it has been submitted for any degree/diploma or any other academic award anywhere before, except where due references have been made. In my opinion, the project report fulfills the requirements for the award of the degree of Master of Science.
	\end{doublespace}
	\vspace*{2cm}
	\noindent
	\begin{minipage}[t]{0.42\textwidth}
		\raggedright
		\rule{\textwidth}{0.55pt}\\
		\begin{doublespace}
			Dr. B. Uma Shankar\\
			Associate Professor\\
			Machine Intelligence Unit\\
			Indian Statistical Institute, Kolkata\\
			B.T. Road, Dunlop, 700108, West Bengal, India.
		\end{doublespace}
	\end{minipage}
	\hfill
	\begin{minipage}[t]{0.42\textwidth}
		\raggedright
		\rule{\textwidth}{0.55pt}\\
		\begin{doublespace}
			Dr. Satyaki Mazumder\\
			Assistant Professor\\
			Department of Mathematics and Statistics\\
			Indian Institute of Science Education and Research, Kolkata\\
			Mohanpur, Nadia, 741246, West Bengal, India.
		\end{doublespace}
	\end{minipage}
	\newpage
	\thispagestyle{plain}
	\begin{center}
		\textbf{\Large Acknowledgements}
	\end{center}

	\begin{doublespace}
		{I would like to express my sincere gratitude to my supervisor Dr. B. Uma Shankar for his guidance, advice, suggestions, and encouragement throughout the work. His incredible insights and advice have helped me to overcome the hurdles of the project report. He has always guided me to be an excellent fellow in ML/DL domain and, overall, a good human being, for which I am forever indebted to him.
			
		I am grateful to the Machine Intelligence Unit of ISI Kolkata for providing me the required resources and an exceptional lab and the Department of Mathematics and Statistics, IISER Kolkata, for allowing me to carry out this project at ISI Kolkata. I am also thankful to the Department of Science and Technology, Government of India, and IISER Kolkata for providing financial support for this work through the INSPIRE Fellowship. I am highly thankful to all my professors at IISER Kolkata, especially the statistics unit and the newly opened computer science department, for their invaluable courses and top-notch professors. And to the IISER Kolkata family as a whole, where I spent the best five years of my life.
			
		I am immensely grateful to Dr. Kashinath Jena (mathematics teacher) at my intermediate school for encouraging me to pursue mathematics as my specialization. I want to express my deep gratitude towards Dr. Varun Dutt, who first introduced me to the notion of research in ML/DL, and for the opportunity to be a part of his team during my summer internship at IIT Mandi. Plus, the invaluable exposure I got throughout the training is an exceptional addition to my overall credibility, which greatly enhanced my interests in ML/DL. I also wish to thank Sudeep Da (Ph.D. Student at ISI Kolkata), with whom I had enlightening discussions on multi-label classification, and Subhash Da for rescuing me out in any technical crisis I got trapped.
			
		I am deeply thankful to my friends at IISER Kolkata for their support and encouragement over these five years. Thanks to my roommates and first friends, Ramyak (the Leonardo-da-Vinci of IISER Kolkata), Maya (the sleeping beauty), and Dehury (my brother in crime) (who weren't my roommates technically, but I used to stay in their room more than my own) for all the memories and late-night gossips. Thanks to our one and only Debashis Parida (the great thinker of Odisha and our culture's savior) and Devesh (the game planner and chef) for treating me as their younger brother. Thanks to Panda and Sanapala (the pop-culture guy), my first roommates at IISER Kolkata, for teaching and preparing me a night before semester exams and recommending superb south Indian movies and songs. A huge thanks to Rakesh and Jiban Bhai for their outstanding group training and maintaining the same passion and enthusiasm within the team towards badminton. I want to thank our girl's badminton team members and my friends Nilu and Balu for their dedication and true sportsmanship spirit. Lastly, a great shoutout to my batchmates, namely, Siddhikant (the all-rounder chocolate guy), Shekhar (the substitute player), Dipjyoti (the one-liner guy), Sandip (the topologist), Anirban (the randomly walking dark horse prodigy), Santanu (my lifesaver), Ashsis (the pubg guy), Narayan, and Pallab (the high-society guys), and many others to create a warm and friendly atmosphere.
			
		I would also like to extend my heartfelt thanks to the open-source support forums like Stack Overflow, Stack Exchange. Plus, the blog and code sources like Machinelearningmastery, Towards Data Science, Medium, Github, Paperswithcode, Kaggle, Analytics Vidhya, and open-source frameworks like TensorFlow, Keras, and Python community for developing and providing enormous resources, which became critical for this work.
			
		Last but not least, I would like to express my profound gratitude towards my parents and my brother Sandeep, who have always supported me and allowed me to make my own decisions all these years, which helped me realize my potential. They have been a constant source of inspiration for me both in my life's academic and personal aspects. This would not have been possible without you.}
	\end{doublespace}
	\newpage
	\thispagestyle{plain}
	\vspace*{\fill}
	\begin{center}
		{\Large \copyright\ Copyright by\\
		\vspace*{0.25cm}
		Aditya Kumar Singh\\
		\vspace*{0.25cm}
		July 2021\\
		\vspace*{0.25cm}
		All Rights Reserved.}
	\end{center}
	\vfill
	\newpage
	\thispagestyle{plain}
	\vspace*{\fill}
	\begin{center}
		\LobsterTwo
		\begin{doublespace}
			\textsl{\large To my loving parents, Dileswar and Dipali Singh}\\
			\textsl{\large for their endless love, support, and tenacious words of encouragement.}
		\end{doublespace}
	\end{center}
	\vfill
	\newpage
	\thispagestyle{plain}
	\vspace*{\fill}
	\begin{center}
		\begin{minipage}[c]{\textwidth}
			\begin{doublespace}
				{\Large\textsl{\enquote{A single neuron in the brain is an incredibly complex machine that even today we don’t understand. A single \enquote*{neuron} in a neural network is an incredibly simple mathematical function that captures a minuscule fraction of the complexity of a biological neuron.}}}
			\end{doublespace}
			\vspace*{0.2cm}
			\begin{flushright}
				\begin{minipage}[c]{0.5\textwidth}
					\rm --- Andrew Yan-Tak Ng, Co-founder of Google Brain and Coursera, Director of Stanford AI Lab
				\end{minipage}
			\end{flushright}
			
		\end{minipage}
	\end{center}
	\vfill
	\newpage
	\tableofcontents
	\newpage
	
	\listoftables
	\addcontentsline{toc}{section}{\protect\numberline{}\listtablename}
	
	\newpage
	
	
	
	\listoffigures
	\addcontentsline{toc}{section}{\protect\numberline{}\listfigurename}
	\pagebreak
	\rhead{Abstract}
	\vspace*{1cm}
	\section*{\centering {Abstract}}
	\addcontentsline{toc}{section}{\protect\numberline{}Abstract}
	\vspace*{0.5cm}
	
	Acquiring information on large areas on the earth's surface through satellite cameras allows us to see much more than we can see while standing on the ground. This assists us in detecting and monitoring the physical characteristics of an area like land-use patterns, atmospheric conditions, forest cover, and many unlisted aspects. The obtained images not only keep track of continuous natural phenomena but are also crucial in tackling the global challenge of severe deforestation. Among which Amazon basin accounts for the largest share every year. Proper data analysis would help limit detrimental effects on the ecosystem and biodiversity with a sustainable healthy atmosphere. This report aims to label the satellite image chips of the Amazon rainforest with atmospheric and various classes of land cover/ land use through different machine learning and superior deep learning models. Evaluation is done based on the F2 metric, while for loss function, we have both sigmoid cross-entropy as well as softmax cross-entropy. Images are fed indirectly to the machine learning classifiers after only features are extracted using pre-trained ImageNet architectures. Whereas for deep learning models, ensembles of fine-tuned ImageNet pre-trained models are used via transfer learning. Our best score achieved so far with F$ _{2} $ metric $ = 0.927 $.
	\newline
	\vspace{0.5cm}
	\\
	\emph{Keywords:} Fine tuning; Ensemble learning; Transfer learning; Image classification; Stratified Cross-validation; ImageNet Models.
	
	\newpage
	\pagenumbering{arabic}
	\thispagestyle{empty}
	\rhead{Introduction}
	\section[Chapter 1]{\Large Chapter 1}
	{\Huge \textbf{Introduction}}
	\vspace*{0.8cm}
	\hrule
	\subsection[Overview]{\Large Overview}
	
	This fast-changing world has put a heavy toll on mother nature. For the sake of humankind, we have become human-centric and paving the path for the destruction of our habitat in the name of \enquote{development.} Ranging from deforestation and global warming to outbreaks of the pandemic are the consequences of our actions or the better way to put this up; this is the fruit we reap for the seeds we have sown. According to National Geographic \cite{wiki_amazon}, the most threatening fact is that more than 20\% of the Amazon rainforest is already gone and is currently vanishing at a rate of 20,000 square miles a year. Amazon rainforest, being the largest rainforest in the world \cite{wiki_amazon} with an area span of 6,000,000 sq. km., represents over half of the planet's rainforest and comprises the largest and most biodiverse tract of tropical rainforest in the world. Plus, it produces 20\% of oxygen found on earth, making it called \enquote{the lungs of the world.} According to Dr. Pacala, if the current trend continues, our planet will soon face four major concerns: climate, food, water, and biodiversity \cite{wilson_amazon}.\\
	
	Given this dearth of reliable and precise estimates, a comprehensive understanding of natural and anthropogenic changes in Amazon is still lacking. And for that Government and Scientists need a way to monitor these actions. One way to tackle this challenge is to regularly monitor from above through satellite and drones, capture high-quality images/videos, and analyze them to understand practical solutions. Rudimentary methods for detecting and quantifying deforestation, especially in the Amazon, have proven to be inadequate. Firstly, the existing models cannot often differentiate between human-caused and natural forest loss. Secondly, the coarse-resolution imagery taken up through these models doesn't allow the detection of small-scale deforestation and local forest degradation. For example, selective logging, which involves logging only selected tree species, can conceal significant logging for low-resolution images.\\
	
	Recent improvements in satellite imaging technology and widespread satellite deployment have allowed for more accurate quantification of broad and minute changes on earth at frequent intervals. These methods effectively address the problem with earlier methods, but still robust methods for performing this have not yet been developed. One technique could be to design automated classification of these images based on the atmospheric conditions and land use, which will enable us to respond quickly and effectively to deforestation and other human-made harmful encroachments.\\
	
	Planet Labs Inc, a satellite imaging company, released a dataset of more than 1,00,000 images from the Amazon basin on the Kaggle platform in 2017 under the challenge - \textit{Planet: Understanding the Amazon from Space} \cite{kaggle-planet}. The dataset composed of both labeled (about 40,000) and unlabeled images (the rest) of 3-5 meter resolution where the challenge is to design an algorithm(s) to label satellite image chips w.r.t. atmospheric conditions and various classes of land cover or land use. The problem is a \underline{multi-label} classification problem with a total of 17 labels: 4 labels for the weather conditions and the rest for land cover/ land use patterns. Each image has one of four atmospheric labels and zero or more of 13 ground labels. By definition, cloudy images have zero ground labels, as none should be visible. Some ground features are anthropogenic (habitation, slash burn, mining), while others are natural (water, blooming, blowdown). More than 90\% of images are labeled \enquote{primary,} implying forest cover in them. Six of the other ground labels appear in <1\% of images but are often the ones we are particularly interested in identifying.
	
	\subsection[What did we do?]{\Large What did we do?}
	
	Before approaching our approach, let's first state the definition \cite{multi-label-book} of multi-label classification.
	\begin{definition}
		Let $\mathcal{X}$ denote the input space, with data samples $X \in A_{1} \times A_{2} \times$ $\ldots \times A_{f}$, being $f$ the number of input attributes and $A_{1}, A_{2}, \ldots, A_{f}$ arbitrary sets. Therefore, each instance $X$ will be obtained as the cartesian product of these sets.
	\end{definition}
	\begin{definition}
		Let $\mathcal{L}$ be the set of all possible labels. $\mathcal{P}(\mathcal{L})$ denotes the powerset of $\mathcal{L}$, containing all the possible combinations of labels $l \in \mathcal{L}$ including the empty set and $\mathcal{L}$ itself. $k=\left|\mathcal{L}\right|$ is the total number of labels in $\mathcal{L}$.
	\end{definition}
	\begin{definition}
		Let $\mathcal{Y}$ be the output space, with all the possible vectors $Y$, $ Y \in \mathcal{P}(\mathcal{L})$. The length of $Y$ always will be $k$. Basically, $ \mathcal{Y} = $ collection of all vectors in $\{0,1\}^{k}$ whose length will be always $k$.
	\end{definition}
	\begin{definition}
		Let $\mathcal{D}$ denote a multilabel dataset, containing a finite subset of $A_{1} \times$ $A_{2} \times \ldots \times A_{f} \times \mathcal{P}(\mathcal{L})$. Each element $(X, Y) \in \mathcal{D} \mid X \in A_{1} \times A_{2} \times \ldots \times A_{f}, Y \in \mathcal{P}(\mathcal{L})$ will be an instance or data sample. $n=\left|\mathcal{D}\right|$ will be the number of elements in $\mathcal{D}$.
	\end{definition}
	\begin{definition}
		Let $\mathcal{F}$ be a multilabel classifier, defined as $\mathcal{F}: \mathcal{X} \rightarrow \mathcal{Y}$. The input to $\mathcal{F}$ will be any instance $X \in \mathcal{X}$, and the output will be the prediction $Z \in \mathcal{Y}$. Therefore, the prediction of the vector of labels associated with any instance can be obtained as $Z=\mathcal{F}(X)$.
	\end{definition}
	We approach the problem starting with a naive idea of implementing 17 different single-label binary classifiers that predict the presence of a particular atmosphere or ground feature. These classifiers are generally machine-learning models (XGBoost \cite{xgboost}, Random Forest \cite{random-forest}, Gradient Boosting Classifier \cite{gbc-1, gbc-2}, Extra Trees Classifier \cite{etc}, and LDA \cite{lda}) to which preprocessed images in the form of flattened vectors are fed to train them against the given labels. Further, we deployed multi-label CNN frameworks for image classification, starting with a baseline simple deep CNN architecture, which performs at par with the machine learning models used before. However, unlike the baseline model trained from scratch, we then build our multi-label classifier (an ANN) on top of pre-trained CNN architectures that have performed well in single-label ImageNet Large-Scale Visual Recognition Challenge (ILSVRC) \cite{imagenet}. Namely, we experiment with all ResNet-V2 models \cite{resnet-v2}, VGG16 and VGG19 \cite{vgg}, Xception \cite{xception}, all DenseNet models \cite{densenet}, InceptionV3 \cite{inception-v3}, and Inception-ResNetV2 \cite{inception-resnet-v2}. Comparison among all of them is made based on the F$ _{2} $ score \cite{f-beta} over the validation set, which is a function of the number of false negatives (FN) and false positives (FP), with a larger emphasis on FN. Our model takes satellite images as input and produces a dynamically sized set of tags as output based on the cutoff threshold for each of 17 features that decide which labels to keep and which not.\\

	The rest of the report details the satellite imagery data and its various aspects, the strategies we adopted to deal with them, the pre-processing stage, all in Chapter \ref{chap2}, different ML and DL architectures, and the efficacy of individual experiments in maximizing the F2 score in Chapter \ref{chap3} and \ref{chap4}. Meanwhile, the last chapter, \ref{conclusion}, briefs all the different approaches that we shall perform in the future and the conclusion.
	
	\subsection[Related Work - Reviewing previous literature]{\Large Related Work - Reviewing previous literature}
	
	Previous research on multi-label classification (MLC) and remote-sensing image analysis (namely satellite image classification) motivated us to further build our model off and some general experimentation ideas from the benchmark baseline framework. MLC has been studied in various domains, such as document classification, bioinformatics, and the labeling of multi-media resources. To tackle this problem of MLC, BP-MLL \cite{28-1}, a neural network algorithm, used a novel pairwise ranking loss function to exploit the correlation among labels. But later, cross-entropy \cite{18-1} was found to be good as the pairwise ranking loss function, and the neural network that uses this as the loss function can encode dependencies among labels properly. With this intuition and seeing the performance of several successful models, such as \cite{14-1} that used cross-entropy loss for multi-label image annotation, we even decided to go with this.\\
	
	Coming to satellite image classification, it has been a task of interest ever since the 1970s, when the first multi-spectral remote sensing imagery got available. Earlier, the overall approach has remained conceptually the same -- obtain satellite observations, extract a feature vector for each image, run a classification algorithm, and produce classification labels \cite{1-2, 2-2}. Till 2000, these feature vectors were either manually encoded by humans or were constructed using some unsupervised algorithms like image filtering, PCA, or clustering. Thereafter some simple regression-based algorithm \cite{1-2, 5-2, 4-2} train upon these features to produce multiple labels. Unfortunately, these models were not that powerful or particularly illuminating. According to a study of multi-label satellite image classification algorithms, \enquote{there has been no demonstrable improvement in classification performances over 15 years (1989 -- 2003)\cite{1-2}.}\\
	
	Later, in the 2000s, image processing communities of remote sensing groups began collaborating with artificial intelligence and computer vision researchers to advance classification techniques \cite{4-2, 2-2, 8-2, 3-2}. Hybrid methods that used a combination of manual tagging and traditional ML techniques became more prevalent \cite{2-1}. Several works have explored ML approaches for the problem of land use classification, such as used Decision Trees \cite{20-1}, considered Random Forest \cite{9-1}, and experimented with Support Vector Machines (SVM) \cite{12-1} with Radial Basis Function (RBF) and polynomial kernels.\\
	
	As of now, with increased computing power, more neural network-based frameworks have become feasible for use in large-scale remote sensing image analysis. Researchers are increasingly using these approaches with appreciably improved results. These state-of-the-art techniques use a multitude of fully connected and convolutional layers with non-linear activations to project the image data onto a predefined feature space used to train models to predict probabilities of given labels \cite{13-2, 8-2, 9-2, 14-2}. One no longer needs to annotate the images manually to prepare the training data. This approach allows models to directly engineer an image's feature vector from raw pixel values based on the specific label we want to predict \cite{1-2, 15-2}.\\
	
	Recently, Castelluccio et al. \cite{11-3} used pre-trained CNN, CaffeNet \cite{12-3}, and GoogleNet \cite{inception-v1} (now called Inception-V3) to classify remote sensing images for land use policy. To reduce the computational cost required for training and to reduce design time, they used the pre-trained weights (i.e., they transfer the weights of the model to their customized model -- popularly called transfer learning \cite{transfer_learning}) and built a small refined network on top. This technique is what one can witness in this paper, where the binary sigmoid cross-entropy function is embedded in the last layer to decide given label applies or not based on a certain threshold. Many research has played while including these thresholds as model parameters \cite{10-2, 7-2}, with considerable success.\\
	
	Several other models have sought to address this challenge by applying statistical techniques, NLP, and other approaches \cite{8-2, 11-2, 12-2, 13-2}. For example, \textit{DeepSAT} uses several CNN architectures to predict four or six different land cover classes \cite{9-2}. Next, multi-instance multi-label approaches use multi-layer perceptrons to extract regional features, then pass to a second stage that is meant to capture connections between labels and regions as well as correlations between labels \cite{11-2, 2-2}. Finally, another unique approach was proposed by Wang et al. \cite{8-2} that uses the CNN-RNN method to extract image features and then predict a sequence of labels (RNN) that captures dependencies among labels while maintaining label order invariance.\\
	
	Our methodology described elaborately in Chapter \ref{chap3} seeks to evaluate and synthesize some novel successful techniques within the scope of this MLC problem. First, we will experiment with ML models for which ResNet-50 will act as feature descriptors. Next, we will see some deep CNN frameworks from a unique angle and verify their performance against the first.
	
	\newpage
	\thispagestyle{empty}
	\rhead{Dataset}
	\section[Chapter 2]{\Large Chapter 2}\label{chap2}
	{\Huge \textbf{Dataset}}
	\vspace*{0.8cm}
	\hrule
	\vspace*{0.6cm}
	The Kaggle dataset provided for the challenge was collected over one year, starting in 2016. They come in two formats, 3-channel (RGB) JPEG and 4-channel (RGB + Near IR) GeoTIFF (Geo-referenced TIFF), both with a resolution of $ 256\times 256 $ sampled from a large $ 6600\times 2200 $ pixel \enquote{Planetoscope Scene} \cite{kaggle-planet}.
	\begin{figure}[h!]
		\centering
		\copyrightbox[b]{\includegraphics[width=\linewidth]{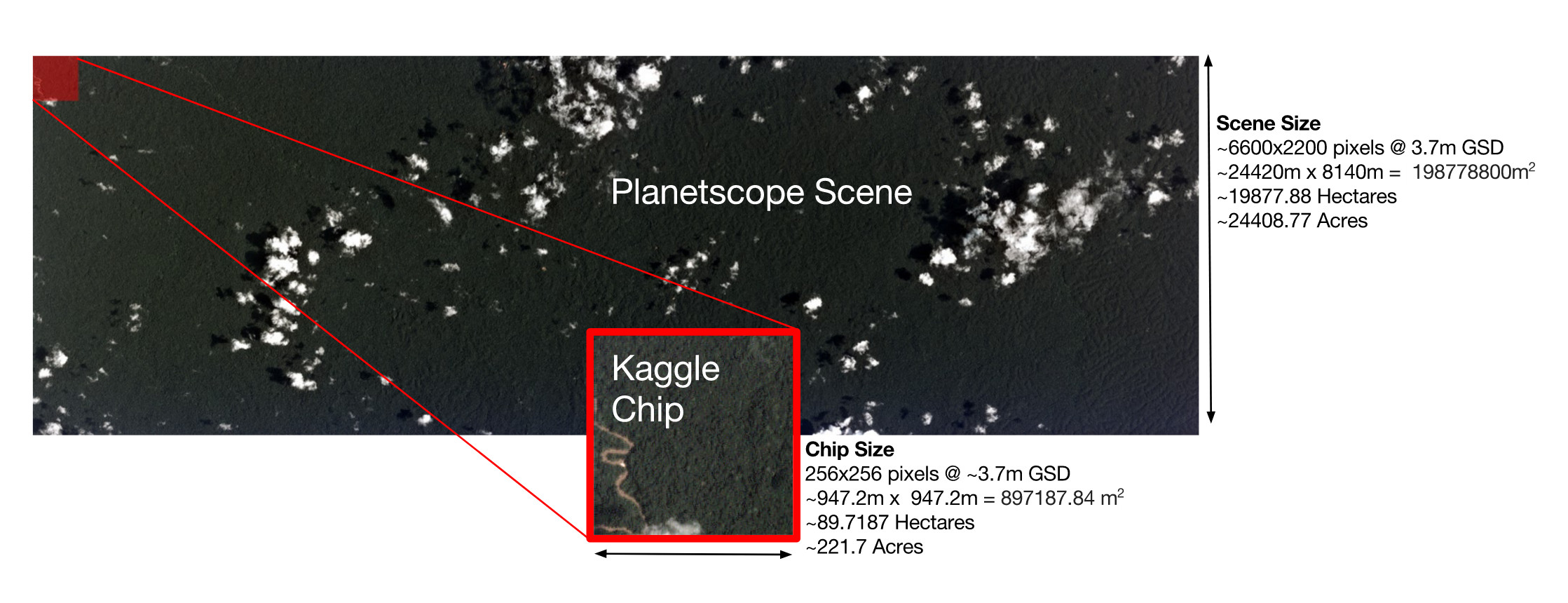}}
		{Source: \href{https://storage.googleapis.com/kaggle-competitions/kaggle/6322/media/chipdesc.jpg}{Kaggle Challenge - Planet: Understanding the Amazon from Space}}
		\caption{Planetoscope Scene}
		\label{fig:1}
	\end{figure}
	An 8-bit integer represents each pixel in each band of JPEG, while for GeoTIFF, it is represented by a 16-bit integer. The near-infrared band is helpful while analyzing atmospheric conditions, such as clouds. Since the GeoTIFF images have calibration issues and mislabeling, as per Kaggle's discussion, we restrict ourselves to JPEG format image chips. The data has a ground sample distance (GSD) of $ 3.7m $ and an orthorectified pixel size of $ 3m $.\\
	
	Overall, there are a total of $ 1,01,670 $ images, out of which $ 40,479 $ images have the associated ground truth labels and are used for training our models, whereas the rest don't have any tags, are of no use to us. The actual labels of these $ 61,191 $ test images were with Kaggle for evaluation of the predictions made by the participants. We performed Stratified 5-fold cross-validation (See, \ref{strcv}) (i.e., splitting the dataset into equally sized folds while maintaining the count of classes in each fold approximately the same) for a generalized evaluation. In total, there are 17 classes, and each image can have multiple classes. These 17 tags can broadly be broken into three groups: atmospheric conditions, common land cover or land use phenomena (representing natural phenomena), and rare land cover or land use phenomena (mainly due to human-made actions). Let's unfold these broad categories and have a look at all possible labels.
	\begin{enumerate}[leftmargin=*]
		\item Atmospheric label:
		\begin{enumerate}[leftmargin=*]
			\item Clear - for transparent weather
			\item Cloudy - Only clouds are visible; the ground scene is obscured.
			\item Partly cloudy - Patch of scattered clouds; ground partially visible.
			\item Haze - Misty or foggy weather; ground visibility is translucent.
		\end{enumerate}
		\item Common land cover/ land use:
		\begin{enumerate}[leftmargin=*]
			\item Primary - For primary tropical forest cover.
			\item Agriculture - Mainly farmlands.
			\item Cultivation - Shifting cultivation in rural areas.
			\item Habitation - Indicates clusters of artificial homes; includes dense urban centers to rural villages along the banks of rivers.
			\item Water - Represents water bodies like rivers and lakes.
			\item Roads - For facilitating transportation.
			\item Bare-ground - Naturally occurring tree-free areas.
		\end{enumerate}
		\item Rare land cover/ land use:
		\begin{enumerate}[leftmargin=*]
			\item Slash-and-burn - A subset of shifting cultivation labels.
			\item Selective logging - A practice of selectively removing high-value tree species from the rainforest (teak and mahogany), which appears as winding dirt roads from space.
			\item Blooming - A natural phenomenon where a particular species of flowering tree blooms at the same time.
			\item Conventional mining - Large-scale legal mining operations.
			\item Artisinal mining - Small-scale mining operations.
			\item Blowdown - Toppling of giant rainforest trees which make open areas visible from space occurred due to windthrow during downburst.
		\end{enumerate}
	\end{enumerate}
	\begin{figure}[H]
		\centering
		\copyrightbox[b]{\includegraphics[width=0.82\linewidth, height=3.8cm]{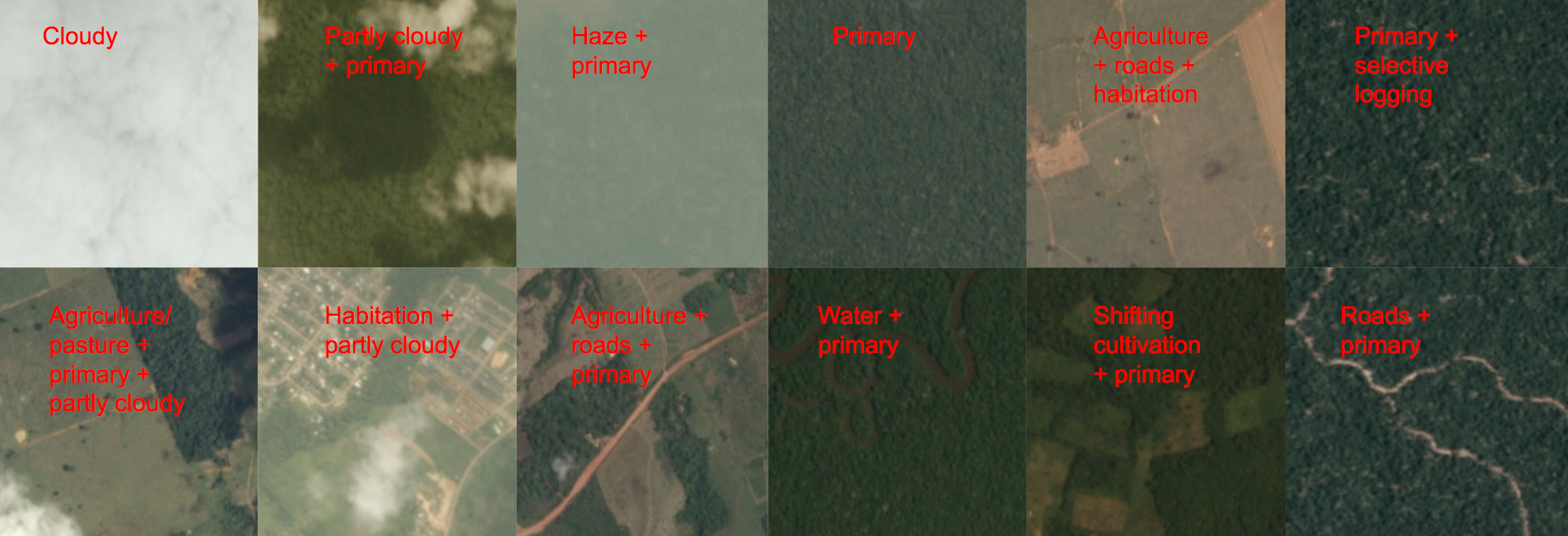}}
		{Source: \href{https://storage.googleapis.com/kaggle-competitions/kaggle/6322/media/chipdesc.jpg}{Kaggle Challenge - Planet: Understanding the Amazon from Space}}
		\caption{Sample labeled chips}
		\label{fig:sample_chip}
	\end{figure}
	To know more about labels, see \hyperref[appendix]{Appendix}.\\
	Figure \ref{fig:data_hist} illustrates the distribution of training images across different class labels.
	\begin{figure}[H]
		\centering
		\includegraphics[width=0.6\linewidth]{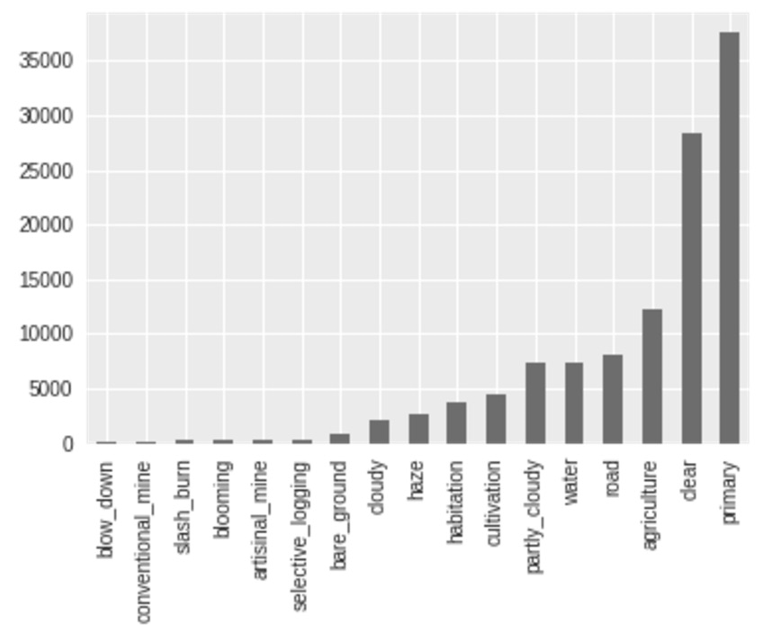}
		\caption{Number of images per class}
		\label{fig:data_hist}
	\end{figure}
	Inferring from the histogram, one can point out a considerable imbalance in the class data, i.e., some of the classes have very few training images associated with it. Hence, this is one of the more prominent challenges. Later in this section, we will figure out how to deal with this imbalance thing. Upon closer inspection, we see a structure in the labels. And the inter-relationships among labels can be exploited to design better classifiers and post-processing to weed out misclassification. For example, an image of a cloud should have no common or rare label as nothing is visible through clouds. Again when there is a water body, one would usually see greenery surrounding that (say primary forest) or human habitation or agriculture. We will deploy some methods that consider the correlation among labels and showcase the difference in the evaluation score (F$ _{2} $ score) while considering the inter-dependence among labels against their independence.
	\newpage
	\thispagestyle{empty}
	\rhead{Methods}
	\section[Chapter 3]{\Large Chapter 3}\label{chap3}
	{\Huge \textbf{Methods}}
	\vspace*{0.8cm}
	\hrule
	\subsection[ML methods]{\Large ML methods}
	Referring to various previous works \cite{comp-1, comp-2, comp-3, comp-4, comp-5}, we have a clear idea that CNN-based frameworks perform better than ML classifiers. But all they have tried is with XGBoost \cite{xgboost} or RandomForest model \cite{random-forest} - one of the best classifiers heavily used in classification challenge. Ignoring other ML models is a bit unfair because we don't know how that will perform on the given data. Since it is not guaranteed that the state-of-the-art model will always perform nicely, we opt for three more models like Gradient Boosting Classifier \cite{gbc-1}, LDA \cite{lda}, and Extra Trees Classifier \cite{etc}. We have this simple idea of aggregating 17 independent binary classifiers assuming that no labels are correlated to others. Hence, the whole setup acts as our first multi-label classifier.\\
	
	Since images are unstructured data, they can't be fed directly into ML models as they are matrices rather than being vectors. And, unlike crowdsourcing, we need a function that automatically does this -- feature assigning. So we need a function that reduces or represents these matrices into vector form. One solution can be flattening the matrix through the \enquote*{vec} operation to form a flattened vector. But with this method we invite several problems:
	\begin{enumerate}[leftmargin=*]
		\item Since our image consists of 3 channels (RGB), each of $ 256 \times 256 $ dimensions, the vector would be $ 256\times 256\times3 = 196608 $, which is enormously huge. With this many features, even a good model will perform worse due to one big reason -- the curse of dimensionality.
		\item Since our objective is to detect all the given 17 features, not segregate those features pixel-wise (segmentation), vector representation of image will become redundant as it incurs much irrelevant information. For example, a road can be encoded to some code, say $ 010 $. In other words, if somehow we could be able to map the image to a reduced form where each feature's presence or absence has some code, our job to then classify images according to these encodings would be significantly simplified. Further, we need to find a highly efficient map to extract all the relevant information occurring in the image. 
	\end{enumerate}
	Seeing the performance of models used in the ILSVRC challenge \cite{imagenet} like ResNet50 \cite{resnet}, where it stood among the best multiclass classification models (for 1000 classes), we chose this as our feature descriptor (function for an image to vector representation). To get the base framework, we removed the ANN (Artificial Neural Network) attached to the top of the frozen pre-trained CNN framework of ResNet-50. Now the output layer of our modified network is a GAP (Global Average Pooling) [See, \ref{subsec:dl_methods}] that outputs a vector of size $ 2048\times1 $, which is not that large compared to the feature vector $ 196608 $ and, at the same time, not too small to get ignored. After constructing our new encoded dataset of size = (total no. of samples$  \times 2048 $), we quickly built 17 independent ML binary classifiers with the help of the \enquote*{MultiOutput} class of scikit-learn \cite{sklearn} and trained it with the new data.\\
	
	{\noindent Before} heading towards how we trained and what other methods we devised in our training, let's discuss elaborately five different ML models we used in this project and what parameters we intend to keep for those models with what values and why?
	
	\subsubsection[LDA]{LDA}
	
	Linear discriminant analysis (LDA) \cite{lda} (also called normal discriminant analysis (NDA), or discriminant function analysis) is a generalization of Fisher's linear discriminant that, through a linear combination of features, characterizes or separates two or more classes of objects. And the result is used as a linear classifier or, more commonly, a dimensionality reduction technique. During training, it learns the most discriminative axes between the classes, which then is used to define a hyperplane onto which the data are projected. This projection respects the condition that the classes must be kept as far as possible.\\
	
	{\noindent\textbf{Assumptions}}: The analysis is pretty sensible to outliers, and the smallest group must be larger than the number of input features.
	\begin{enumerate}[leftmargin=*]
		\item \textit{Multivariate normality}: Input variables must follow Gaussian distribution for each level of the grouping variable.
		\item \textit{Homogeneity of variance/covariance (homoscedasticity)}: Covariances among the classes (say number of classes $ =n $) are identical, so ${ \Sigma_{0}=\Sigma_{1}= \ldots = \Sigma_{n} = \Sigma}$ with covariance matrix being full rank (can be tested with Box's M statistic). However, it has been suggested that when covariances are equal, use linear discriminant analysis, or else, the quadratic discriminant analysis may be used.
		\item \textit{Multicollinearity}: With an enhanced correlation between independent variables, the predictive ability can decline.
		\item \textit{Independence}: Participants are assumed to be randomly sampled, and their scores on a given variable must be independent of each other.
	\end{enumerate}
	
	{\noindent\textbf{Discrimination Rules}}: There are three discrimination rules for this method to date, namely 
	\begin{itemize}[leftmargin=*]
		\item \textit{Maximum likelihood}, where $ x $ is assigned to the group that maximizes population density,
		\item \textit{Bayes Discriminant Rule}, which goes along the maximum likelihood method, and instead of population density, it goes for maximum $ \pi_{i}f_{i}(x)$, where $ \pi_{i} $ represents the prior probability of that classification, and $ f_{i}(x) $ represents the population density.
		\item  And lastly, \textit{Fisher's linear discriminant rule}, where the ratio between $ SS_{between} $ and $ SS_{within} $ is maximized, and a linear combination of the predictors is found to predict the group.
	\end{itemize}
	
	{\noindent We'll} use Fisher's linear discriminant rule for our purpose, which is claimed to be better for high dimensional data than the existing one. Let's understand it in the following steps:
	\begin{enumerate}[leftmargin=*]
		\item First, compute the within-class scatter matrix, i.e., $$ S_{w} = \sum\limits_{i=1}^{c}S_{i} $$ where $ c = $ the total number of distinct classes, $ S_{i} = \sum\limits_{j=1}^{n_{i}}(x_{ij} - m_{i})(x_{ij} - m_{i})^{T} $, $ m_{i} = \dfrac{1}{n_{i}}\sum\limits_{j=1}^{n_{i}}x_{ij} $, $ x $ is a sample (i.e. row) and $ n_{i} $ is the total number of samples in class $ i $. Then compute between-class scatter matrix, .e., $$ S_{B} =  \sum\limits_{i=1}^{c}n_{i}(m_{i}-m)(m_{i}-m)^{T}$$ where $ m = \dfrac{1}{n}\sum\limits_{k=1}^{n}x_{k} $ and $ n = $ total number of samples.
		\item Second, estimate the eigenvectors and corresponding eigenvalues for the dot product of inversed within-class scatter matrix and between-class scatter matrix ($ S_{W}^{-1}S_{B} $).
		\item Sort the eigenvalues and select the top $ k $ (as per the requirement of dimensions) as the eigenvectors with the highest eigenvalues carry the most information about the data distribution.
		\item Construct a new matrix containing the eigenvectors that map to the chosen $ k $ eigenvalues.
		\item Obtain the new features (i.e., LDA components) using the dot product of the data and the matrix from step 4.
	\end{enumerate}

	\subsubsection[Random Forest]{Random Forest}
	
	Random forests \cite{random-forest} or \textit{random decision forests} are an \hyperref[sec:ensemble]{ensemble learning algorithm}, generally trained via the \hyperref[sec:bagging]{bagging} method for classification, regression, and other tasks that operate by taking an ensemble of \href{https://en.wikipedia.org/wiki/Decision_tree_learning}{decision trees} \cite{decision-tree-learning} at training time. For classification tasks, the class selected by most trees is the output of the random forest, while for regression tasks, it is the mean or average prediction of the individual trees. While constructing nodes of decision trees top-down, we use different \href{https://en.wikipedia.org/wiki/Decision_tree_learning#Metrics}{metrics} \cite{dtl-metrics} to measure the \enquote{best} split among features.\\ 
	
	{\noindent Random forests} use a modified tree learning algorithm that picks a random subset of the features at each candidate split in the training process called \enquote{feature bagging.} By doing this, the correlation among trees in an ordinary bootstrap sample can be reduced. For instance, if one or a few features are robust predictors for the target output, they will be selected in many $ B $ trees, making them more correlated.\\
	
	{\noindent Usually,} for a classification problem with $ n $ features, $ \sqrt{n} $ (rounded down) features are used in each split. At the same time, for regression, the authors recommend $ n/3 $ (rounded) with a minimum node size of $ 5 $ as the default. In practice, the problem decides the best values for these parameters, and they should be treated as hyperparameters.
	
	\subsubsection[Extra Trees Classifier]{Extra Trees Classifier}
	
	Extra Trees Classifier \cite{etc} (also called \textit{Extremely Randomized Trees}) is a derivative of Random Forest Classifier where one further step of randomization is added while constructing nodes. The two main differences between Extra Trees and Random Forest  are:
	\begin{enumerate}[leftmargin=*]
		\item Each tree uses the whole learning sample (rather than a bootstrap sample) for its training.
		\item Second, the top-down splitting in the tree learner is randomized. Instead of computing the locally optimal cut-point for each predictor under consideration (based on, e.g., \href{https://en.wikipedia.org/wiki/Information_gain}{information gain} \cite{ig} or the \href{https://en.wikipedia.org/wiki/Gini_impurity}{Gini impurity} \cite{gi}), a random cut-point is selected from a uniform distribution within the feature's empirical range (in the tree's training set). Then, among all the randomly generated splits, the split that generates the highest score is selected to split the node. Like ordinary random forests, we can also specify the number of randomly selected features to be considered at each node. Default values for this parameter are $ \sqrt{n} $ for classification and $ n $ for regression, where $ n= $ the number of features in the model.
	\end{enumerate}
	
	\subsubsection[Gradient Boosting Classifier (GBC)]{Gradient Boosting Classifier (GBC)}
	
	Gradient boosting \cite{gbc-1} is a machine learning technique for regression, classification, and other tasks, which produces a prediction model by combining weak \enquote{learners} into a single strong learner (ensemble of weak prediction models, typically decision trees) in an iterative fashion. In gradient boosted trees, a decision tree is usually a weak learner, which outperforms random forest. Here models are built in a stage-wise manner as other \hyperref[sec:boosting]{boosting} methods do, and it generalizes them by optimizing an arbitrary differentiable loss function.\\
	
	Let's take an example in the \textit{least-squares regression} setting, where our goal is to \enquote{teach} a model $ F $ to predict values $ \hat{y} = F(x) $ by minimizing the mean squared error $ \dfrac{1}{n}\sum_{i=1}^{n}\left(\hat{y}_{i} - y_{i}\right)^{2} $, where
	\begin{itemize}[leftmargin=*]
		\item $ \hat{y}_{i} = $ the predicted value $ F(x) $
		\item $ y_{i} = $ the observed value
		\item $ n = $ number of samples
	\end{itemize}
	Consider a Gradient Boosting Method of $ M $ stages and we chose some imperfect model $ F_{m} $, where $ 1\leq m\leq M$. Suppose $ F_{m} $ simply returns $ \hat{y}_{i} = \bar{y} $ for some low $ m $, where $ \bar{y} = $ mean of $ y $. To improve $ F_{m} $, our algorithm adds a new estimator, $ h_{m}(x) $ that transforms to some new model 
	\begin{align*}
		F_{m+1}(x) &= F_{m}(x) + h_{m}(x) = y\\
		\implies h_{m}(x) &= y - F_{m}(x)
	\end{align*}
	which implies GB algorithm tries to fit $ h $ to the residual $ y - F_{m}(x) $. That basically points to the conclusion that $ F_{m+1} $ attempts to rectify its predecessor $ F_{m} $ by correcting its error. Now the fact that residuals $ h_{m}(x) $ is equal to the negative gradients of the mean squared error (MSE) loss function (with respect to $ F(x) $), which generalize this idea to loss function other than squared error.
	\begin{align*}
		L_{MSE} &= \dfrac{1}{2}\left(y - F(x)\right)^{2}\\
		h_{m}(x) &= -\dfrac{\partial L_{MSE}}{\partial F} = y - F(x)
	\end{align*}
	This specifies GB could be specialized to gradient descent algorithm, and its generalized version can be plugged in a diferent loss and gradient.\\
	\vspace*{0.5cm}
	\noindent
	\scalebox{0.86}{\begin{algorithm}[H]
			\SetAlgoLined
			\KwIn{training set $\left\{\left(x_{i}, y_{i}\right)\right\}_{i=1}^{n}$, a differentiable loss function $ L(y, F(x)) $, number of iterations $ M $.}
			1. Initialize model with a constant value: $$ F_{0}(x) = \underset{\gamma}{\arg\min}\sum\limits_{i=1}^{n}L(y_{i}, \gamma) $$
			2. \For{$ m = 1 $ to $ M $}
			{
				(a) Compute pseudo-residuals: $$r_{im} = -\left[\dfrac{\partial L(y_{i}, F(x_{i}))}{\partial F(x_{i})}\right]_{F(x) = F_{m-1}(x)}$$ for $ i = 1, 2, \ldots, n $\
				
				(b) Fit a base (or weak) learner (e.g., tree) $ h_{m}(x) $ to pseudo-residuals, i.e., train it using the training set $\left\{\left(x_{i}, r_{im}\right)\right\}_{i=1}^{n}$\
				
				(c) Compute multiplier $ \gamma_{m} $ by solving the following optimization problem in one dimension: $$\gamma_{m} = \underset{\gamma}{\arg\min}\sum\limits_{i=1}^{n}L\left(y_{i}, F_{m-1}(x_{i})+ \gamma h_{m}(x_{i})\right)$$\
				
				(d) Update the model: $$F_{m}(x) = F_{m-1}(x) + \gamma_{m}h_{m}(x)$$
			}
			3. \KwOut{$ F_{M}(x) $}
			\caption{Gradient Boosting Algorithm}
	\end{algorithm}}\\
	\noindent
	At the $ m^{th} $ step in gradient boosting a decision tree $ h_{m}(x) $ is fitted to pseudo-residuals. Let $ K_{m} $ = the number of its leaves. Then the possible number of partitions of the input space by the decision tree $ = K_{m} $ which produces disjoint regions viz. $ R_{1}, \ldots, R_{K_{m}} $. In each region the tree predicts a constant value. Using the indicator notation, the image of $ h_{m} $ for input $ x $ can be written as the sum: $$h_{m}(x) = \sum\limits_{k=1}^{K_{m}}b_{km}\mathbf{1}_{R_{km}}(x)$$ where $ b_{km} $ is the value predicted by the tree in the region $ R_{km} $.
	\subsubsection[XGBoost]{XGBoost}
	
	As said by \href{https://www.quora.com/What-is-the-difference-between-the-R-gbm-gradient-boosting-machine-and-xgboost-extreme-gradient-boosting/answer/Tianqi-Chen-1}{\textit{Tianqi Chen}} \cite{tc-quora}, the author of the XGBoost algorithm, that both XGBoost and GBC follow the principle of gradient boosting. However, the difference is there in modeling details, where specifically, XGBoost used a more regularized model formalization to control over-fitting, which gives it better performance.
	
	\subsection[Cross-Validation]{\Large Cross-Validation}
	
	We invoked Stratified 5-fold Cross-Validation (CV) \cite{lda, cv-1, cv-2, cv-3, cv-4} strategy with F$ _{2} $ as our evaluation metric to have a fair comparison among the models. Why F$ _{2} $? We will discuss that later. Now the question comes why we chose the CV method for comparison, which is computationally more expensive; why not just the score on prediction of test data? How CV ensures fairness in this game? And what is stratified? Let's discuss one by one in detail what they all imply.
	\begin{itemize}[leftmargin=*]
		\item \textbf{Cross-validation}: It (also called \textit{rotation estimation} or \textit{out-of-sample testing}) is a model validation technique for assessing how accurately a predictive model will perform in practice. The goal of cross-validation is to flag problems like overfitting or selection bias and give an insight into how the model will generalize to an unseen dataset. There are several sub-methods in cross-validation, among which one is Stratified K-Fold cross-validation. But before that, we'll know about what is \enquote{holdout} method.
		\item \textbf{Holdout Method}: In this method, we remove a part of the training data, which we call validation set (or \textit{dev set} or \textit{development set}), and use it to obtain predictions from the model trained on the rest of the data. However, suppose the validation set is too small, then, in that case, our model predictions will be imprecise, and there is a high chance that it will be suffering from a high variance issue as it is not ensured which data points will end up in the validation set. Plus, the result might be different for different sets, like the test set. 
		\item \textbf{Solution}: In K-Fold cross-validation methods, the data is randomly partitioned into k equal-sized subsets. Now, of the k subsets, one subset is retained as the validation set, and the rest k-1 subsets are put together to form a training set. And in this manner,  this "holdout" method is repeated k times. The error estimation from k trials is averaged to get the total effectiveness of our model in the form of a single estimate. We can see, every instance gets to be in a validation set exactly once and in a training set k-1 times. And this significantly reduce the bias as we use most of the data for fitting. The variance, too, drops considerably as most of the data is also being used in the validation set. Due to this, we generally use K-Fold CV as a measure of model evaluation.
		\item \textbf{Stratified K-Fold cross-validation \label{strcv}}: In stratified k-fold cross-validation, the \textit{mean response value} is approximately equal in all the partitions of the given data. As in binary classification, each partition contains roughly the exact proportions of the two types of class labels.
	\end{itemize}
	
	Despite the high computational cost, CV shows us a clear picture of the generalizability as it covers both the strength and weaknesses of every model at each split, which brings all of them to the same ground. So, for example, training the same model independently on five mutually exclusive data tends to produce five different results due to the uniqueness of each split. And taking their average indeed ascertain about the performance it will show when it's not too overfitted nor too biased - just the perfect.\\
	
	{\noindent\textbf{\textit{Remarks}:} Peformance will be discussed in Chapter \ref{chap4}.}
	
	\subsection[DL Methods]{\Large DL Methods}\label{subsec:dl_methods}
	
	So far, so good. Now we'll implement the most promising group of frameworks that guarantees to perform well on image datasets and has consistently been used over by most data scientists and computer vision engineers for their tasks and challenges. Unlike previous methodologies, in this one, the DL architecture directly feeds upon the images and uses their pixel pattern to arrive at a meaningful decision about all the 17 labels. Here the image is being pre-processed multiple times throughout the network before outputting the required information. Each layer can be considered a function that is nothing but convolution filters that slide upon the input stack of matrices in left to right and top to the bottom manner with a user input stride. After taking the sum of the element-wise product (for Conv2D) and maximum value (as in Maxpooling) for each filter, we get a stack of matrices for each filter, which then visits the same operations or gets flattened to produce the information so far processed in vector form.\\
	
	{\noindent Before moving into some complex terminology}, let's discuss them in brief.
	\begin{enumerate}[leftmargin=*]
		\item \textit{Input Layer}: Usually, the raw input of the image is held in this layer in the form of a stack of matrices obtained from the corresponding image with width $ = n_{w} $, height $ = n_{h} $, and depth = number of channels $ = n_{c} $.
		\item \textit{Convolutional layer} (or \textit{Conv2D} layer): At this layer, the dot product between all the kernels and the input image patch or the transformed image patch occurs. Suppose a total of $ 12 $ filters, each of size $ n_{f}\times n_{f}\times3 $, are used for this layer, then we'll get an output volume of dimension $ 32 \times 32 \times 12 $ (for "same" padding) for an input image of $ 32\times32\times3 $.
		\item \textit{Activation Layer}:  This layer computes an element-wise activation function to the output of the preceding convolution layer. Common activation functions are $ ReLU: \max(0, x) $, $ sigmoid: 1/(1+e^{-x}) $, $ tanh $, Leaky RELU, etc. Our case study will use $ ReLU $ in hidden layers and $ sigmoid $ and $ softmax $ in the output layer.
		\item \textit{Padding}: It is a process of image transformation where simply layers of zero are added to the input image in order to avoid the following issues.
		\begin{enumerate}[leftmargin=*]
			\item When on an image of size $ n\times n $, a filter of size $ f\times f $ is run, we get the resultant image to be of size $ (n-f+1)\times(n-f+1) $. For $ f>1 $, the image shrinks whenever convolution is performed. And this sets an upper limit to the number of times such an operation could be executed before the image reduces to nothing, thereby restraining us from building deeper networks.
			\item Further, the pixels on the corners and the edges of the image are used much less than those in the middle. For example, when on a 6*6 image, a $ 3\times3 $ filter is convoluted, then the pixel at one of the corner is visited only once, while the one at the edge is seen 3 times, and the pixel at $ 3^{rd} $ row and $ 3^{rd} $ column touched 9 times.
		\end{enumerate}
		Again there are two types of padding:
		\begin{enumerate}[leftmargin=*]
			\item \textit{Valid padding}: Here, as the name suggests, no alteration is made to the image, i.e., no padding is done.
			\item \textit{Same Padding}: Here, as the name suggests, we add that amount of padding layers such that the output image has the exact dimensions as the input image. In our case, we've used only this padding.
		\end{enumerate}
		\item \textit{Stride}: This is a parameter of the neural network's filter that defines the amount of kernel movement in terms of pixels shifts over the image, video, or any input matrix. When the stride is one, the filters will move $ 1 $ pixel or one unit. When the stride is two, we carry the filters to $ 2 $ pixels at a time. This component of the neural network handles the compression of image and video data.
		\item \textit{Max-Pooling}: It is a \textit{sample-based discretization process} that down-samples an input representation (image, hidden-layer output matrix, etc.) by selecting the maximum element from the region of the feature map covered by the filter. This helps reducing dimensionality and allows assumptions to be made about features contained in the sub-matrices binned. Hence, after the max-pooling layer, the output feature map includes the most prominent features of the previous feature map.\\
		Why max-pooling?: 
		\begin{itemize}[leftmargin=*]
			\item This is done to correct over-fitting by providing an abstracted form of the image.
			\item It further reduces the computational expense by reducing the number of parameters to learn and allows fundamental translation invariance to the internal representation.
		\end{itemize}
		\item \textit{Average-Pooling}: It provides the mean pixel values present in the region of the feature map covered by the kernel. Thus, unlike max-pooling, which gives the most prominent feature in a particular patch of the feature map, average pooling provides the average features present in that patch.
		\item \textit{Global Average-Pooling (GAP)}: It helps to reduce each channel in the feature map to a single real value. For example, an $ n_{h} \times n_{w} \times n_{c} $ feature map is reduced to $ 1 \times 1 \times n_{c} $ feature map, equivalent to using a max-pooling filter of full size, i.e., dimension $ nh\times nw $.
		\item \textit{Batch-Normalization (BN)}: It is a technique proposed by \textit{Sergey Ioffe} and \textit{Christian Szegedy} \cite{bn} in 2015 for training artificial neural networks that normalizes the layers' inputs by re-centering and re-scaling for each mini-batch.And this satisfies the assumptions made by the subsequent layer about the spread and distribution of the preceding layer's output which stabilizes the learning process and makes it faster by dramatically reducing the number of training epochs required to train deep networks. Plus, it can act as a regularization method that avoids the overfitting of the model. Therefore, it is often placed just after defining the sequential model and after the convolution and pooling layers.\\
		
		To date, there isn't any particular reason behind its effectiveness. And it is still under discussion. The prime intention with BN is to mitigate the issue of \textit{internal covariate shift}, where random parameter initialization and the following changes in the distribution of each layer's inputs affect the learning rate of the network. To know more about internal covariate shift, please click \href{https://en.wikipedia.org/wiki/Batch_normalization#Motivation:_The_phenomenon_of_internal_covariate_shift}{here}.\\
		
		\textbf{BN Transform Procedure \cite{bn}}:\\
		It simply fixes the means and variances of each layer's input. Ideally, this normalization should be performed over the whole batch, i.e., the entire training set, but to use this jointly with \href{https://en.wikipedia.org/wiki/Stochastic_optimization}{stochastic optimization} methods, it is impractical to use the global information. Hence, BN is limited to each mini-batch in the training procedure.\\
		Denote $ B_{i} = i^{th}$ batch of size $ m $ of the entire training set $ \forall\ i = 1, \ldots, n $, where a total of $ n $ splits from $ m $ sized training set is possible.\\
		\scalebox{0.95}{\begin{algorithm}[H]
				\SetAlgoLined
				\KwIn{For a layer, we have $ d $-dimensional input, $ x = (x^{(1)}, \ldots, x^{(d)}) $ where each dimension of the input is normalized (i.e., re-centered and re-scaled) independently.}
				\For{$ i = 1 $ to $ n $}{
					1. Compute $ \mu_{B_{i}} = \dfrac{1}{m}\sum\limits_{j=1}^{m}x_{ij} $ and $ \sigma_{B_{i}}^{2} = \dfrac{1}{m}\sum\limits_{j=1}^{m}\left(x_{ij}-\mu_{B_{i}}\right)^{2} $\;
					2. Denote $ \hat{x}_{ij}^{(k)} = \dfrac{x_{ij}^{(k)} - \mu_{B_{i}}^{(k)}}{\sqrt{\sigma_{B_{i}}^{(k)^{2}} + \epsilon}} $, where $ k \in \{1, \ldots, d\} $ and $ j \in \{1, \ldots, m\} $; $ \mu_{B_{i}}^{(k)} $ and $ \sigma_{B_{i}}^{(k)^{2}} $ are the per-dimension mean and variance respectively. $ \epsilon = $ arbitrarily small positive value added in the denominator for numerical stability. If $ \epsilon $ is not considered, then $ \hat{x}_{i}^{(k)} $ have zero mean and unit variance.\;
					3. Transformation Step: Restoring the representation power of the network $$y_{ij}^{(k)} = \gamma_{i}^{(k)}\hat{x}_{ij}^{(k)} + \beta_{i}^{(k)}$$ where $ \gamma_{i}^{(k)} $ and $ \beta_{i}^{(k)} $ are learned subsequently in the optimization procedure.
				}
				\KwOut{$ y_{i}^{(k)} = $ the transformed version of $ x_{i}^{(k)} $}
				\caption{Batch-Normalization Algorithm}
		\end{algorithm}}
	
		Formally, Batch-Normalization transform $ BN_{\gamma_{i}^{(k)}, \beta_{i}^{(k)}} $ maps $ x_{i}^{(k)} $ to $ y_{i}^{(k)} $ , i.e., $ y_{i}^{(k)}  = BN_{\gamma_{i}^{(k)}, \beta_{i}^{(k)}}(x_{i}^{(k)})$ which is then passed to other network layers, while the normalized output $ x_{i}^{(k)} $ remains internal to the current layer.
	\end{enumerate}
	\newpage
	\subsubsection[Our First DL Model - Baseline Model]{\large Our First DL Model - Baseline Model}
	
	Let's see our first CNN framework, which serves as our baseline model.
	\begin{figure}[H]
		\centering
		\includegraphics[width=0.8\linewidth, height=0.75\linewidth]{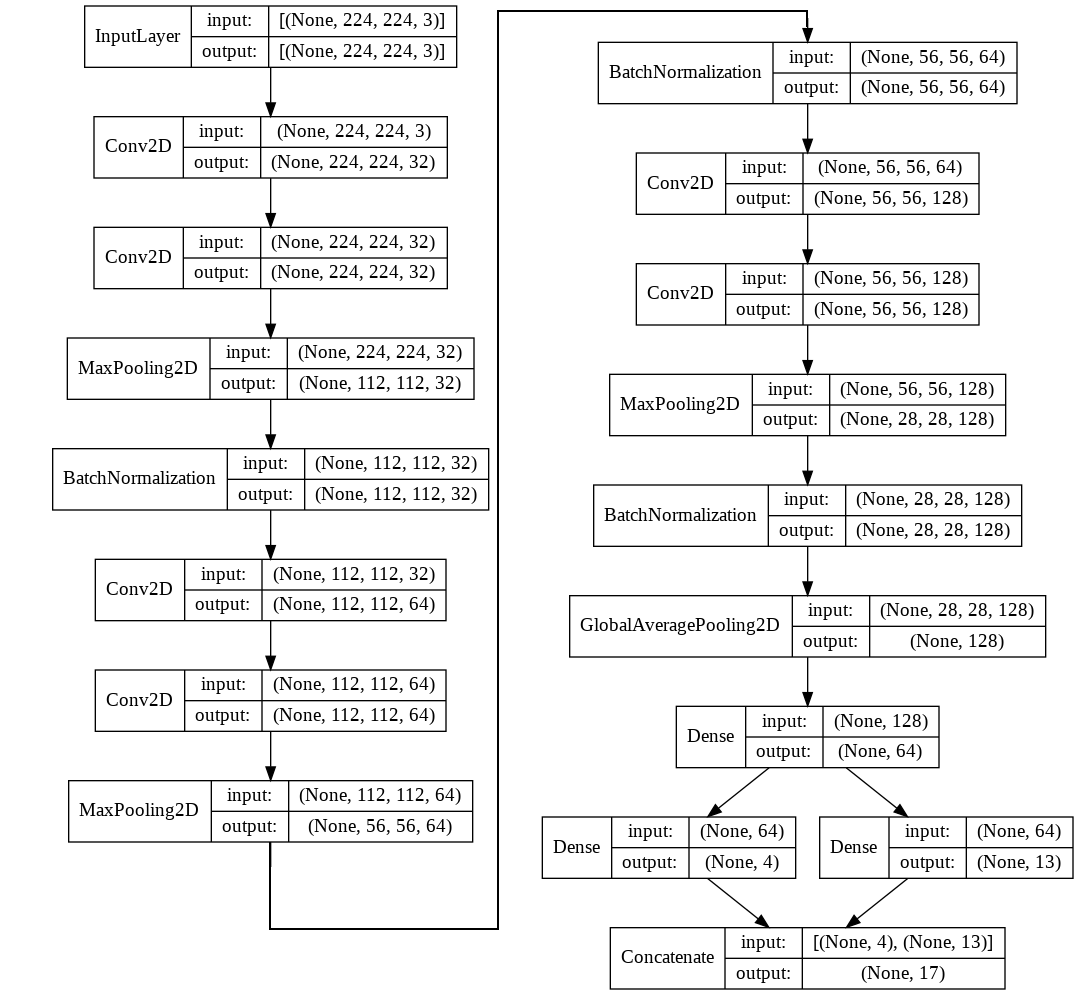}
		\caption{Baseline Model}
		\label{fig:baseline}
	\end{figure}
	The output vector of $ 64\times 1 $ consists of information regarding weather and land use/ land cover. At this point, we attach a dense neural network whose number of hidden layers and number of neurons in those layers is decided according to the performance in different topologies. Thus, the output layer is again a dense layer of 17 units which ideally should provide us the collection of labels corresponding to the features present.\\
	
	Unlike in multiclass classification, where each instance has only one label, here, our training images were by no means limited to one weather label. However, this softmax loss function, which is advantageous for single-label classification, is not well suited for multi-label classification for various reasons, most glaringly due to the fact that each label is no longer mutually exclusive in a multi-label classification task. And the common fix for this is to alter the loss function from softmax CE loss to sigmoid CE loss. Sigmoid CE loss (or Binary Cross Entropy  - BCE) is primarily used for binary classification and so can evaluate which labels out of $ N (= 17) $ possibility should be set to \enquote*{on} as opposed to selecting a single best label \cite{10-2}.\\
	
	{\noindent Mathematically,}
	\begin{itemize}[leftmargin=*]
		\item The standard (unit) softmax function $ \sigma:\mathbb{R}^{K} \longrightarrow [0,\ 1]^{K}$ is defined by the formula $$\sigma(\mathbf{z})_{i} = \dfrac{e^{z_{i}}}{\sum_{j=1}^{K}e^{z_{j}}}$$ for $ i = 1,\ldots, K $ and $ \mathbf{z} = (z_{1}, \ldots, z_{k})\in \mathbb{R}^{K} $, where $ K = $ input feature-space dimension.
		\item Sigmoid curve is a characteristic \enquote*{S}-shaped curve with following equation, where $ S:\mathbb{R}\longrightarrow \mathbb{R} $
		$$S(x) = \dfrac{1}{1 + e^{-x}}$$ with the property that $ S(x) = 1 - S(-x) $.
	\end{itemize}
	However, this approach does not use a unique structure within our dataset, motivating our more advanced softmax-sigmoid joint loss approach.  In order to exploit unique weather labels (as every satellite image has one and only one weather label) and multiple (zero to many) land use/land cover labels, we moved towards a hybrid loss function that applied softmax loss for weather tags (with four dense neurons of the last layer) and sigmoid CE loss for the rest of the labels (with 13 dense neurons).\\
	
	We had used Adam \cite{adam} with a batch size of 128 (depending upon GPU memory and model size) for the optimization process. And this configuration will be the same for the rest of the upcoming CNN models. So let's see what Adam with \textit{AMSGrad = True} \cite{amsgrad} supposedly means.
	\begin{itemize}[leftmargin=*]
		\item Adam (short for \textbf{Ada}ptive \textbf{M}oment Estimation) is different from classical stochastic gradient descent as it combines the specialties of two other extensions of stochastic gradient descent, namely:
		\begin{enumerate}[leftmargin=*]
			\item Adaptive Gradient Algorithm (\textit{AdaGrad}) maintains a per-parameter learning rate to improve performance on problems with sparse gradients.
			\item Root Mean Square Propagation (\textit{RMSProp} or \textit{RMSP}) is also an adaptive learning algorithm that maintains a per-parameter learning rate that is adapted based on the mean gradients for the weight. Further, it uses second moments of the gradients.
		\end{enumerate}
		Now let's unravel the methods and steps behind this optimization technique one by one.\\
		Adam involves a combination of both \textit{momentum} and \textit{RMSP}.
		\begin{enumerate}[leftmargin=*]
			\item \textit{Momentum}: It helps our gradient descent algorithm to accelerate by taking into consideration the \enquote{exponentially weighted average} of the gradients. Averaging the gradients makes the algorithm converge towards minima at a faster pace.
			$$w_{t+1} = w_{t} - \alpha_{t} m_{t}$$ where, $$m_{t} = \beta m_{t-1} + (1-\beta)\left[\dfrac{\partial L}{\partial w_{t}}\right]$$
			\begin{itemize}[leftmargin=*]
				\item $ m_{t} = $ aggregate of gradients at current time $ t $ (at $ t=0 $, $ m_{t} = 0 $)
				\item $ m_{t-1} = $ aggregate of gradients at time $ t-1 $
				\item $ w_{t} = $ weights at time $ t $
				\item $ w_{t+1} = $ weights at time $ t+1 $
				\item $ \alpha_{t} = $ learning rate at time $ t $ 
				\item $ \dfrac{\partial L}{\partial w_{t}} $ = derivative of Loss function w.r.t weights at time $ t $
				\item $ \beta = $ Moving average parameter (const $ = 0.9 $)
			\end{itemize}
			\item \textit{RMSP}: It is an upgraded version of AdaGrad where instead of taking the cumulative sum of squared gradient descents like in AdaGrad, it takes the \enquote{exponential moving average.}
			$$w_{t+1} = w_{t} - \dfrac{\alpha_{t}}{\sqrt{v_{t}} + \epsilon}\left[\dfrac{\partial L}{\partial w_{t}}\right]$$ where, $$v_{t} = \beta v_{t-1} + (1-\beta)\left[\dfrac{\partial L}{\partial w_{t}}\right]^{2}$$
			\begin{itemize}[leftmargin=*]
				\item $ v_{t} = $ sum of square of past gradients (i.e., sum of $ \dfrac{\partial L}{\partial w_{t-1}} $ for the preceding time $ = t-1 $). (Initially, $ v_{t} = 0 $)
				\item $ \epsilon = $ A small positive constant $  = 10^{-8} $
			\end{itemize}
			\item Now combining the above two, we get the following: $$m_{t} = \beta_{1} m_{t-1} + (1-\beta_{1})\left[\dfrac{\partial L}{\partial w_{t}}\right]$$ and $$v_{t} = \beta_{2} v_{t-1} + (1-\beta_{2})\left[\dfrac{\partial L}{\partial w_{t}}\right]^{2}$$
			Parameters used:
			\begin{itemize}[leftmargin=*]
				\item $ \epsilon =  $ A small positive constant ($ 10^{-8} $) to avoid \enquote*{division by 0} error when ($ v_{t} = 0 $).
				\item $ \beta_{1} \ \&\ \beta_{2} = $ decay rates of average of gradients in the above two methods. ($ \beta_{1} = 0.9\ \&\ \beta_{2} = 0.999 $)
				\item $ \alpha = $ Step size parameter or learning rate (default $ = 0.001 $)
			\end{itemize}
			Since, \enquote*{$ m_{t} $} and \enquote*{$ v_{t} $} are both initialized as zero, it is observed that they both have a tendency to be \enquote*{biased towards zero} as both $ \beta_{1}$ and $ \beta_{2} = 1 $. To fix this, we compute \enquote*{bias corrected} \enquote*{$ m_{t} $} and \enquote*{$ v_{t} $,} which also control the weights while reaching the global minimum to prevent high oscillation when near it. \enquote*{Bias Corrected} formula
			\begin{align*}
				\hat{m}_{t} &= \dfrac{m_{t}}{1- \beta_{1}^{t}}\\
				\hat{v}_{t} &= \dfrac{v_{t}}{1- \beta_{2}^{t}}
			\end{align*}
			Plugging the bias-corrected weight parameters \enquote*{$ \hat{m}_{t} $} and \enquote*{$ \hat{v}_{t} $}, we get our general equation in Adam as follows: $$w_{t+1} = w_{t} - \hat{m}_{t}\left(\dfrac{\alpha}{\sqrt{\hat{v}_{t}} + \epsilon}\right)$$
			\item Summarising all the above steps, we get as follows.\\
			\begin{algorithm}[H]
				\SetAlgoLined
				\KwIn{$ \alpha $: Stepsize}
				\KwIn{$ \beta_{1},\ \beta_{2} \in \mathopen{]}0, 1\mathclose{]} $: Exponential decay rates for the moment estimates}
				\KwIn{$ L(w) $: Stochastic objective function with parameters $ w $}
				\KwIn{$ w_{0} $: Initial parameter vector}
				$ m_{0} \longleftarrow 0$  (Initialize $ 1^{st} $ moment vector)\;
				$ v_{0} \longleftarrow 0$  (Initialize $ 2^{nd} $ moment vector)\;
				$ t \longleftarrow 0$  (Initialize timestep)\;
				\While{$ w_{t} $ not converged}{
				$ t \longleftarrow t+1 $\;
				$ g_{t}\longleftarrow \nabla_{w}L_{t}(w_{t-1}) $ (Get gradients w.r.t. stochastic objective at timestep $ t $)\;
				$ m_{t} \longleftarrow \beta_{1} m_{t-1} + (1-\beta_{1}).g_{t} $ (Update biased first moment estimate)\;
				$ v_{t} \longleftarrow \beta_{2} v_{t-1} + (1-\beta_{2}).g_{t}^{2} $ (Update biased second raw moment estimate)\;
				$ \hat{m}_{t} \longleftarrow \dfrac{m_{t}}{1- \beta_{1}^{t}} $ (Compute bias-corrected first moment estimate)\;
				$ \hat{v}_{t} \longleftarrow \dfrac{v_{t}}{1- \beta_{2}^{t}} $  (Compute bias-corrected second raw moment estimate)\;
				$ w_{t+1} = w_{t} - \hat{m}_{t}\left(\dfrac{\alpha}{\sqrt{\hat{v}_{t}} + \epsilon}\right) $ (Update parameters)\;
				}
				\KwOut{$ w_{t} $ (Resulting parameters)}
				\caption{Adam, our proposed algorithm for stochastic optimization. $ g_{t}^{2} $	indicates the elementwise square $ g_{t}\odot g_{t} $. All operations on vectors are element-wise. With $ \beta_{1}^{t} $ and $ \beta_{2}^{t} $ we denote $ \beta_{1} $ and $ \beta_{2} $ raised to the power $ t $.}
			\end{algorithm}
		\end{enumerate}
		\item With AMSGrad \cite{amsgrad} $ = True $, we seek to fix a convergence issue with Adam-based optimizers. It uses the maximum of past squared gradients $ v_{t} $ rather than the exponential average to update the parameters:\\
		\begin{algorithm}[H]
			\SetAlgoLined
			\KwIn{Same as above}
			$ m_{0} \longleftarrow 0$  (Initialize $ 1^{st} $ moment vector)\;
			$ v_{0} \longleftarrow 0$  (Initialize $ 2^{nd} $ moment vector)\;
			$ t \longleftarrow 0$  (Initialize timestep)\;
			\While{$ w_{t} $ not converged}{
				$ t \longleftarrow t+1 $\;
				$ g_{t}\longleftarrow \nabla_{w}L_{t}(w_{t-1}) $ (Get gradients w.r.t. stochastic objective at timestep $ t $)\;
				$ m_{t} \longleftarrow \beta_{1} m_{t-1} + (1-\beta_{1}).g_{t} $ (Update biased first moment estimate)\;
				$ v_{t} \longleftarrow \beta_{2} v_{t-1} + (1-\beta_{2}).g_{t}^{2} $ (Update biased second raw moment estimate)\;
				$ \hat{m}_{t} \longleftarrow \dfrac{m_{t}}{1- \beta_{1}^{t}} $ (Compute bias-corrected first moment estimate)\;
				$ \hat{v}_{t} \longleftarrow \dfrac{v_{t}}{1- \beta_{2}^{t}} $  (Compute bias-corrected second raw moment estimate)\;
				$ \hat{v}_{t} \longleftarrow \max\left(\hat{v}_{t-1}, v_{t}\right) $ (The only update in AMSGrad)\;
				$ w_{t+1} = w_{t} - \hat{m}_{t}\left(\dfrac{\alpha}{\sqrt{\hat{v}_{t}} + \epsilon}\right) $ (Update parameters)\;
			}
			\KwOut{$ w_{t} $ (Resulting parameters)}
			\caption{AMSGrad, an improvement to Adam.}
		\end{algorithm}
	\end{itemize}

	\subsubsection[ImageNet Models]{\large ImageNet Models}

	As discussed earlier, these models \cite{keras-app} are claimed to give a very accurate performance in the Image classification problem. Thus, for us, the headache for deciding on deep network architecture and the breathtaking work of training such a vast network are both resolved. But the issue with such a network is their training. As they have learned the distribution of previous data, their weights can respond well and perform better in prior datasets than on our image data, even though the objective of both works is the same. So now, here comes the concept of fine-tuning. In this method, a bunch of top layers in the given pre-trained model, where the complex features are learned from the image, are allowed to train (update their weights) with a meager learning rate. This helps our model get acclimatized to our dataset, i.e., essentially learning the data distribution. More formally, \textit{fine-tuning} is the method in which parameters of a model must be adjusted very precisely in order to fit with specific observations. This is an approach of transfer learning, where we train the same model with the holdout set. Usually, we reduce the learning rate such that it does not have a major influence on the previously set weights.\\
	
	{\noindent Now} let's state the transfer learning operation \cite{transfer-learning-def} mathematically to have a clear insight into this idea.
	\begin{definition}[Domain]
		A domain $\mathcal{D}$ is composed of two parts, i.e., a feature space $\mathcal{X}$ and a marginal distribution $P(X)$. In other words, $\mathcal{D}=\{\mathcal{X}, P(X)\} .$ And the symbol $X$ denotes an instance set, which is defined as $X=\left\{\mathbf{x} \mid \mathbf{x}_{i} \in \mathcal{X}, i=\right.$ $1, \cdots, n\}$
	\end{definition}
	\begin{definition}[Task]
		A task $\mathcal{T}$ consists of a label space $\mathcal{Y}$ and a decision function $f$, i.e., $\mathcal{T}=\{\mathcal{Y}, f\} .$ The decision function $f$ is an implicit one, which is expected to be learned from the sample data.
	\end{definition} 
	
	{\noindent Some} machine learning models actually output the predicted conditional distributions of instances. In this case, $f\left(\mathbf{x}_{j}\right)=\left\{P\left(y_{k}|\mathbf{x}_{j}\right)|y_{k} \in \mathcal{Y}, k=1, \cdots,\left| \mathcal{Y}\right|\right\}$.\\
	In practice, a domain is often observed by a number of instances with/without the label information. For example, a source domain $\mathcal{D}_{S}$ corresponding to a source task $\mathcal{T}_{S}$ is usually observed via the instance-label pairs, i.e., $D_{S}=\left\{(\mathbf{x}, y) \mid \mathbf{x}_{i} \in \mathcal{X}^{S}, y_{i} \in \mathcal{Y}^{S}, i=1, \cdots, n^{S}\right\}$ an observation of the target domain usually consists of a number of unlabeled instances and/or limited number of labeled instances.
	
	\begin{definition}[Transfer Learning]
		Given some/an observation(s) corresponding to $m^{S} \in \mathbb{N}^{+}$ source domain(s) and task(s) (i.e., $\left.\left\{\left(\mathcal{D}_{S_{i}}, \mathcal{T}_{S_{i}}\right) \mid i=1, \cdots, m^{S}\right\}\right)$, and some/an observation(s) about $m^{T} \in \mathbb{N}^{+}$ target domain(s) and task(s) (i.e., $\left.\left\{\left(\mathcal{D}_{T_{j}}, \mathcal{T}_{T_{j}}\right) \mid j=1, \cdots, m^{T}\right\}\right)$, transfer learning utilizes the knowledge implied in the source domain(s) to improve the performance of the learned decision functions $f^{T_{j}}(j=$ $\left.1, \cdots, m^{T}\right)$ on the target domain $(s)$
	\end{definition}
	
	{\noindent In} our case $m^{S}=m^{T}=1$.\\
	
	{\noindent Now} let's discuss one by one all the models of the ImageNet challenge we used in our project.
	
	\paragraph{ResNet-V2}
	
	First in our list is ResNet50-V2 and its relatives ResNet101-V2 and ResNet152-V2 \cite{resnet-v2}. Seeing this, the first question that comes to our mind is what is the fundamental difference between all these models, and secondly, how this $ 2^{nd} $ version is different from the first one? Before that, we'll discuss the basic building blocks of ResNet50, which are also called \textit{residual} or \textit{identity block} and \textit{convolutional block}.
	\begin{itemize}[leftmargin=*]
		\item \textit{Speciality of Fundamental Blocks:} Due to this underlying structure, the deeper neural network throws less error than expected. In theory, as more layers are added to a neural network, the training error should decrease monotonically. But in practice, however, for a traditional neural network, it will reach a point where the training error will start increasing. And hence, ResNet does not suffer from this problem. The training error will keep decreasing as more layers are added to the network. In a residual block, the activation of a layer is fast-forwarded to a deeper layer in the neural network, which we call a skip connection.
		\begin{figure}[H]
			\centering
			\includegraphics[width=0.8\linewidth, height=0.36\linewidth]{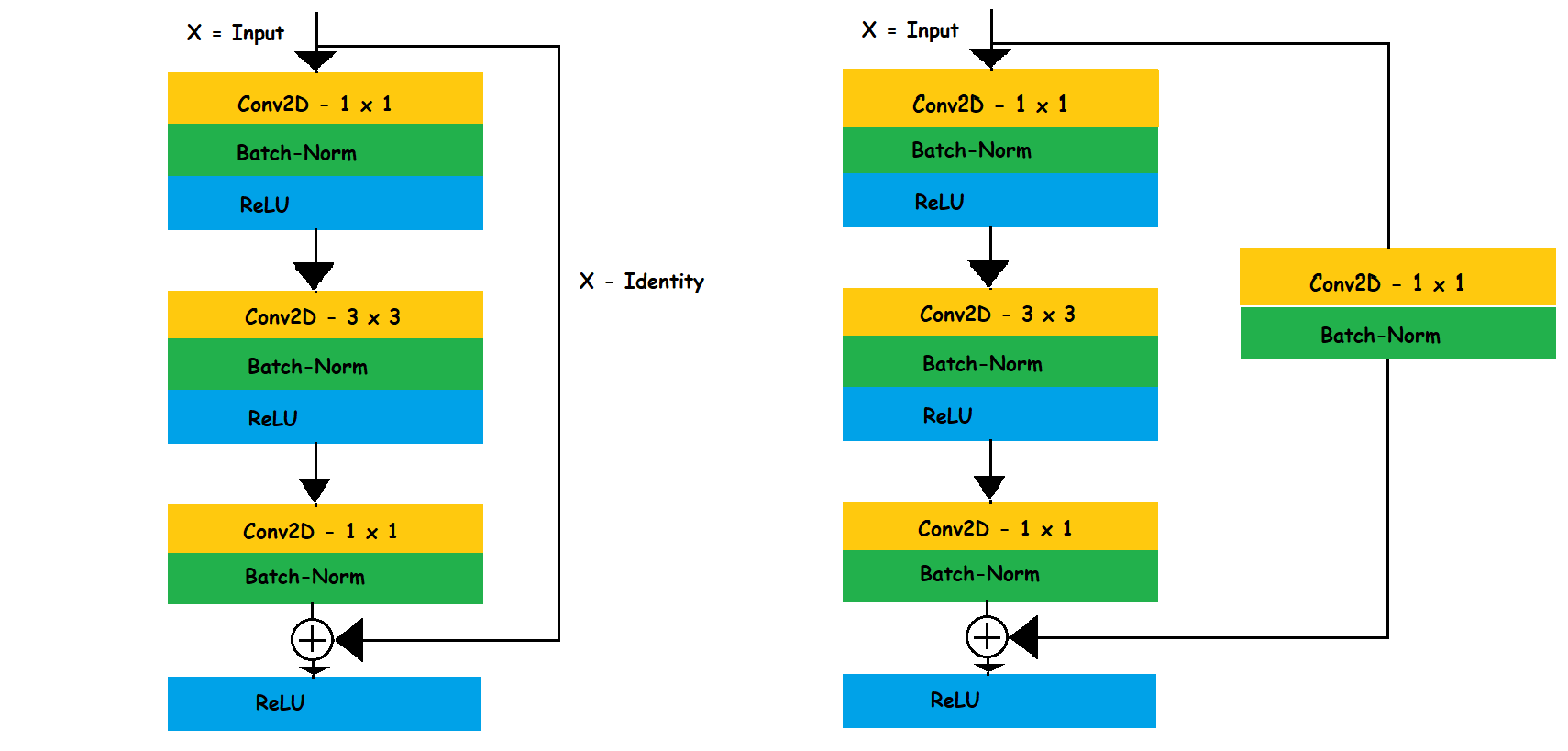}
			\caption{Left: Residual or Identity Block, Right: Convolutional block}
			\label{fig:residual_conv_block}
		\end{figure}
		\item Below is the summary of our full ResNet50 model and its other variants.
		\begin{figure}[H]
			\centering
			\copyrightbox[b]{\includegraphics[width=\linewidth]{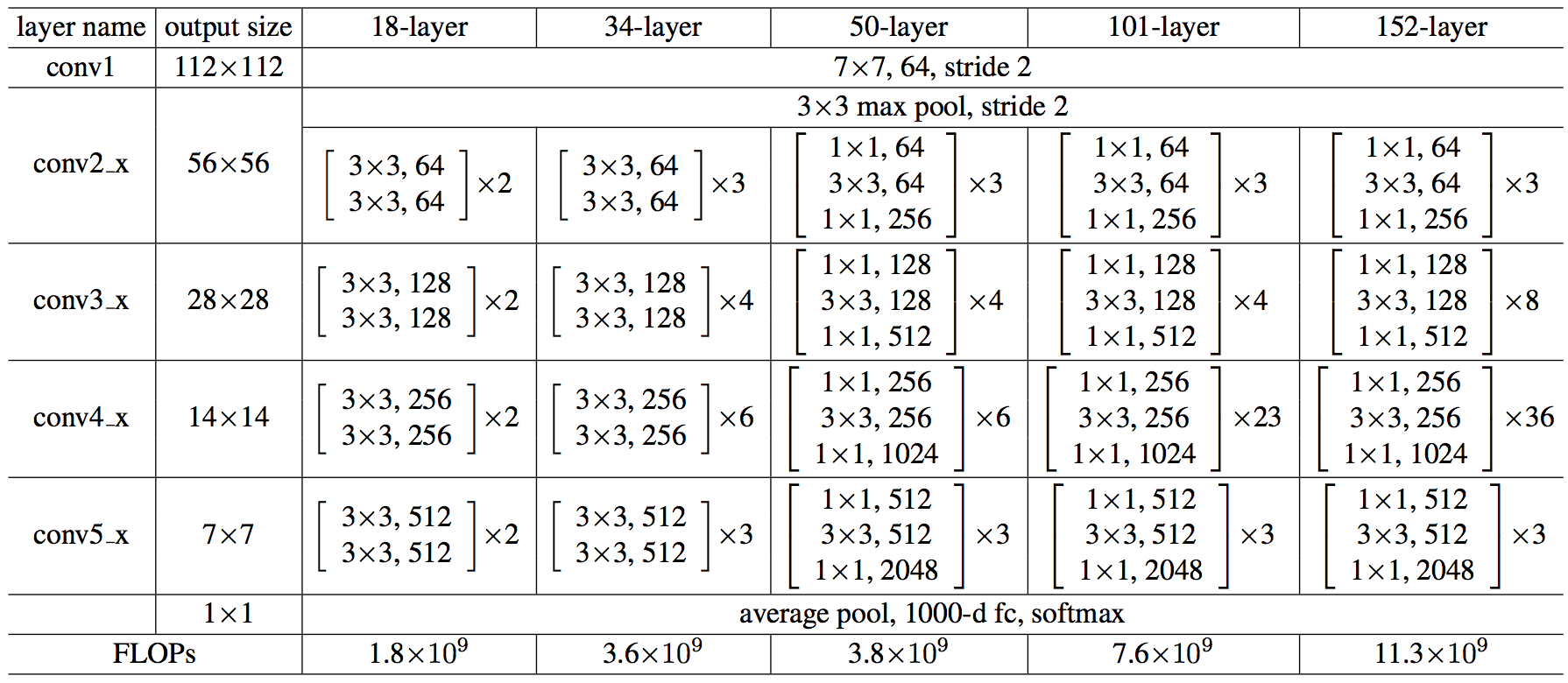}}
			{Source:\href{https://arxiv.org/pdf/1603.05027.pdf}{Identity Mappings in Deep Residual Networks}}
			\caption{Summary of Resnet50 and its variant}
			\label{fig:resnet_arch}
		\end{figure}
		\item \textit{Remarks:} Every image is pre-processed before it is being passed through the model. And this pre-processing function is uniquely associated with each ImageNet model as it ensures that the image fed to the model should have the format (i.e., range of pixel values) same as those on which the respective model is trained before. To date, we have three pre-processing functions, namely, \enquote*{\textit{tf},} \enquote*{\textit{caffe},} and \enquote*{\textit{torch}.}
		\begin{enumerate}[leftmargin=*]
			\item  \enquote{\textit{tf}} scales pixels between -1 and 1, sample-wise. Let's see the corresponding mathematical formulation for this pre-processing. Say, $ image\_arr = [x, y, z] $ is the image array in RGB format, where $ x, y, z \in [0, 255]^{2} \cap \mathbb{Z}^{2} $. Then,
			\begin{enumerate}[leftmargin=*]
				\item $ image\_arr = image\_arr/127.5 = \dfrac{image\_arr - 0}{(255 - 0)/2} $ (to scale it in between $ [0, 2] $).
				\item $ image\_arr = image\_arr - 1 $ (that shifts image array within $ [-1, 1] $).
			\end{enumerate}
			\item \enquote{\textit{caffe}} converts the images from RGB to BGR, then zero-centers each color channel with respect to the ImageNet dataset, without scaling. Again, say, $ image\_arr = [x, y, z] $  in RGB format, where $ x, y, z \in [0, 255]^{2} \cap \mathbb{Z}^{2} $. Then,
			\begin{enumerate}[leftmargin=*]
				\item $ image\_arr = [z, y, x] $ (conversion from RGB to BGR format).
				\item $ image\_arr = image\_arr - [103.939, 116.779, 123.68] $ (i.e., $ 103.939 $ is subtracted from all elements in 'B' channel, $ 116.779 $ from 'G', and $ 123.68 $ from 'R').
			\end{enumerate}
			\item \enquote{\textit{torch}} scales pixels between $ 0 $ and $ 1 $ and then normalizes each channel with respect to the ImageNet dataset. Say, $ image\_arr = [x, y, z] \in \left([0, 255]^{2} \cap \mathbb{Z}^{2}\right)^{3} $ in RGB format, then,
			\begin{enumerate}[leftmargin=*]
				\item $ image\_arr = \dfrac{image\_arr}{255.0} $
			\end{enumerate}
		\end{enumerate}

		\item ResNet50 prefers the \enquote{caffe} mode for pre-processing with a bottleneck residual block design for performance enhancement.
		\begin{table}[H]
			\centering
			\scalebox{0.8}{
				\begin{tabular}{|c|c|c|}
					\hline
					caffe & tf & torch\\
					\hline
					\hline
					ResNet50&ResNet50-V2&DenseNet121\\
					ResNet101&ResNet101-V2&DenseNet169\\
					ResNet152&ResNet152-V2&DenseNet201\\
					VGG16&Inception-V3&\\
					VGG19&Inception-ResNet-V2&\\
					&Xception&\\
					\hline
				\end{tabular}}
			\caption{Pre-processing mode for all ImageNet Models we used in our project \cite{tf-preprocess}}
			\label{table:pre-processing}
		\end{table}
		\item \textit{ResNet-V1 vs ResNet-V2}: The dissimilarity between version 2 (ResNet-V2 \cite{resnet-v2}) from the older one \cite{resnet} is due to the mismatched ordering of layers in the fundamental building block (i.e., both residual block and convolutional block). In version 2, instead of post-activation, pre-activation of weight layers is done, which is shown below.
		\begin{figure}[H]
			\centering
			\includegraphics[width=0.52\linewidth, height=0.45\linewidth]{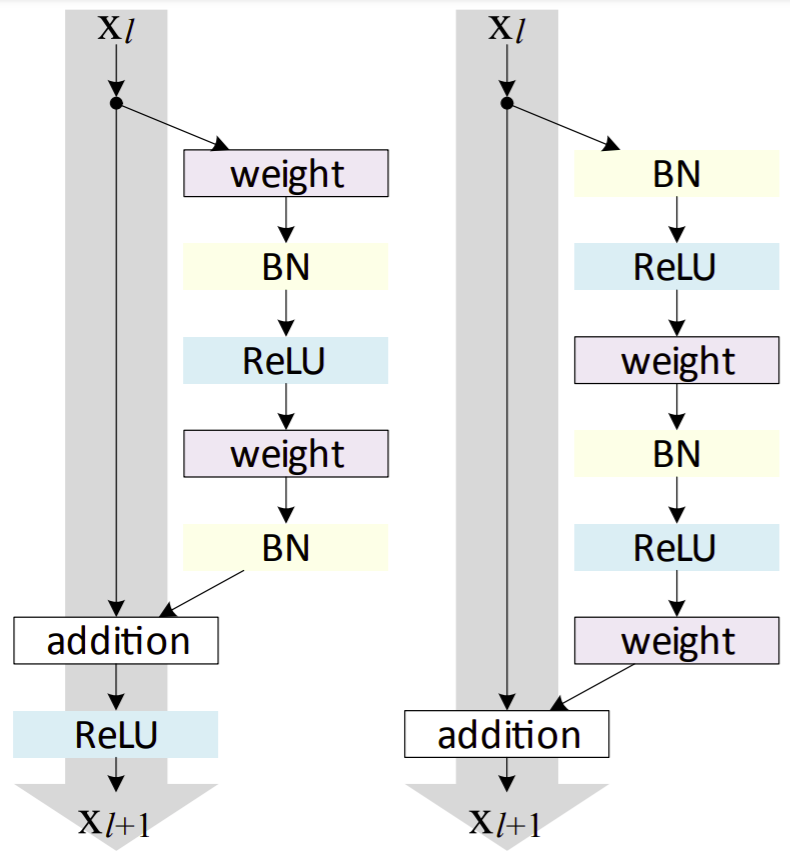}
			\caption{Left: ResNet-V1; Right: ResNet-V2 \cite{resnet-v2}}
			\label{fig:resv1_v2}
		\end{figure}
		\item Significant differences between these two versions are listed as follows:
		\begin{enumerate}[leftmargin=*]
			\item The first version adds the second non-linearity (i.e., the activation function) after the addition operation is performed between $x_{l}$ and $F(x_{l})$ and then transfers it to the next block as the input. At the same time, the second version cleared the path of $x_{l}$ to $x_{l+1}$ in the form of identity connection by removing the last non-linearity. And this ensures the output of add operation is passed as it is to the next block.
			\item ResNet-V2 performs convolution on the input only after BN and ReLU are applied on it, while in the first version, BN and ReLU are applied after convolution (i.e., weight matrix multiplication)
			\item Pre-processing mode for ResNet-V2 is \enquote{tf,} while for V1, it is \enquote{caffe.}
		\end{enumerate}
		\item Similarly, ResNet101-V2 and ResNet152-V2 differ from ResNet50-V2 in terms of the number of total convolutional filters used. For ResNet101-V2, it is 101, while for ResNet152-V2, it is 152.
	\end{itemize}
	\begin{figure}[H]
		\centering
		\includegraphics[width=\linewidth]{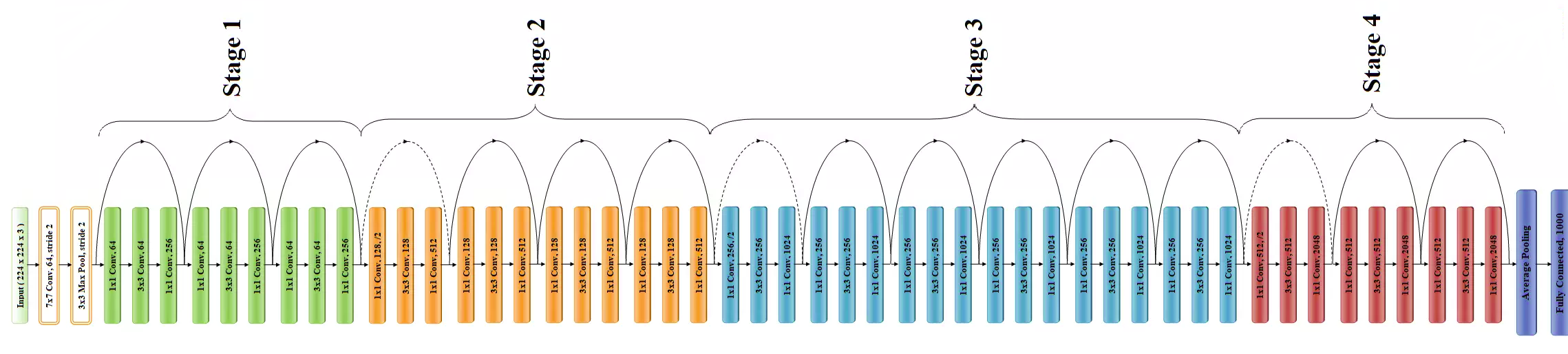}
		\caption{Full Architecture of ResNet-50}
		\label{fig:resnet50}
	\end{figure}
	\paragraph{VGG}
	
	Second, on our list are VGGG16 and VGGG19 \cite{vgg}, where VGG stands for \textit{Visual Geometry Group} of University of Oxford. As in ResNet50-V2, here also we will exhibit some fundamental blocks for VGG16 and VGGG19, called VGG blocks. This network is one of the derivatives of AlexNet \cite{alexnet} that addresses a crucial aspect of CNN's -- depth. AlexNet, on its first appearance, won the ILSVRC-2012 challenge and proved itself to be one of the most capable object-detection and classification algorithms. $ ReLU $ instead of $tanh$, overlapping pooling and optimization with multiple GPU are some of the critical features of AlexNet.\\
	
	{\noindent Now the VGG, while based on AlexNet, has several contrasting features:}
	\begin{enumerate}[leftmargin=*]
		\item VGG uses small receptive fields ($ 3\times3 $ with a stride of 1) instead of large receptive fields like Alexnet ($ 11\times11 $ with stride $ = 4 $). Instead of one, $3 ReLU $ units make the decision function more discriminative.
		\item With $ 1\times1 $ convolutional layers in VGG, the decision function become more non-linear without incorporating any change in receptive fields.
		\item Due to small-sized convolution filters, VGG has many weight layers, which again leads to improved performance.
	\end{enumerate}
	VGG, like ResNet, is made up of some basic repeating blocks.
	\begin{figure}[H]
		\centering
		\begin{subfigure}[b]{0.49\textwidth}
			\centering
			\includegraphics[width=0.64\textwidth, height=0.57\linewidth]{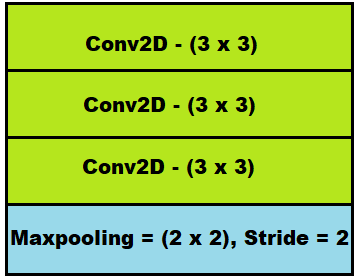}
			\caption{VGG16 Block}
			\label{fig:vgg16_block}
		\end{subfigure}
		\hfill
		\begin{subfigure}[b]{0.49\textwidth}
			\centering
			\includegraphics[width=0.64\textwidth, height=0.57\linewidth]{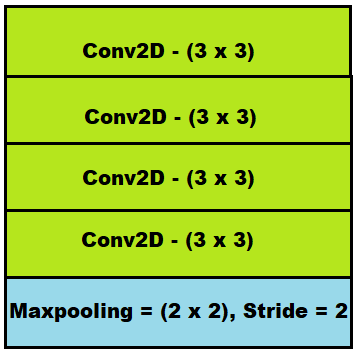}
			\caption{VGG19 Block}
			\label{fig:vgg19_block}
		\end{subfigure}
		\caption{VGG Fundamental Blocks}
		\label{fig:vgg_blocks}
	\end{figure}
	\begin{figure}[H]
		\centering
		\includegraphics[width=\linewidth, height=4.2cm]{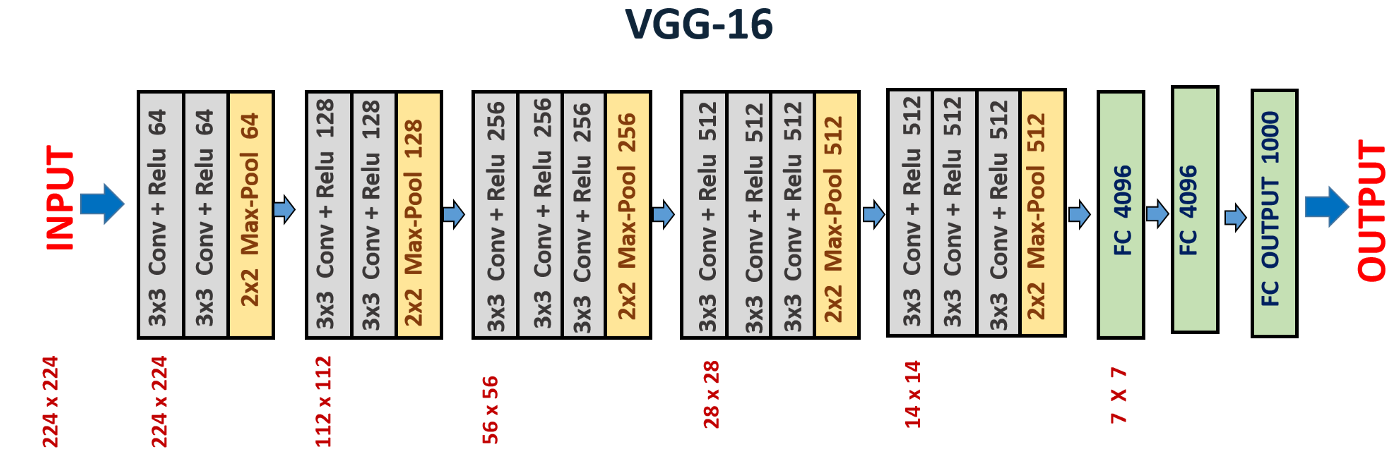}
		\caption{VGG16 Full Architecture}
		\label{fig:vgg16_full}
	\end{figure}
	\begin{figure}[H]
		\centering
		\copyrightbox[b]{\includegraphics[width=0.85\linewidth]{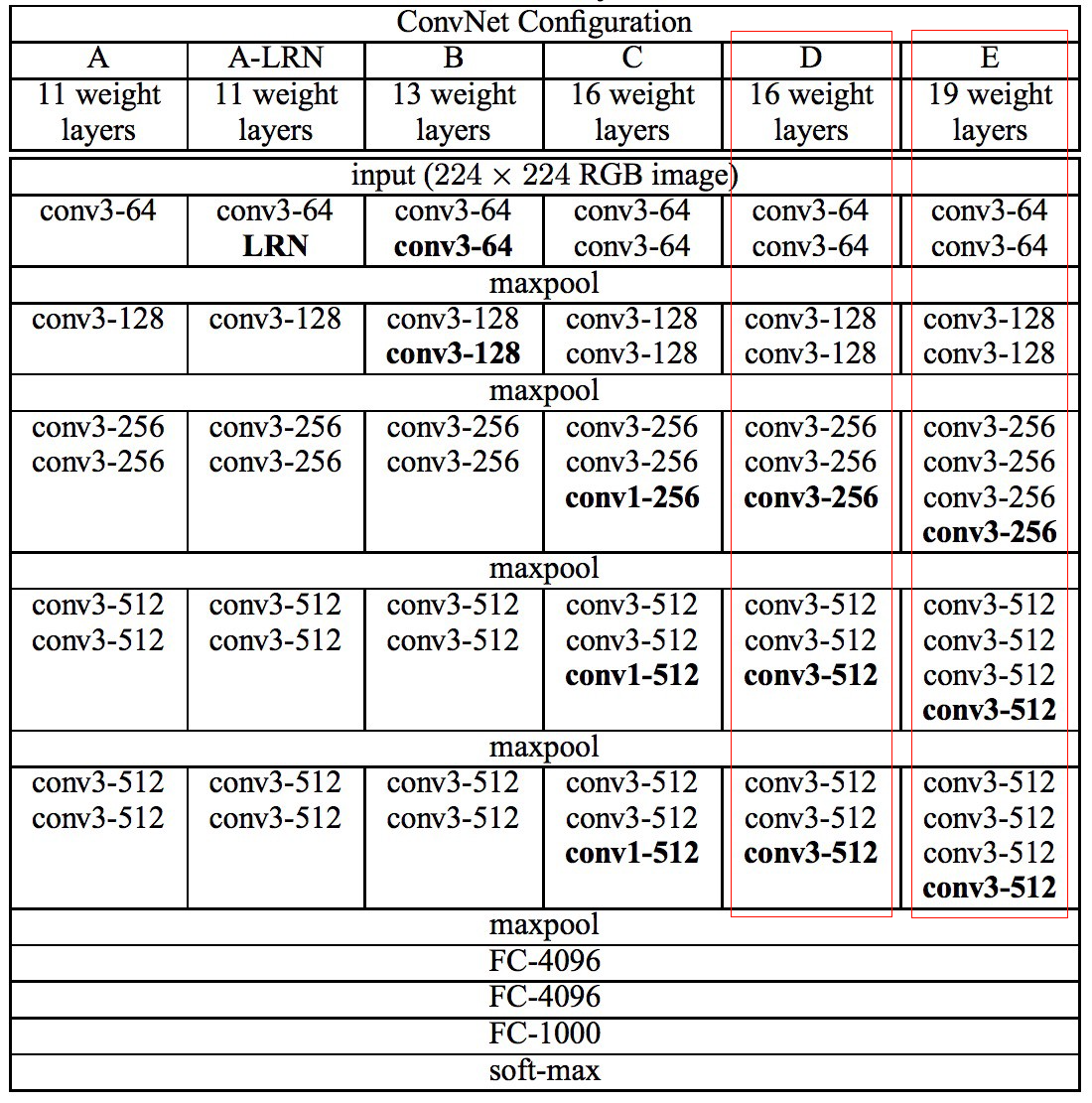}}
		{Source: \href{https://arxiv.org/pdf/1409.1556v6.pdf}{V\lowercase{ERY DEEP CONVOLUTIONAL NETWORKS FOR LARGE-SCALE IMAGE RECOGNITION}}}
		\caption{VGG16 vs. VGG19}
		\label{fig:vgg16vs19}
	\end{figure}
	\textit{VGG16 vs. VGGG19}: \enquote{$ 16 $} and \enquote{$ 19 $} stand for the number of weight layers in the network (columns D and E in figure \ref{fig:vgg16vs19}). VGG19 has three more \textit{conv3} layers.
	\newpage
	\paragraph{InceptionV3}
	
	This deep CNN architecture originated from GoogleNet (or Inception-V1) \cite{inception-v1}, developed by Google, which set the state-of-the-art for classification and detection in the ILSVRC-2014 challenge. The prime specialty of Inception-V1 is the improved utilization of the computing resources inside the network. One great feature of its architecture is to address the extreme variation of image sizes as well as the location of an object of concern in the images. For instance, the image shown below speaks about the difference in the area occupied by each dog.
	\begin{figure}[H]
		\centering
		\copyrightbox[b]{\includegraphics[width=\linewidth]{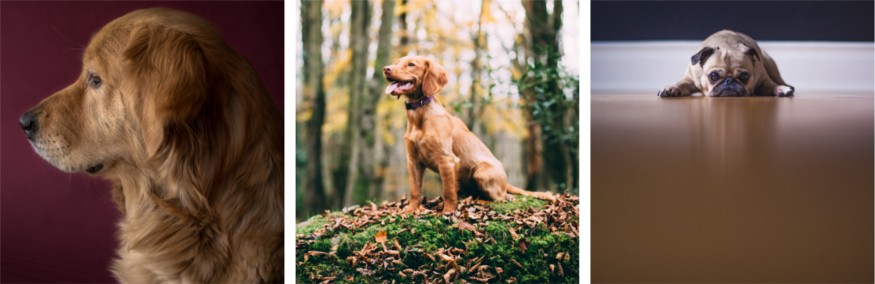}}
		{Source: \href{https://unsplash.com/}{Unsplash}}
		\caption{From right: A dog occupying very little space, a dog occupying a part of it, and a dog occupying most of the image.}
		\label{fig:dog}
	\end{figure}
	{\noindent And due to this, choosing the right kernel size for the convolution operation becomes resilient. For information that is distributed more globally, a larger kernel is preferred.} Plus, we're very aware that profound networks are prone to overfitting, and it's heavy to update the gradients through the entire network, which may lead to a vanishing gradient issue. And finally, stacking deep convolutional operation is computationally expensive with large time complexity.\\
	
	{\noindent\textit{Solution:}} And to mitigate all the above headaches, the article \cite{inception-v1} introduces an architecture that has filters with multiple sizes operating on the same level, which makes the network a bit \enquote{wider} rather than \enquote{deeper.}
	\begin{figure}[H]
		\centering
		\begin{subfigure}[b]{0.49\textwidth}
				\centering
				\includegraphics[width=\textwidth, height=4.2cm]{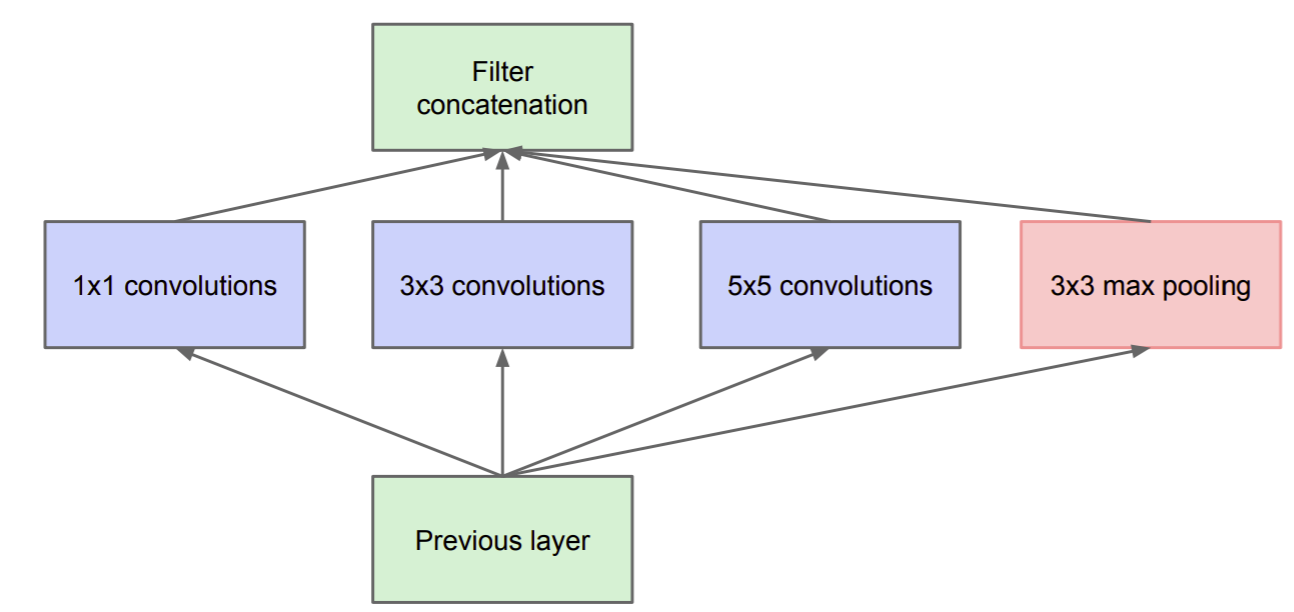}
				\caption{Inception module, naive version}
				\label{fig:inception_naive}
			\end{subfigure}
			\hfill
			\begin{subfigure}[b]{0.49\textwidth}
				\centering
				\includegraphics[width=\textwidth, height=4.2cm]{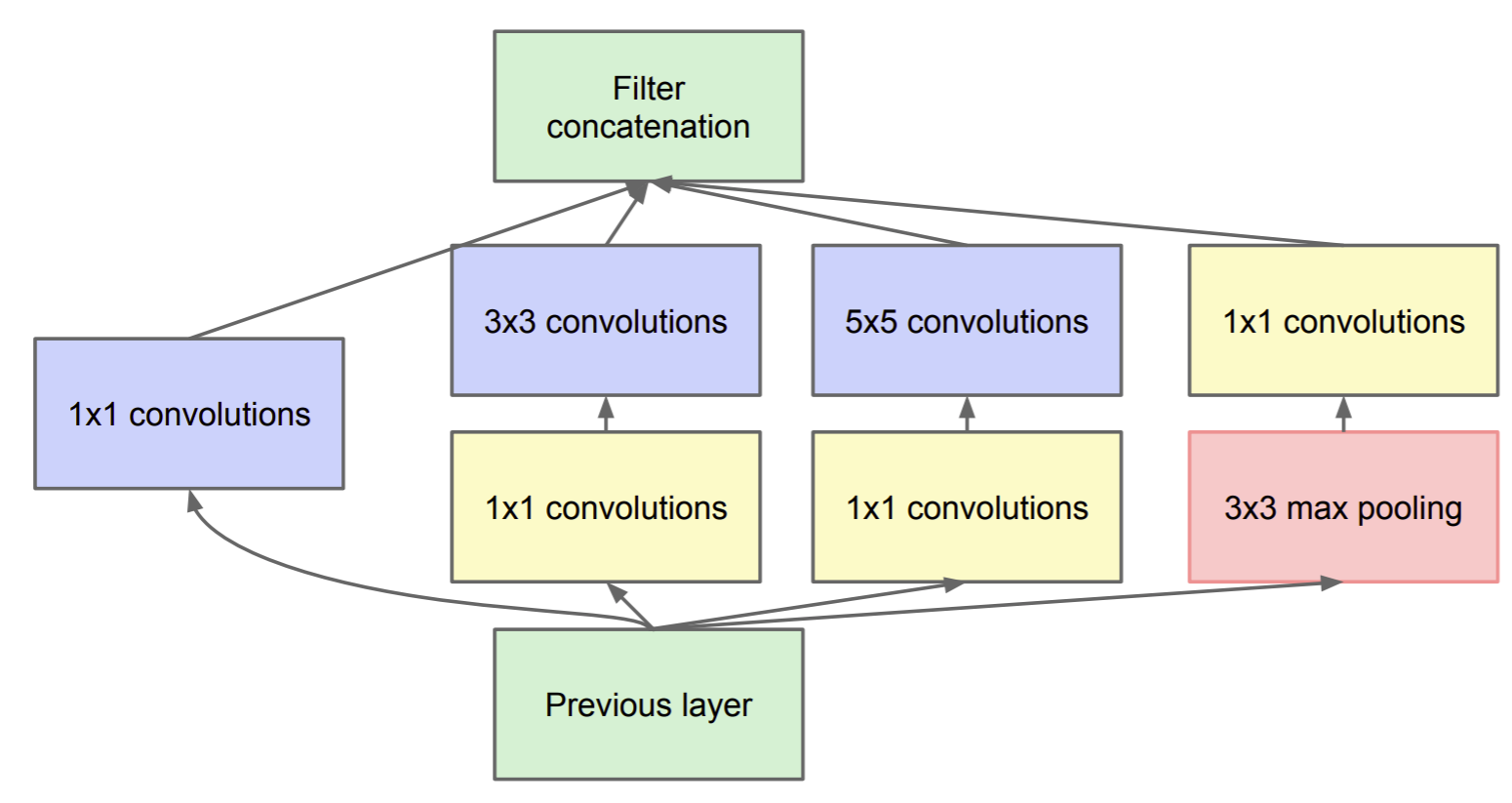}
				\caption{ Inception module with dimension reductions}
				\label{fig:inception_reduction}
		\end{subfigure}
		\caption{Inception module \cite{inception-v1}}
		\label{fig:inception_module}
	\end{figure}
	To alleviate the computational cost incurred by such module as in figure \ref{fig:inception_naive}, the authors used an extra $ 1\times1 $ convolution before $ 3\times3 $ and $ 5\times5 $ convolutions to limit the number of input channels as shown in figure \ref{fig:inception_reduction}.\\
	
\begin{figure}[H]
		\centering
		\begin{subfigure}[b]{0.48\textwidth}
			\centering
			\includegraphics[width=\textwidth]{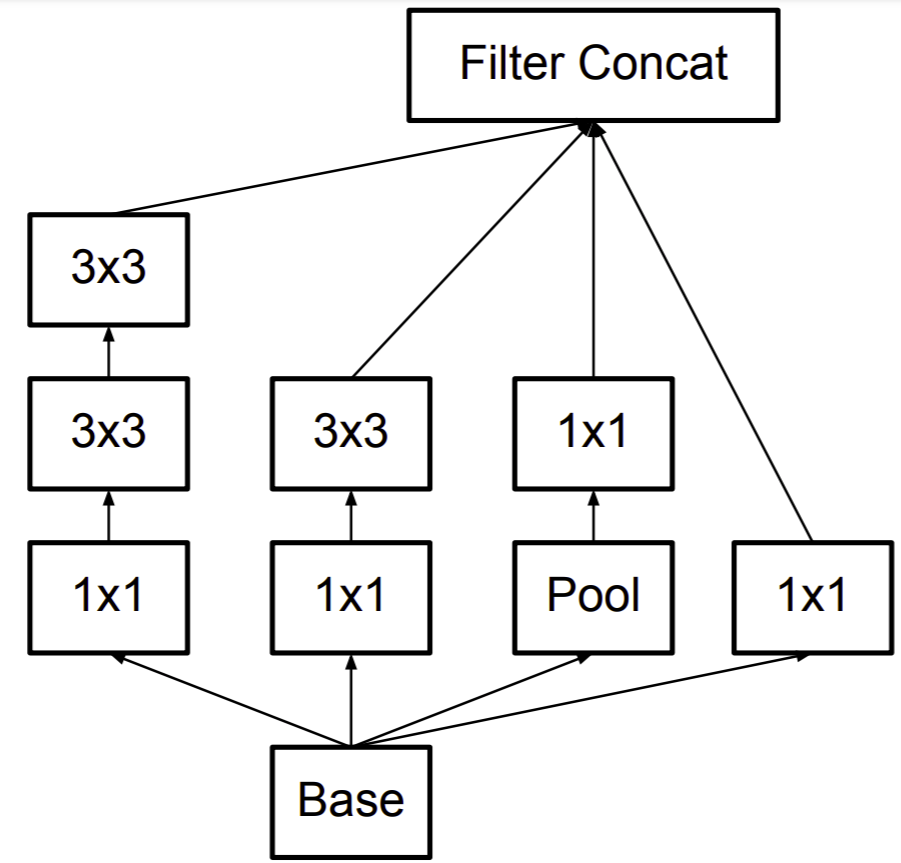}
			\caption{Inception modules where each $ 5\times5 $ convolution is replaced by two $ 3 \times 3 $ convolution,}
			\label{fig:5*5-3*3}
		\end{subfigure}
		\hfill
		\begin{subfigure}[b]{0.48\textwidth}
			\centering
			\includegraphics[width=\textwidth]{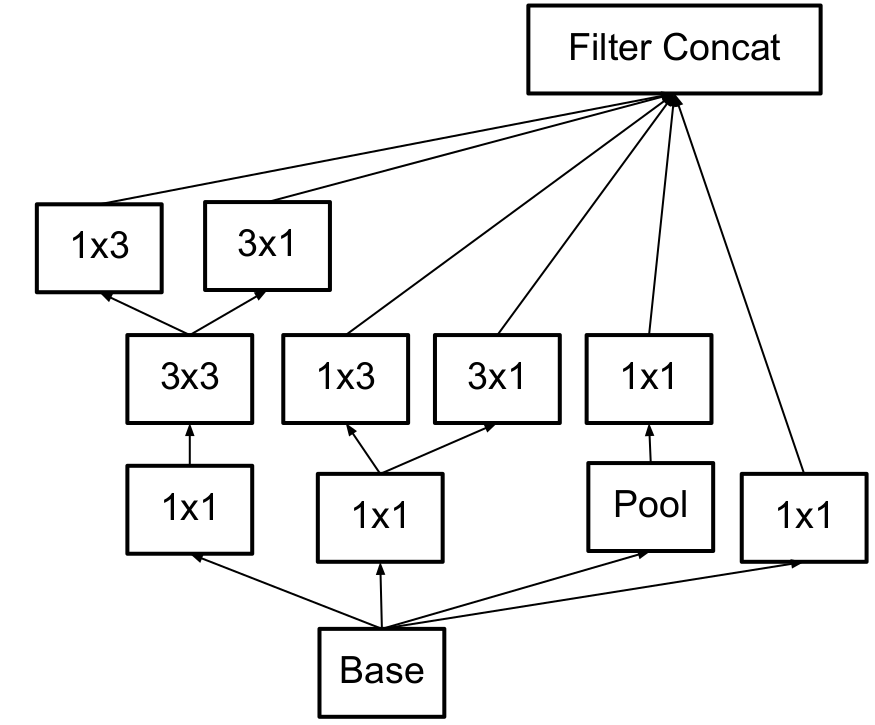}
			\caption{Inception modules with expanded the filter bank outputs.}
			\label{fig:3*3-1*3-3*1}
		\end{subfigure}
		\caption{\centering Modified Inception Module \cite{inception-v3}}
		\label{fig:modified_inception}
	\end{figure}

	{\noindent With Inception-V2, the following shortcomings of Inception-V1 were addressed.}
	\begin{enumerate}[leftmargin=*]
		\item Reducing the representational bottleneck didn't alter the dimensional of the input drastically, which otherwise may have caused huge information loss.
		\item Using ingenious factorization methods such as spatial separable convolution, convolutions can be made computationally efficient.
	\end{enumerate}
	{\noindent\textit{Solution:}} By factorizing $ 5\times5 $ convolution to two $ 3\times3 $ convolution operations (as in \ref{fig:5*5-3*3}), our performance improves by $ 2.78 $ times. Now representing those $ 3\times3 $ convolution, in terms $ 1\times3 $ and $ 3\times1 $ convolution (as in \ref{fig:3*3-1*3-3*1}) further made our process 33\% cheaper.
	{\noindent Next, the filter banks in the module were expanded (made wider instead of deeper) to remove the representational bottleneck.}\\
	
	{\noindent Now}, the Inception-V3 \cite{inception-v3} incorporated all the above upgrades in version 2 and, in addition, used the following:
	\begin{enumerate}[leftmargin=*]
		\item RMSProp optimizer
		\item Factorized $ 7\times7 $ convolutions
		\item Batch Norm in the auxiliary classifiers to function then as regularizers.
		\item Label smoothing (a regularizing component) is added to the loss formula to prevent the network from overfitting.
	\end{enumerate}
	\begin{figure}[H]
		\centering
		\copyrightbox[b]{\includegraphics[width=\linewidth]{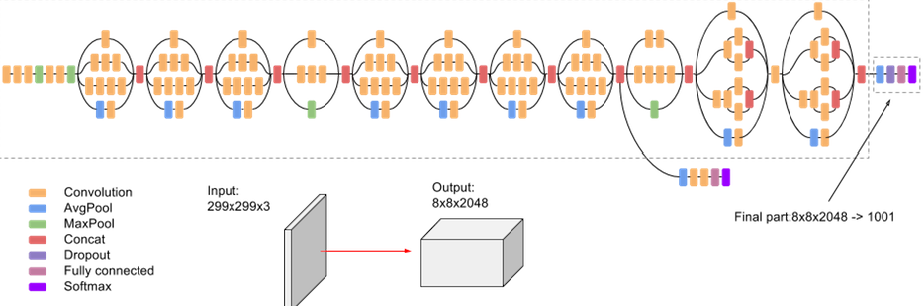}}
		{Source: \href{https://cloud.google.com/tpu/docs/inception-v3-advanced}{Inception v3}}
		\caption{Inception-V3 Full Architecture}
		\label{fig:inceptionv3_full}
	\end{figure}
	\paragraph{Inception-ResNet-V2}
	
	This architecture from Google Inc. is a hybrid of InceptionNet and ResidualNet where the inception architecture is combined with residual connections. As proved in the article \cite{inception-resnet-v2}, the residual connections indeed accelerate the training of Inception networks significantly. And even this architecture outperforms its relatives (expensive Inception networks without residual connections) by a thin margin. Further in ILSVRC-2012 image classification challenge, these variations have improved the single-frame recognition performance. Unlike ResNet and InceptionNet, this model has several fundamental building blocks, which we list as follows.
	\begin{figure}[H]
		\centering
		\begin{subfigure}[b]{0.49\textwidth}
			\centering
			\includegraphics[width=\textwidth]{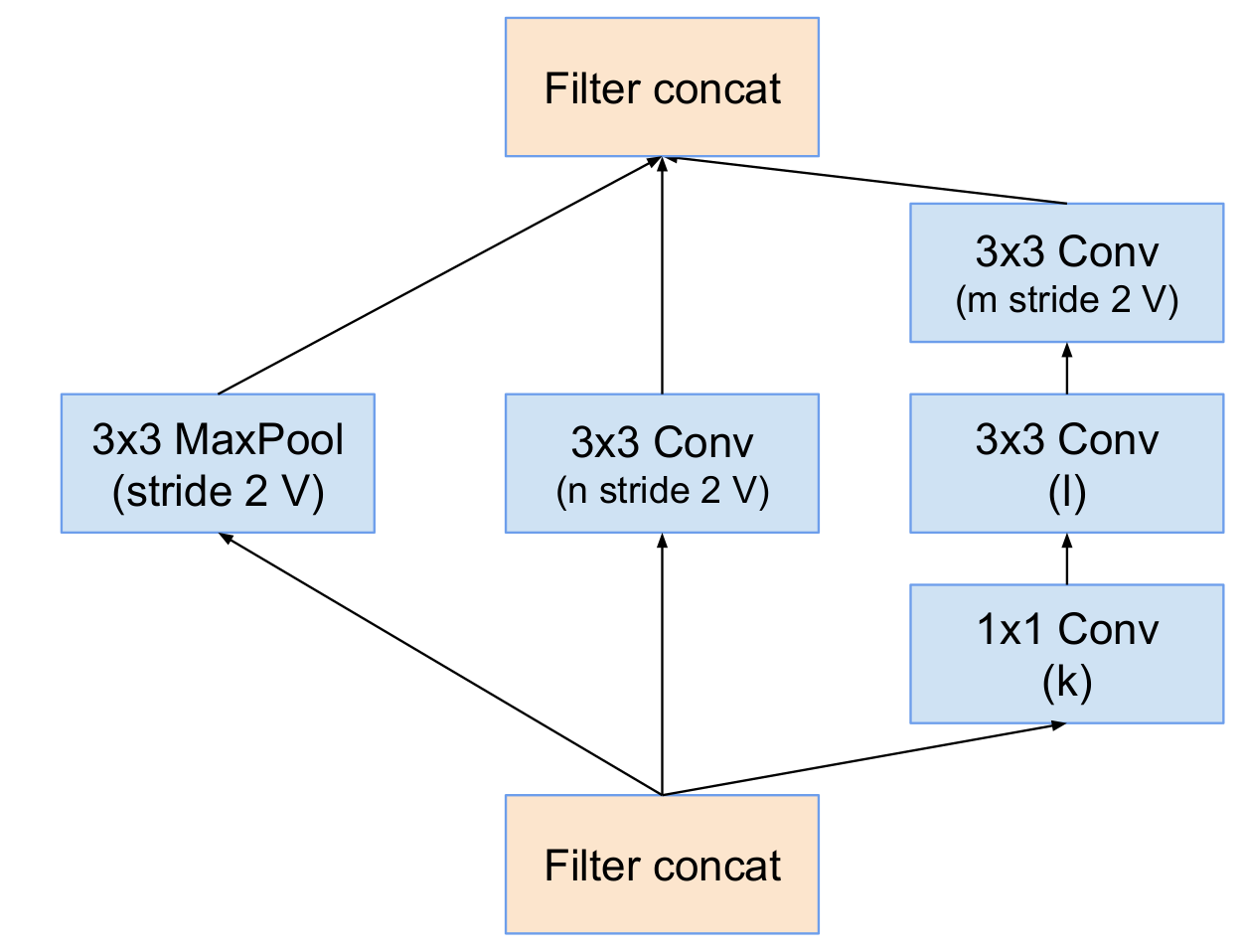}
			\caption{Reduction-A: The schema for $ 35 \times 35 $ to $ 17 \times 17 $ reduction module}
			\label{fig:red_A}
		\end{subfigure}
		\hfill
		\begin{subfigure}[b]{0.49\textwidth}
			\centering
			\includegraphics[width=0.93\textwidth, height=0.9\linewidth]{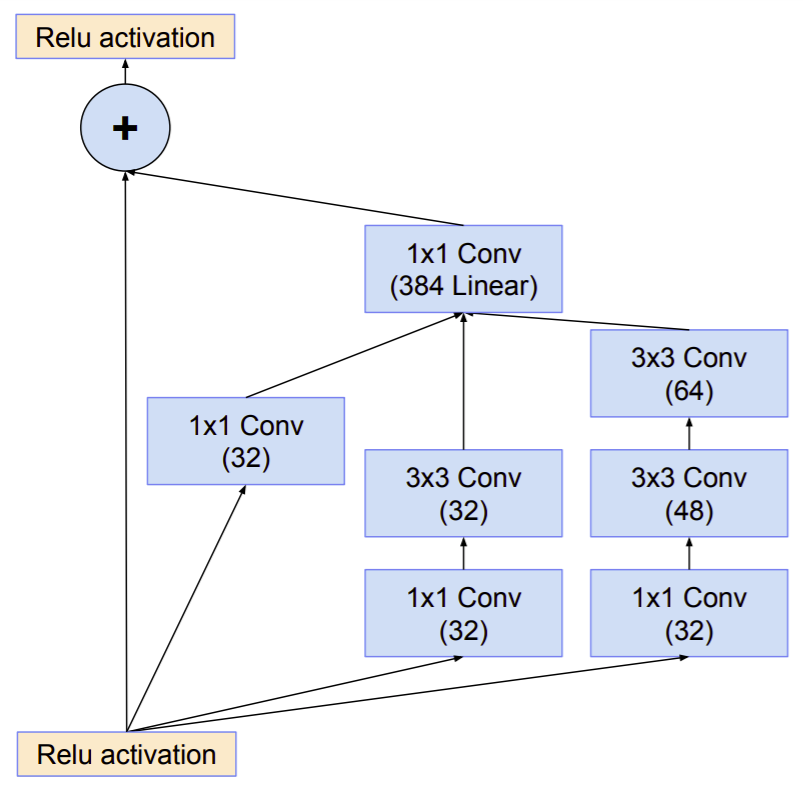}
			\caption{Inception-resnet-A: The schema for $ 35 \times 35 $ grid module of the Inception-ResNet-v2 network}
			\label{fig:inc_res_A}
		\end{subfigure}
		\caption{Reduction-A and Inception-resnet-A module}
		\label{fig:red_A-inc_res_A}
	\end{figure}
	\begin{figure}[H]
		\centering
		\begin{subfigure}[b]{0.48\textwidth}
			\centering
			\includegraphics[width=\textwidth]{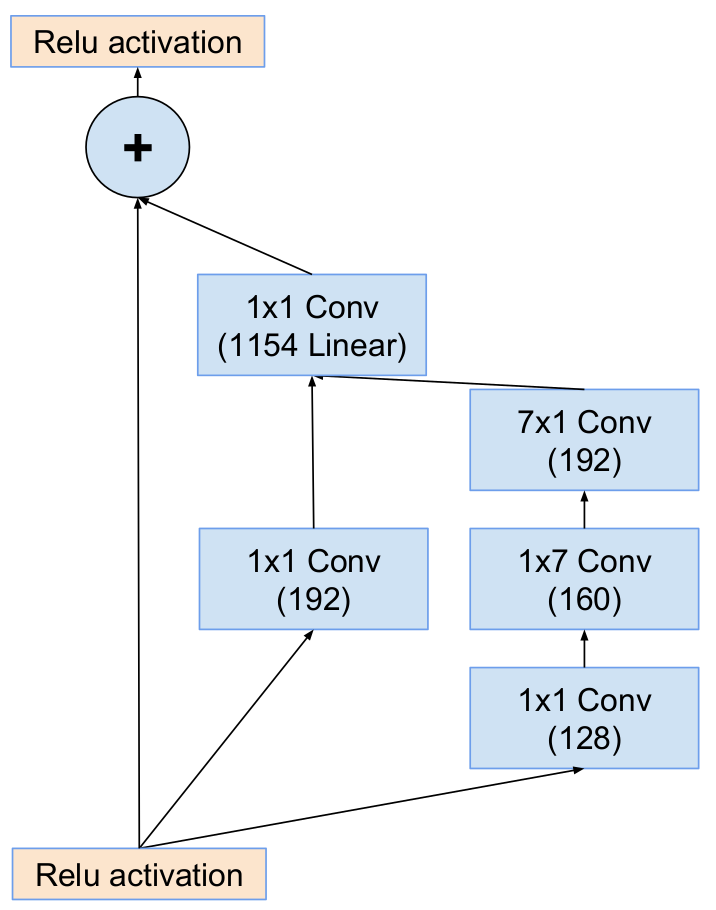}
			\caption{Inception-resnet-B: The schema for $ 17 \times 17 $ grid module of the Inception-ResNet-v2 network}
			\label{fig:inc_res_B}
		\end{subfigure}
		\hfill
		\begin{subfigure}[b]{0.48\textwidth}
			\centering
			\includegraphics[width=\textwidth]{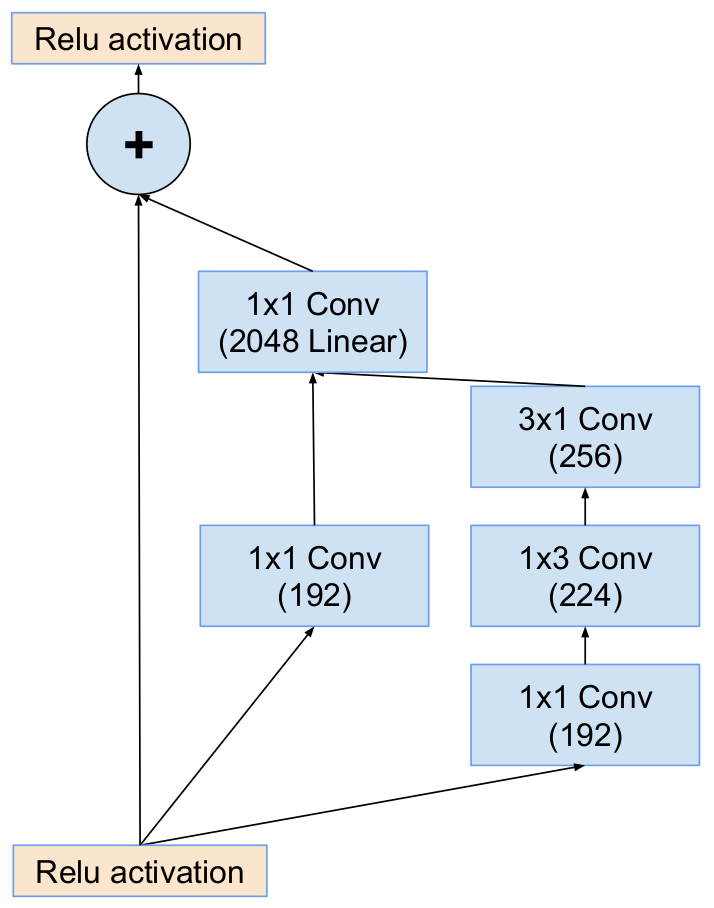}
			\caption{Inception-resnet-C: The schema for $ 8 \times 8 $ grid module of the Inception-ResNet-v2 network}
			\label{fig:inc_res_C}
		\end{subfigure}
		\medskip
		\begin{subfigure}[b]{\textwidth}
			\centering
			\includegraphics[width=\textwidth]{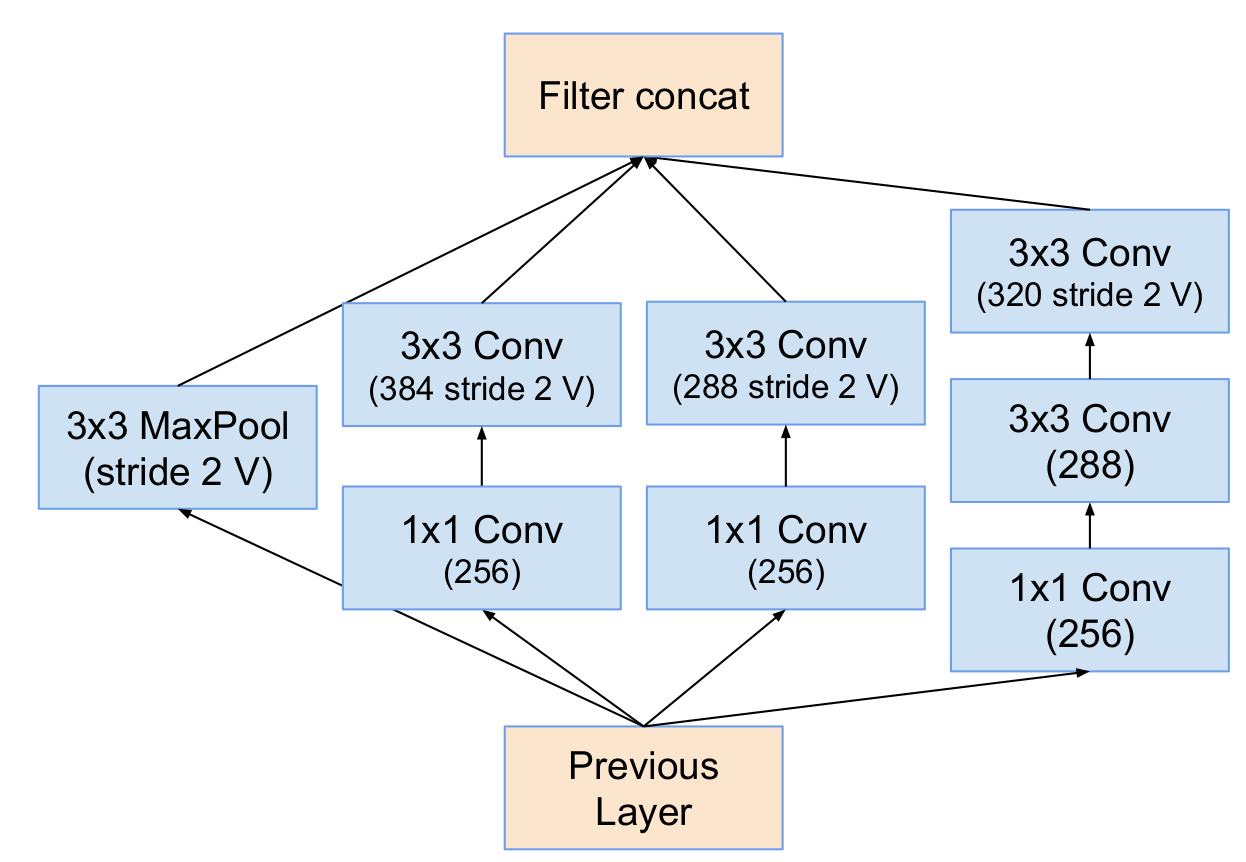}
			\caption{Reduction-B: The schema for $ 17 \times 17 $ to $ 8 \times 8 $ grid-reduction module}
			\label{fig:red_B}
		\end{subfigure}
		\caption{Inception-resnet-B, Inception-resnet-C, and Reduction-B modules}
		\label{fig:red_B-inc_res_B_C}
	\end{figure}
	\begin{figure}[H]
		\centering
		\includegraphics[width=\textwidth]{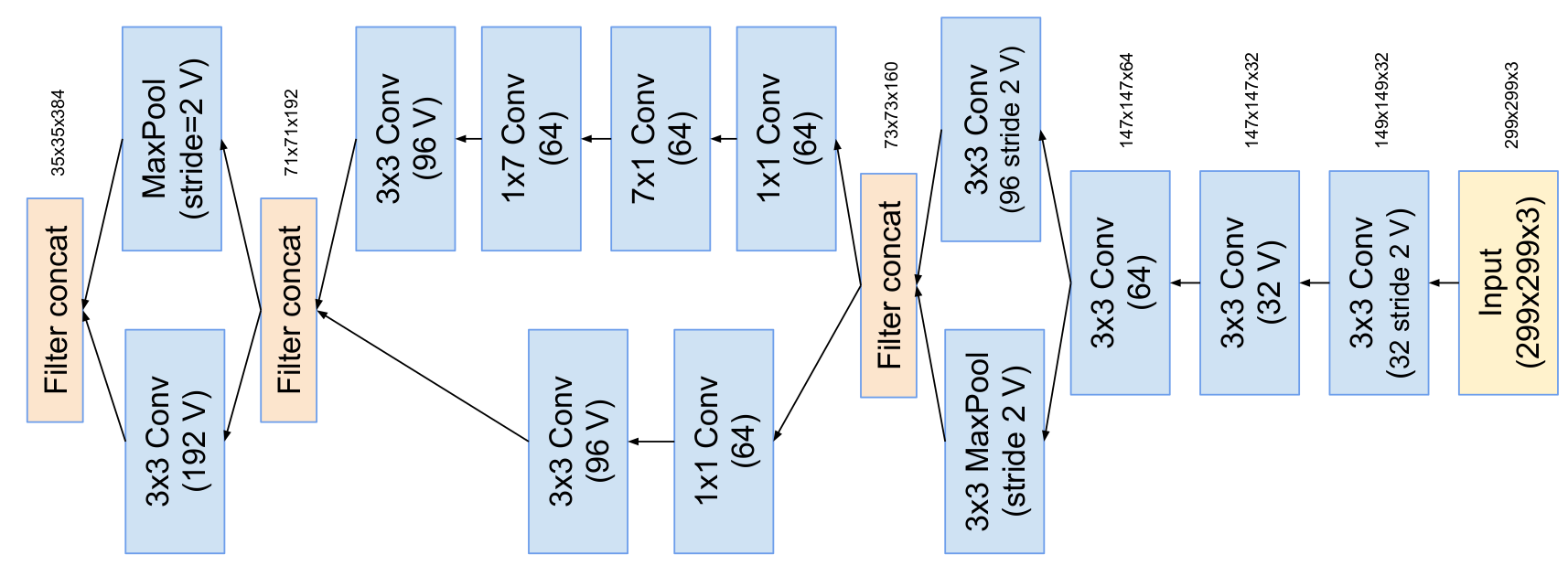}
		\caption{Stem: The input part of Inception-ResNet-V2}
		\label{fig:stem_inc_res_v2}
	\end{figure}
	{\noindent And the final model of Inception-ResNetV2 looks as follows:}
	\begin{figure}[H]
		\centering
		\includegraphics[width=\linewidth]{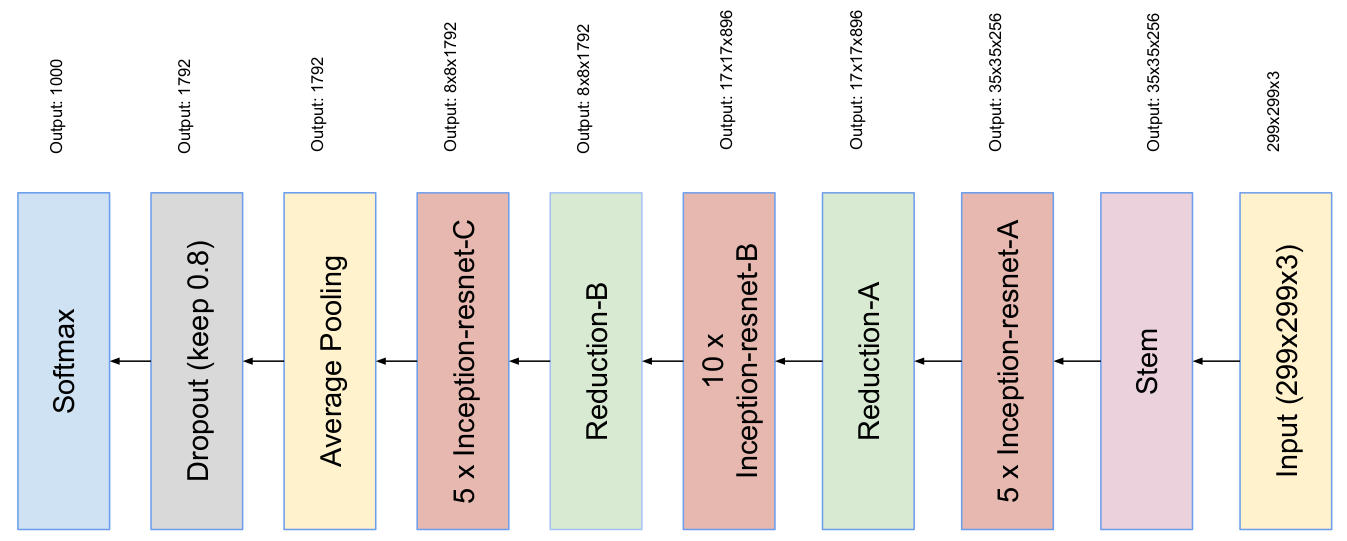}
		\caption{Incpetion-ResNet-V2 full architecture}
		\label{fig:inc_res_full}
	\end{figure}
	\paragraph{DenseNet}
	
	Like ResNet, DenseNet \cite{densenet} also addresses the vanishing gradient problem by simplifying the connectivity pattern between layers of other architectures such as:
	\begin{itemize}[leftmargin=*]
		\item High Way Networks
		\item Residual Networks
		\item Fractal Networks
	\end{itemize}
	These networks are way deeper than the ResNets but are very efficient to train as they use shorter connections between layers that strengthen the feature propagation. But unlike ResNet50, DenseNet concatenates the output feature maps of the layer with the incoming feature maps rather than summing it up. To ensure maximum information (and gradient) flow, for each layer, the feature maps of all preceding layers are used as inputs, and its feature maps are used as inputs to all subsequent layers. Unlike traditional convolutional networks with $ L $ layers having $ L $ connections -- one between each layer and its next layer -- this network has $ \frac{L(L+1)}{2} $ direct connections. And due to this, it requires fewer parameters than traditional, as there is no need to learn redundant feature maps. Another most important advantage of DenseNets is that instead of drawing representational power from extremely deep comprehensive architectures, it exploits the network's full potential through feature reuse.\\
	
\begin{figure}[H]
		\centering
		\begin{subfigure}[b]{0.49\textwidth}
			\centering
			\includegraphics[width=0.9\textwidth, height=4in]{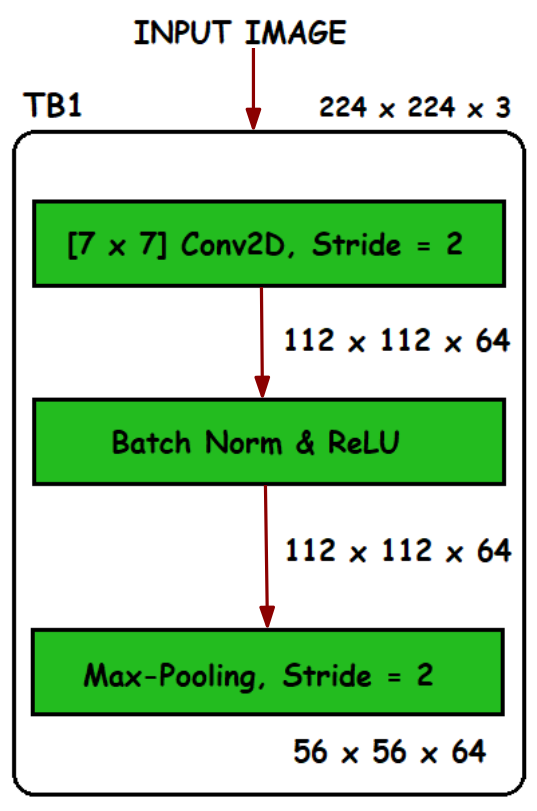}
			\caption{TB1: Transfer-Block 1 (Input block)}
			\label{fig:tb1}
		\end{subfigure}
		\hfill
		\begin{subfigure}[b]{0.49\textwidth}
			\centering
			\includegraphics[width=0.8\textwidth, height=4in]{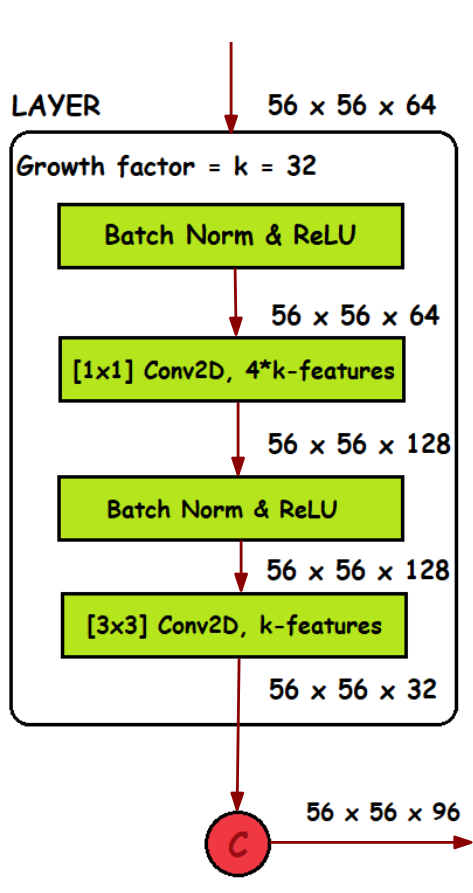}
			\caption{LAYER: Layer in Dense Block 1}
			\label{fig:layer_dense}
		\end{subfigure}
		\caption{Input Block (TB1) and Layer block (basic block to be used in Dense Block)}
		\label{fig:tb1_layer}
	\end{figure}

\begin{figure}[H]
		\centering
		\begin{subfigure}[b]{0.48\textwidth}
			\centering
			\includegraphics[width=\textwidth]{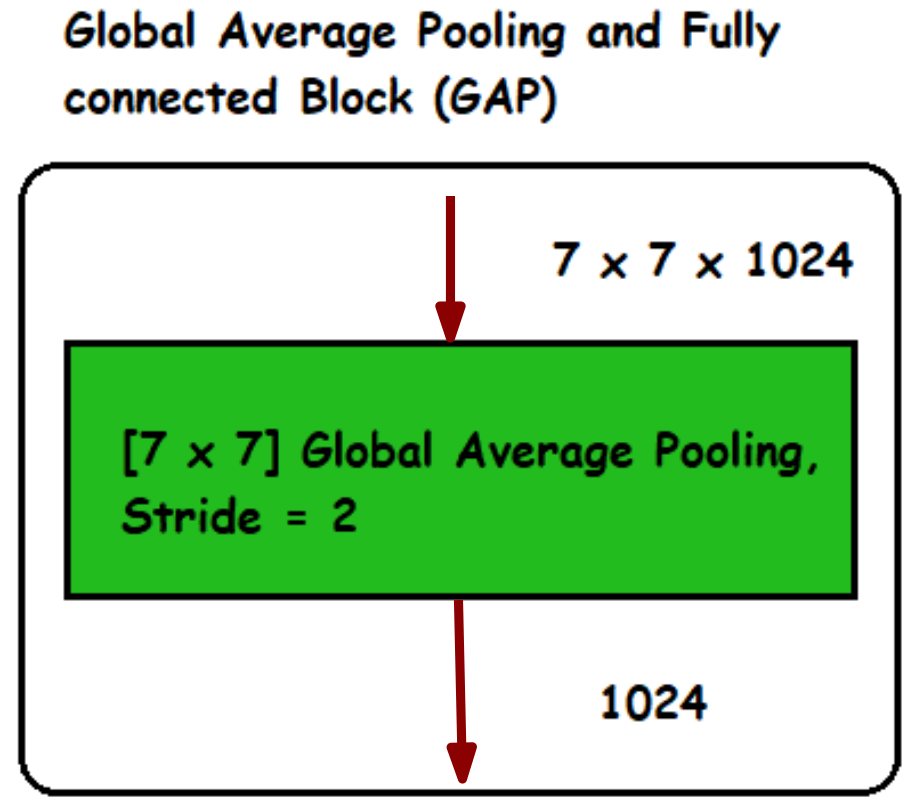}
			\caption{GAP: Last Global Average Pooling Block}
			\label{fig:gap_block_dense}
		\end{subfigure}
		\hfill
		\begin{subfigure}[b]{0.48\textwidth}
			\centering
			\includegraphics[width=0.8\textwidth, height=3.4in]{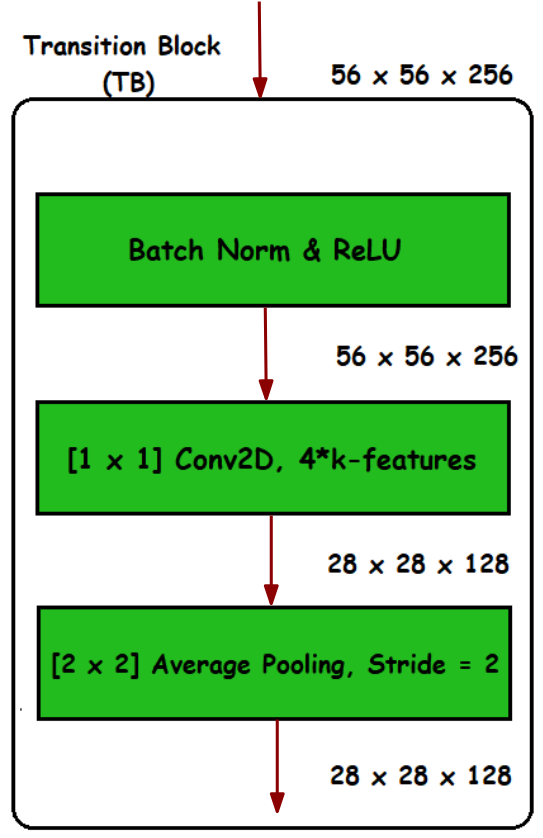}
			\caption{TB: Transfer Block}
			\label{fig:tb_dense}
		\end{subfigure}
		\medskip
		\begin{subfigure}[b]{\textwidth}
			\centering
			\includegraphics[width=\textwidth, height=1.8in]{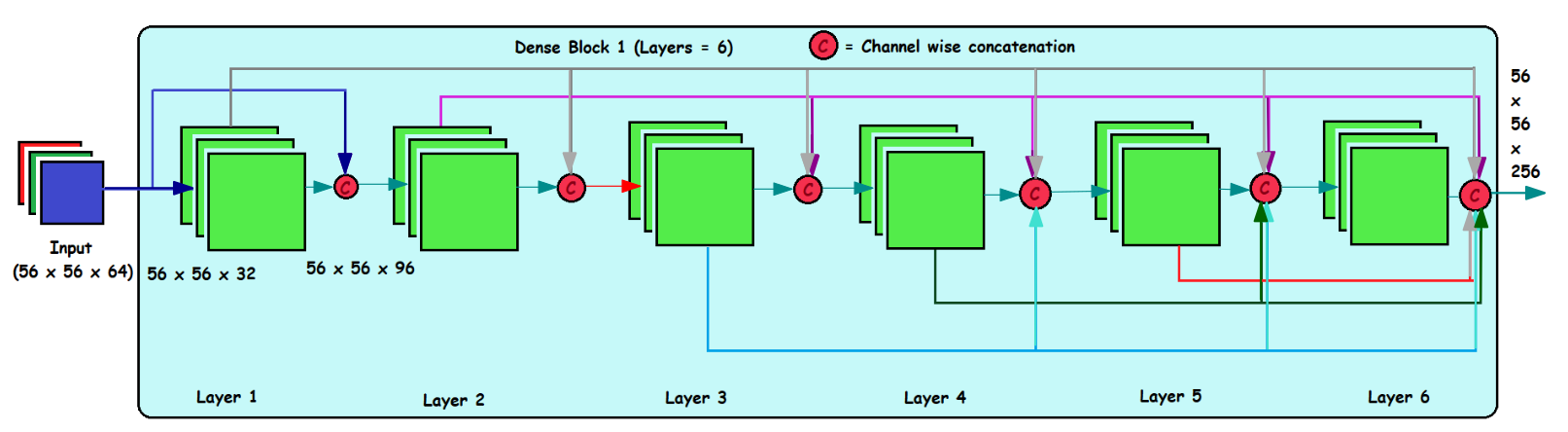}
			\caption{Dense Block 1}
			\label{fig:dense_block_1}
		\end{subfigure}
		\caption{Essential building blocks for DenseNet-121}
		\label{fig:blocks_densenet}
	\end{figure}

	{\noindent DenseNet is made up of two fundamental blocks:}
	\begin{enumerate}[leftmargin=*]
		\item Dense Block (shown in figure \ref{fig:dense_block_1})
		\item Transition Block (shown in figure \ref{fig:tb_dense} and \ref{fig:tb1})
	\end{enumerate}
	{\noindent\textit{Some Important Terms in DenseNet Architecture:}}
	\begin{itemize}[leftmargin=*]
		\item \textit{Growth rate ($ k $)}: This determines the no. of feature maps output by individual layers inside a dense block.
		\item \textit{Transition layer}: It aggregates feature maps from the Dense block and reduces its dimension (e.g., enabling max-pooling with $ 1\times1 $ convolutions).
		\item \textit{Composite functions}: Additional sequence of operations that present in one convolutional layer i.e., Batch Normalization - $ ReLU $ - $ Conv2D $.
	\end{itemize}
	\begin{figure}[H]
		\centering
		\includegraphics[width=\textwidth, height=0.85in]{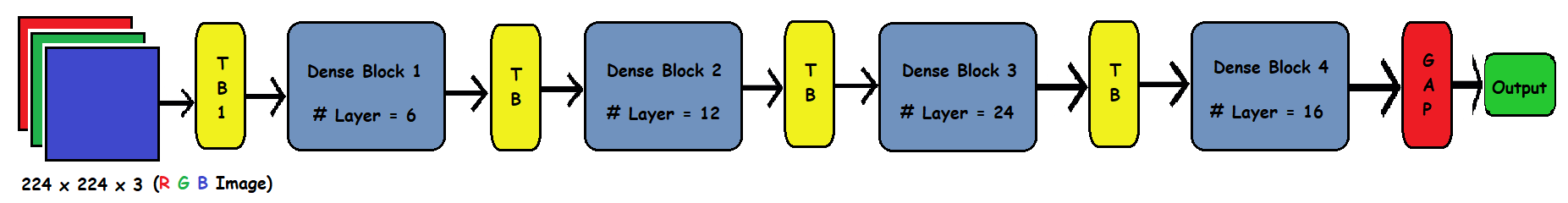}
		\caption{DenseNet-121 Full Architecture}
		\label{fig:dense_full}
	\end{figure}
	\paragraph{Xception Net}
	
	The Xception model \cite{xception} is separated into three flows:
	\begin{center}
		\begin{enumerate*}
			\item Entry Flow
			\item Middle Flow
			\item Exit Flow
		\end{enumerate*}
	\end{center}
	All these three flows include depth-wise separable convolutional neural networks along with convolutional skip connection, the one we saw in ResNet. The specialty of Xception lies in the fact that it uses a separable convolutional layer, which is widely used for the following two reasons:
	\begin{itemize}[leftmargin=*]
		\item Due to the lesser number of parameters, they are easy to handle compared to standard CNN and hence reduce overfitting.
		\item They are computationally cheaper because of fewer computations, making them an appropriate model for mobile vision applications.
	\end{itemize}
	{\noindent Both} Xception and MobileNet \cite{mobilenets} (both proposed by Google) use separable convolutional layers.\\
	
	{\noindent\textit{Computational Complexity}}:
	\begin{enumerate}[leftmargin=*]
		\item In standard convolution operation, for input data of size $ D_{f}\times D_{f}\times M $ ($ D_{f} $ = width and height of the image, $ M = $ no. of channels) when we use $ N $ filters/kernels of size $ D_{k}\times D_{k}\times M $. the total number of multiplication operations use for output size of $ D_{p}\times D_{p}\times N  = \mathbf{N \times D_{p}^{2}\times D_{k}^{2}\times M}$.
		\item While in depth-side separable convolutions, we go through two significant steps:
		\begin{enumerate}[leftmargin=*]
			\item Depth-wise convolution or filtering stage: Here, $ M $ number of filters of size $ D_{k}\times D_{k}\times1 $ convoluted over $ D_{f}\times D_{f}\times M $ sized image -- one kernel on each channel -- to obtain $ D_{p}\times D_{p}\times M $ intermediate output. And for this operation, we require a total of $ M\times D_{p}^{2}\times D_{k}^{2} $  multiplication.
			\item Point-wise convolution/ combination stage: Here, $ N $ number of $ 1\times1\times M $ sized kernels are applied over the intermediate output to produce a total of $ N $, $ D_{p}\times D_{p}\times 1 $ sized matrix, which is then stacked to give the outcome of size $ D_{p}\times D_{p}\times N $. And the total cost of this operation $ = M\times D_{p}^{2}\times N $.
		\end{enumerate}
		\item Now comparing the complexity of depth-wise separable convolution with standard convolution, we get a ratio of $ = \dfrac{M\times D_{p}^{2}\times \left[D_{k}^{2} + N\right]}{M\times D_{p}^{2}\times D_{k}^{2}\times N} = \mathbf{\dfrac{1}{N} + \dfrac{1}{D_{k}^{2}}} $
		\item For example, consider $ N = 100 $ and $ D_{k} = 512 $.Then the ratio becomes $ 0.010004 $, which says Depth-wise separable convolution network performs $ 100 $ times lesser multiplication than the standard one.
	\end{enumerate}
	\begin{figure}[H]
		\centering
		\includegraphics[width=\linewidth]{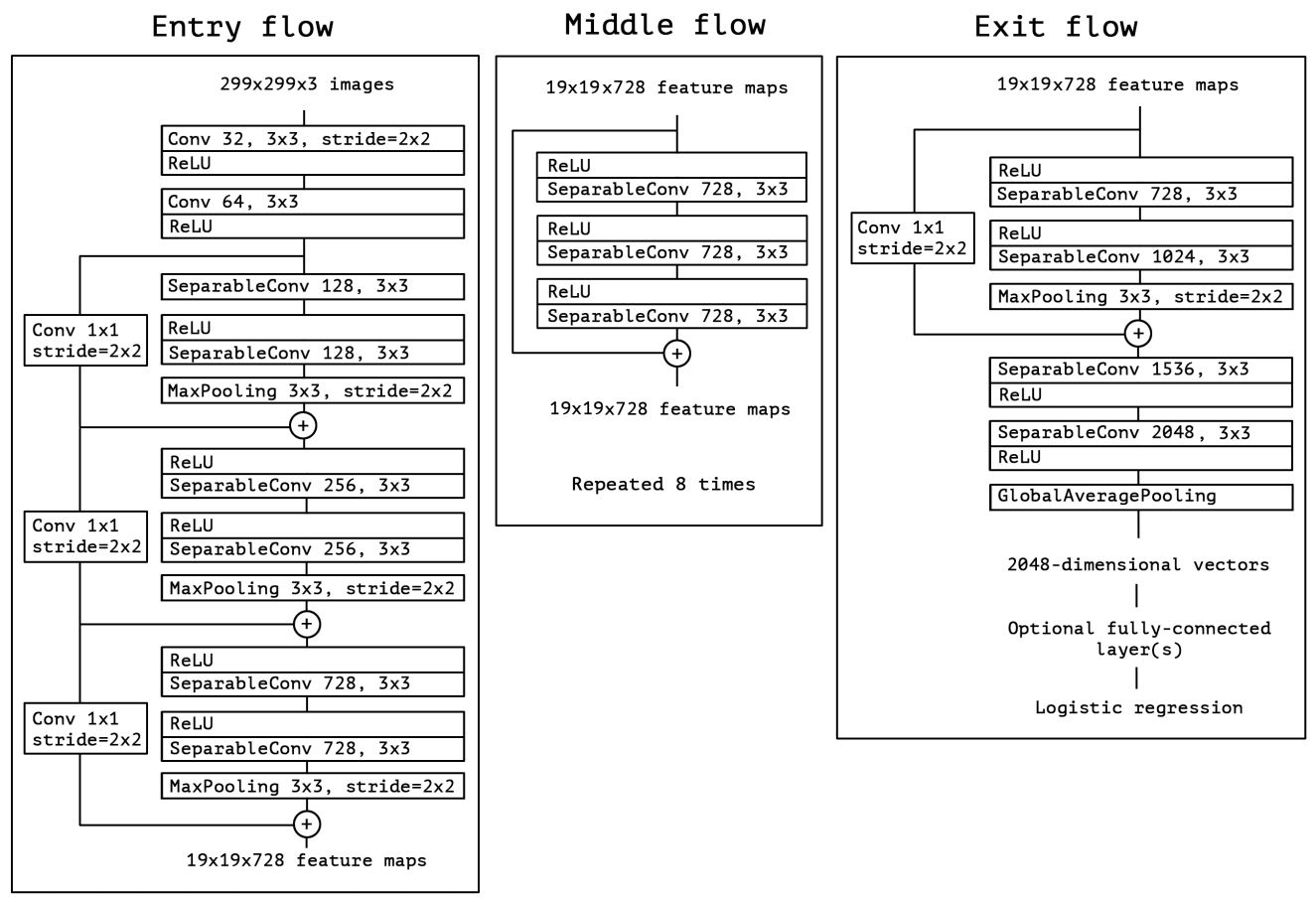}
		\caption{Xception Architecture \cite{xception}}
		\label{fig:xception}
	\end{figure}
	\newpage
	\thispagestyle{empty}
	\rhead{Experiment and Results}
	\section[Chapter 4]{\Large Chapter 4}\label{chap4}
	{\Huge \textbf{Experiment and Results}}
	\vspace*{0.8cm}
	\hrule
	\subsection[Evaluation Metric]{\Large Evaluation Metric}
	
	We have chosen F$ _{2} $ and F$ _{1} $ as our evaluation metric, with F$ _{2} $ being the primary one and F$ _{2} $ and F$ _{1} $, both are special cases of F$ _{\beta} $ score, which are defined as:
	$$ F_{\beta} = (1 + \beta^{2})\dfrac{pr}{\beta^{2}p + r} $$ for $ \beta = 2 $ and  $ 1 $, respectively, and $ p = \dfrac{TP}{TP + FP} =  $ precision while  $ r = \dfrac{TP}{TP + FN} = $ recall.\\
	
	{\noindent(\emph{Remarks:} TP = True Positive, FP = False Positive, TN = True Negative, FN = False Negative)}\\
	
	The final F$ _{2} $  and F$ _{1} $ scores are mean F$ _{2} $ and  mean F$ _{1} $  scores, which are calculated by averaging the F$ _{2} $ and F$ _{1} $ scores for each image in the test/  Val set. Consider that F$ _{1} $ is the harmonic mean of precision and recall $ \dfrac{1}{F_{1}} = \dfrac{1}{2}\left(\dfrac{1}{p}+\dfrac{1}{r}\right) $, giving equal weightage to precision and recall. While in the case of F$ _{2} $, more weight is conferred to recall as $ \dfrac{1}{F_{2}} = \dfrac{4}{5r} + \dfrac{1}{5p} $. Thus, the reason behind why we have made F$ _{2} $ our primary score is due to the importance/weightage it gives to recall. Since our dataset has a considerable amount of rare label images, the models couldn't learn those features accurately and falsely predict their absence in many such images even if they actually present, essentially increasing false negatives (FN) and thereby lowering the recall. To penalize such mistakes, F$ _{2} $ scoring, which prioritizes recall, becomes a better classification scoring criterion. Further, there are several averaging standards for the $ F_{\beta} $ \cite{metrics}, the precision, and the recall that differs in terms of support of the classes like, 
	\begin{itemize}[leftmargin=*]
		\item \textit{Macro-averaging}: It caters equal weight to each class and computes the average per class scores but simultaneously ignores label imbalance. Say, we have $ k $ classes in the given dataset. And let their corresponding True Positive, True Negative, False Positive, and False Negative counts be $ TP^{i},$ $ TN^{i},$ $ FP^{i} $, and $ FN^{i} \ \forall \ i = 1, 2, \cdots, k$, respectively. Then
		\begin{align*}
			\text{Macro-precision} &= \dfrac{1}{k}\sum\limits_{i=1}^{k}p^{i},\quad\text{where } p^{i} = \dfrac{TP^{i}}{TP^{i} + FP^{i}}\\
			\text{Macro-recall} &= \dfrac{1}{k}\sum\limits_{i=1}^{k}r^{i},\quad\text{where } r^{i} = \dfrac{TP^{i}}{TP^{i} + FN^{i}}\\
			\text{Macro-}F_{\beta} &= \dfrac{1}{k}\sum\limits_{i=1}^{k}F_{\beta}^{i},\quad\text{where } F_{\beta}^{i} = \dfrac{\left(1+\beta^{2}\right)\left(p^{i}.r^{i}\right)}{\left(\beta^{2}.p^{i} + r^{i}\right)} = \dfrac{\left(1 + \beta^{2}\right)TP^{i}}{\left(1 + \beta^{2}\right)TP^{i} + FP^{i} + FN^{i}}\\
			&\forall\ i = 1, 2, \cdots, k
		\end{align*} 
		\item \textit{Weighted-averaging}: The only difference here is that users manually input weights for each class. Say, $ w^{1}, w^{2}, \cdots, w^{k} $ be the corresponding weights for $ k $ classes such that $ \sum_{i=1}^{k}w^{i} = 1 $. Then
		\begin{align*}
			\text{Weighted-precision} &= \sum\limits_{i=1}^{k}w^{i}p^{i}\\
			\text{Weighted-recall} &= \sum\limits_{i=1}^{k}w^{i}r^{i}\\
			\text{Weighted-}F_{\beta} &= \sum\limits_{i=1}^{k}w^{i}F_{\beta}^{i}\\
			&\forall\ i = 1, 2, \cdots, k
		\end{align*}
	\end{itemize}
	While in terms of support of the instances, we have two additional averaging criteria.
	\begin{itemize}[leftmargin = *]
		\item \textit{Micro-averaging}: It calculates metrics globally by counting the total true positives (TP), false negatives (FN), and false positives (FP). This aggregates the contributions of all classes to compute the average metric. It takes the support of all classes and instances as a whole, due to which it takes label imbalance and volume of instances into account. If the distribution of classes is symmetrical (i.e., you have an equal number of samples for each class), then macro and micro scores will be the same.\\
		Mathematically,
		\begin{align*}
			\text{Micro-precision} &= \dfrac{\text{Total count of TP}}{\text{Total count of TP} + \text{Total count of FP}} = \dfrac{\sum_{i=1}^{k}TP^{i}}{\sum_{i=1}^{k}TP^{i} + \sum_{i=1}^{k}FP^{i}}\\
			\text{Micro-recall} &= \dfrac{\text{Total count of TP}}{\text{Total count of TP} + \text{Total count of FN}} = \dfrac{\sum_{i=1}^{k}TP^{i}}{\sum_{i=1}^{k}TP^{i} + \sum_{i=1}^{k}FN^{i}}\\
			\text{Micro-}F_{\beta} &= \dfrac{\sum_{i=1}^{k}\left(1 + \beta^{2}\right)TP^{i}}{\sum_{i=1}^{k}\left(1 + \beta^{2}\right)TP^{i} + \sum_{i=1}^{k}FP^{i} + \sum_{i=1}^{k}FN^{i}}
		\end{align*}
		\item \textit{Sample-averaging}: It calculates metrics for each instance and finds their average, meaning it considers each sample to be independent. This rule is only meaningful for multi-label classification. Following is the mathematical formulation for sample-averaging method. Say, we have $ m $ instances in the dataset. And for each instance, we compute the confusion matrix that details count of classes/labels (instead of instances)  in each section, namely, true positive, true negative, false positive, and false negative. Let those corresponding quantities in confusion matrix for $ i^{th} $ instance be $ tp^{i} $, $ tn^{i} $, $ fp^{i} $, and $ fn^{i} $, respectively, $ \forall\ i = 1, 2, \cdots, m $. Then,
		\begin{align*}
			\text{Sample-precision} &= \dfrac{1}{m}\sum\limits_{i=1}^{m}precision^{i},\ \text{where } precision^{i} = \dfrac{tp^{i}}{tp^{i} + fp^{i}}\\
			\text{Sample-recall} &= \dfrac{1}{m}\sum\limits_{i=1}^{m}recall^{i},\ \text{where } recall^{i} = \dfrac{tp^{i}}{tp^{i} + fn^{i}}\\
			\text{Sample-}F_{\beta} &= \dfrac{1}{m}\sum\limits_{i=1}^{m}\text{sample-}F_{\beta}^{i},\\
			&\text{where } \text{sample-}F_{\beta}^{i} = \dfrac{\left(1 + \beta^{2}\right)tp^{i}}{\left(1 + \beta^{2}\right)tp^{i} + fp^{i} + fn^{i}}\ \forall\ i = 1, 2, \cdots, m
		\end{align*} 
	\end{itemize}
	If our goal is to maximize hits (TP and TN) and minimize misses (FP and FN), micro-averaging should be considered. On the contrary, if one values the minority class the most, then macro-averaging is the one. But with this, we would not be able to capture dependency among labels. Since our task is a multi-label classification task with an imbalanced dataset, we chose the \textit{sample-averaging method} for F$ _{2} $, F$ _{1} $, precision, and recall.

	\subsection[Improving Performance]{\Large Improving Performance}
	
	We'll discuss some techniques that we have implemented while training our models and pre-processing images to improve our performance score in the subsequent subsection.
	
	\subsubsection[Data-Augmentation]{\large Data-Augmentation}
	
	Since our architectures added layers onto the existing ImageNet models (which in turn added more parameters), there is a high chance that our model tries to learn the noise and irregularities in the data. Thus we wanted to prevent our model from overfitting. And, given a limited number of training data, we can transform our images such that the exact location of houses, roads, mines, and other features changes by which we can have an ample volume of data with a considerable amount of images w.r.t to rare labels. In order to mitigate these issues, we distorted our training images. In particular, we experimented with random left-right flipping, random up--down image flipping, and random clockwise-anticlockwise $ 90^{0} $ image rotation. By adding random distortions to the training images at each epoch, we prevented the model from seeing the images many times and generally succeeded in avoiding overfitting, which we can verify through the plots and tables in the \enquote{quantitative evaluation} section \ref{quant}.
	\begin{figure}[h!]
		\centering
		\begin{subfigure}{0.5\linewidth}
			\includegraphics[width=0.85\linewidth]{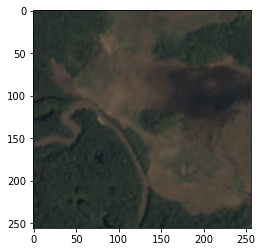}
			\label{fig:Original}
			\caption{Original Image}
		\end{subfigure}\hfil
		\begin{subfigure}{0.5\linewidth}
			\includegraphics[width=0.85\linewidth]{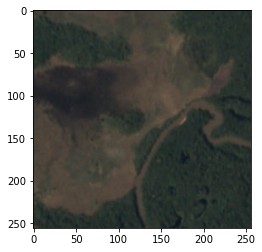}
			\label{fig:Horiz_flip}
			\caption{Horizontally Flipped}
		\end{subfigure}\hfil
	
		\medskip
		
		\begin{subfigure}{0.5\linewidth}
			\includegraphics[width=0.85\linewidth]{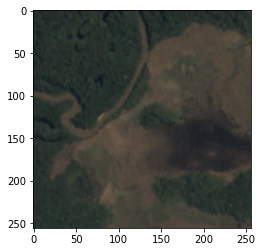}
			\label{fig:Vert_flip}
			\caption{Vertically Flipped}
		\end{subfigure}\hfil
		\begin{subfigure}{0.5\linewidth}
			\includegraphics[width=0.85\linewidth]{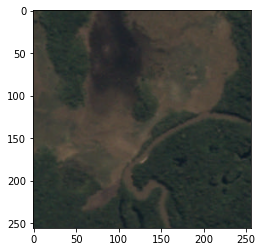}
			\label{fig:Rotated}
			\caption{Rotated Image}
		\end{subfigure}\hfil
		\label{fig:augmentation}
		\caption{Data Augmentation}
	\end{figure}

	\subsubsection[Fine-tuning]{\large Fine-tuning}
	
	As discussed earlier, fine-tunning is a method where the topmost block(s) of a pre-trained model is allowed to update its weights while training with a very minimal learning rate to get accustomed to our data.
	
	\subsubsection[Model Callbacks]{\large Model Callbacks}
	
	Callbacks are essentially a set of functions that are applied at certain stages of the training procedure. Informally, they are the helper instruments, e.g., as in a car, the velocimeter displays fuel usage, speed, tire pressure, etc. These values indicate the level of car safety helping us to drive safely and under complete control. Callbacks are those information clusters along with the feature of self-driving formula that monitor our loss and early stop our model if it hits a dead end and so much more.
	
	\paragraph{EarlyStopping}
	
	It \cite{es} is a form of regularization that prevents overfitting. In an extensive network, during training, there will be a point when the model will stop generalizing and start learning the statistical noise in the training data. Hence, our objective is to train the network long enough to learn the mapping from inputs to outputs, but not the noise.\\ 
	
	Our approach to solving this problem is to treat the number of training epochs as a hyperparameter, i.e., train the model multiple times with different values of epochs and then select that number of epoch for which we get the best results on hold out test set. The downside of the above approach is that it is computationally expensive and time-consuming as it requires training the model multiple times and discards it.\\
	
	An alternate approach is to train the model once for a large no. of training epochs while keeping track of the score on the holdout test set. If the model's performance on the validation set starts to degrade, then the training process is halted. At this point, the model is said to have good generalizability. Usually, two steps are being followed while we want to stop our training early. First, monitoring model performance, and secondly, set a trigger to stop the training as soon as validation performance falls.
	
	\paragraph{ModelCheckpoint}
	
	ModelCheckpoint will save our model as a checkpoint file (in \textit{hdf5} format) after those epochs where our model has performed better than earlier epochs. Here the performance is assessed according to our preferred metric. In our case, the metric is the F$ _{2} $ score, and the more F$ _{2} $ is, the better is our model.
	
	\subsubsection[Ensemble]{\large Ensemble}\label{sec:ensemble}
	
	In simple terms, ensemble learning combines individual learners from heterogenous or homogenous modeling in order to obtain a \enquote{combined leaner} that improves the generalizability (avoiding the risk of overfitting).\\
	
	{\noindent\textit{Dietterich}} \cite{dietterich} given three reasons why ensemble is better than \enquote{single.}
	\begin{enumerate}[leftmargin=*]
		\item Picking only one single classifier, we run the risk of choosing a bad one.
		\item Many algorithms use local search and may not find the optimal classifier. So running several times the learning algorithm and combining the obtained models results in a better approximation to the optimal classifier than why single one.
		\item In most ML problems, the function that can address the query absolutely can't be found. Hence ensemble does provide a way to reach optimal classifier by combining several classifiers.
	\end{enumerate}
	
	{\noindent\textbf{Commmon Types of Ensemble:}}
	
	\paragraph{Bagging}\label{sec:bagging}
	
	A method homogenous weak learners learns independently from each other as a small sample population and combines their prediction following some deterministic averaging process. Example: Random Forest Classifier \cite{random-forest}, Extra Trees Classifier \cite{etc}
	
	\paragraph{Boosting}\label{sec:boosting}
	
	This is an iterative technique where homogenous weak learners learn sequentially in an adaptive manner, i.e., it adjusts the weights of observations based on the last classification. Example: Adaboost \cite{adaboost-1, adaboost-2}, Gradient Boosting Classifier \cite{gbc-1, gbc-2}
	
	\paragraph{Stacking}
	
	It is a method in which homogenous or heterogenous weak learners learn in parallel and train a meta-model on their combined predictions to output the required prediction based on different weak model's predictions.\\
	
	{\noindent Roughly, we can assert that Bagging focuses on getting an ensemble model with less variance while boosting and stacking try to produce less biased strong models with respect to their components. Plus, the variance problem can also be addressed using these techniques.}\\
	
	{\noindent In our case, we have used two forms of stacking ensemble, namely, \textit{weighted stacking ensemble} and \textit{integrated stacking ensemble}. They both differ only on the basis of how they combine their prediction to train meta-model and do they even use meta-model or not.}
	\begin{enumerate}[leftmargin=*]
		\item \emph{Weighted Stacking Ensemble:} This is a weighted majority voting ensemble technique where various models are given certain weights as per the performance. Let's see how these weights are assigned.\\
		The ensemble procedure is as follows: Let $ (y^{j})\in \{0,\ 1\}^{17} $ and $ w^{j}(\in \mathbb{N}) $ be, respectively, the binarized predicted label vector for a given input image and the weight of the $ j^{th} $ model. We form the vector of label votes $ \tilde{v} = (v_{1}, \ldots, v_{17}) $ as $$\tilde{v} = \sum_{j=1}^{n}w^{j}.y^{j}$$ where $ n= $ number of heterogenous learners and the $ j^{th} $ model contributes either 0 or $ w^{j} $ votes to the total number of votes for each label. Given the weights, the maximum number of votes for an instance is $ \sum_{j=1}^{n}w^{j} $. The final predictions $ y = (y_{1}, y_{2}, \ldots, y_{17}) $ are made by applying a threshold to each label against a common threshold value $ \left(th = \dfrac{\sum_{j=1}^{n}w^{j}}{2}\right) $. Specifically, $$y_{i} = 
		\begin{cases}
			1,\ \text{if}\ v_{i} > th\quad \forall\ i = 1, 2, \ldots, 17\\
			0,\ else
		\end{cases}$$
		Also, we can tune our class-wise threshold as per the requirement to get the maximum evaluation score which is F$ _{2} $ in our case. Using precision-recall-curve, we can do so. We'll discuss more this in subsection class-specific threshold. Note that this is a simplistic implementation of the ensemble technique where no meta-model could be seen in action.
		\item \emph{Integrated Stacking Ensemble (ISE):} In this method, several heterogenous sub-models/networks are embedded in a larger multi-headed model/network that learns how to best combine the predictions from each input sub-model. It allows treating this stacked ensemble as a single model. The advantage of this approach is that output of all the submodels is fed directly to the meta-learner. Moreover, it is too possible to update the weights of the submodels in conjunction with the meta-learner model if this is desirable. But in our case, we will first train the sub-models and then use their predictions on the holdout validation set to train the meta-model. Because the meta-model should learn how to map the output of submodels on unseen data to the required labels, not the data on which it is trained. This would indirectly incorporate a high variance/ overfitting issue.\\
		
		Once all our submodels are trained, these are used as a separate input head to the meta-model, which receives $ k $ copies of any input instance ($ k = $ number of input sub-models). Note that the input shape of different submodels can be different, which is adjusted accordingly at the time of input. The output of each submodel is then merged using a simple concatenation merge that produces a single $ k\times17 $-element vector from the 17 class probabilities predicted by each of the k models. Here, the meta-model is basically two-layered or more than two-layered ANN whose hidden layer helps interpret the input to the meta-leaner. The output layer will make its probabilistic prediction of 17-element vector.
	\end{enumerate}
	\subsubsection[Class-wise thresholding]{\large Class-wise thresholding}
	
	It is mainly used to mitigate the class imbalance problem or the scenario where the cost of one type of misclassification is more critical than another type of misclassification. This method chooses the optimal threshold using the decision function output and the ground truth values for the given metric. Decision scores below the decision threshold are labeled as the negative class (0), and those with higher values are labeled as the positive class (1). Using \textit{Precision-Recall curves}, we can select the best value for the decision threshold such that it gives high precision or high recall or both (e.g., in $ F_{1} $ or $ F_{2} $) based on whether our project is precision-recall-oriented or $ F_{\beta} $-oriented, respectively. Usually, the area under the curve (AUC) summarizes the skill of a model across thresholds.\\
	
	Or else, we can go for the ROC curve (\textit{Reciever Operating Characteristics}) curve, which offers the advantage of comparing different models directly in general or for different thresholds. And the AUC is used as a summary of the model skill, like in precision-recall AUC. It should be used when there aren't any (or few in multi-class case) class-imbalance issues. Here, the decision scores can be any classification metric. But the ROC curve itself provides a metric called roc\_auc score, which points to the AUC for the ROC curve.
	
	\subsection[Training]{\Large Training}
	
	Since our ultimate move for improving overall performance is through stacking 11 ImageNet sub-models, we train each of them to decide upon the depth of ANN classifier required to attach at the top of every ImageNet submodel. Plus, we optimize the number of nodes in each layer and the dropout rate within each of them. Furthermore, other hyperparameters, including the optimization algorithm, learning rate, and regularization, need to be tuned. Let's briefly discuss our training step by step.
	
	\begin{enumerate}[leftmargin=*]
		\item[]\textit{Step 1}: We split our training dataset of size $ 40,479 $ into five separate folds, each of approximately equal size, such that the proportion of each class in all the five folds is more or less the same (Stratified CV). We have made two copies of our whole dataset, one with a pixel resolution of $ 224\times224 $ and the other $ 299\times299 $ as per the input requirement of different models.
		\item[]\textit{Step 2}: We executed a five-fold Stratified CV and stacked the model's predictions for holdout validation fold in each loop for later use.
	\end{enumerate}

	While the model is being trained, image augmentations (random left-right flipping, random up-down image flipping, and random clockwise-anticlockwise $ 90^{0} $ image rotation), as well as image pre-processing (done by their respective preprocessor - discussed in the methodology chapter \ref{chap3}) both, are performed by \textit{ImageDataGenerator} function of \textit{Keras} \cite{keras-imagedata}. This function also facilitates batch-wise augmentation and pre-processing (with batch-size $ = 128 $, as more than that could not be loaded into our given memory). As a result, the next batch comes into the picture (loaded into the data generator function) only after the training of the current model finishes on the current batch. Further, for each model, the number of epochs to be fully trained varies, which can be considered a hyperparameter.\\ 
	
	In the first ten epochs, the whole ImageNet model is frozen, and only the top neural network classifier is allowed to train so that the classifier get accustomed to the type of input and output (i.e., basically learning the function that maps reduced information vector ImageNet to the required labels). This approach prevents the top classifier from yielding dumb predictions when the last block of ImageNet is unfrozen after the tenth epoch. Because till that epoch, ANN knows the type of input it expects, and hence even if the input changes a bit from the expected distribution, ANN will adjust itself by modifying its weights within a small margin so that it can produce a sensible output.\\
	
	Furthermore, the learning rate is reduced by $ 10 $ folds from $ 10^{-4} $ to $ 10^{-5} $ in the second phase of training so that the learning curve doesn't go berserk and we can progress to the optimal point on the loss curve (i.e., minima of the loss surface) fluently. We have used \textit{Adam} \cite{adam} as our optimizer with \textit{amsgrad} mode on. For the second phase of training, model callbacks are deployed into the model. With \textit{EarlyStopping}, we monitored validation loss with some patience value decided upon by the user (different for different models), and with \textit{ModelCheckpoint}, we saved the best model observed during training with a maximum F$ _{2} $ score for later use. Next, we forbid \textit{BN} \cite{bn} layer to update its parameters during the training phase because doing so will lose all the statistical information it learned previously and produce unsatisfactory performance. Though it allows faster training as well as a wide range of learning rates without compromising convergence, it is very fragile in the sense that it changes its weights while predicting the holdout test set. And that's why we pre-process our images to that format with which they are trained before, as a result of which our model hardly will distinguish the current data from the previous ones.
	
	\subsection[Results]{\Large Results}\label{quant}
	
	\subsubsection[Results - For Machine Learning Models]{\large Results - For Machine Learning Models}
	
	Here we present our $ F_2 $ score and BCE loss for all the $ 5 $ folds obtained from Stratified 5-Fold splitting. 
	\begin{table}[H]
		\centering
		\begin{minipage}[c]{\textwidth}
			\centering
			\scalebox{0.9}{
			\begin{tabular}{|m{2.5cm}|m{2.5cm}|m{2.5cm}|m{2.5cm}|m{1.5cm}|m{1.5cm}|}
				\hline
				Parameters&n\_components&solver&priors&shrinkage&tol\\
				\hline
				\hline
				Values&$ \min $(number of classes - 1, number of features)&svd = Singular Value Decomposition&Class proportions are inferred from the training data&no shrinkage&$ 0.0001 $\\
				\hline
			\end{tabular}}
			\caption{Parameter values chosen during training for LDA}
		\end{minipage}
		\medskip
		\begin{minipage}[c]{\textwidth}
			\scalebox{0.99}{
				\begin{tabular}{|l|l|l|l|l|l|l|l|}
					\hline
					Model Name & Fold 1 & Fold 2 & Fold 3 & Fold 4 & Fold 5 & Average & Score Type\\
					\hline\hline
					LDA & 0.87700 & 0.88040 & 0.88276 & 0.88433 & 0.88112 & \cellcolor[RGB]{255, 255, 0}0.88112 & F2 Score\\
					\hline
					&0.68651 & 0.68168 & 0.66527 & 0.66794 & 0.67844 & 0.67597 & BCE Loss\\
					\hline
			\end{tabular}}
			\caption{Top Row: F$ _{2} $ Scores, Bottom Row: BCE Loss for LDA}
		\end{minipage}
		\caption{LDA}
		\label{table:LDA}
	\end{table}
	\begin{table}[H]
		\centering
		\begin{minipage}[c]{\textwidth}
			\centering
			\scalebox{0.85}{
			\begin{tabular}{|l|c|}
				\hline
				Parameters&Values\\
				\hline
				\hline
				base\_score&$ 0.5 $\\
				\hline
				max\_depth&10\\
				\hline
				n\_estimators&100\\
				\hline
				eval\_metric&logloss\\
				\hline
				booster&gbtree\\
				\hline
				colsample\_bylevel&1\\
				\hline
				colsample\_bynode&1\\
				\hline
				colsample\_bytree&1\\
				\hline
				gamma&0\\
				\hline
				learning\_rate&0.001\\
				\hline
				objective&\enquote*{binary:logistic}\\
				\hline
			\end{tabular}}
			\caption{Parameter values for XGBoost}
		\end{minipage}
		\medskip
		\begin{minipage}[c]{\textwidth}
			\scalebox{0.99}{
				\begin{tabular}{|l|l|l|l|l|l|l|l|}
					\hline
					Model Name & Fold 1 & Fold 2 & Fold 3 & Fold 4 & Fold 5 & Average & Score Type\\
					\hline\hline
					XGBoost & 0.86041 & 0.86333 & 0.86303 &	0.86794 & 0.86204 &	\cellcolor[RGB]{255, 255, 0}0.86335 & F2 Score\\
					\hline
					&0.69912 & 0.69107 & 0.67832 & 0.66548 & 0.68668 & 0.68413 & BCE Loss\\
					\hline
			\end{tabular}}
			\caption{F$ _{2} $ Scores; BCE Loss}
		\end{minipage}
		\caption{XGBoost}
		\label{table:xgboost}
	\end{table}
	\begin{table}[H]
		\centering
		\begin{minipage}[c]{\textwidth}
			\centering
			\scalebox{0.99}{
			\begin{tabular}{|l|c|}
				\hline
				Parameters & Values\\
				\hline
				\hline
				bootstrap& False\\
				\hline
				n\_estimators& 200\\
				\hline
				criterion& \enquote*{gini}\\
				\hline
				ccp\_alpha& $ 0.0 $\\
				\hline
				max\_depth& None\\
				\hline
				max\_features& \enquote*{auto}\\
				\hline
				max\_leaf\_nodes& None\\
				\hline
				max\_samples& None\\
				\hline
				min\_samples\_leaf& 1\\
				\hline
				min\_samples\_split& 2\\
				\hline
			\end{tabular}}
			\caption{Parameters and their values for ETC}
		\end{minipage}
		\medskip
		\begin{minipage}[c]{\textwidth}
			\scalebox{0.99}{
				\begin{tabular}{|l|l|l|l|l|l|l|l|}
					\hline
					Model Name & Fold 1 & Fold 2 & Fold 3 & Fold 4 & Fold 5 & Average & Score Type\\
					\hline\hline
					ETC & 0.85675 & 0.86142 & 0.85945 & 0.86012 & 0.86179 & \cellcolor[RGB]{255, 255, 0}0.85991 & F2 Score\\
					\hline
					&0.66709 & 0.66815 & 0.66527 & 0.66794 & 0.66844 & 0.66597 & BCE Loss\\
					\hline
			\end{tabular}}
			\caption{F$ _{2} $ Scores; BCE Loss}
		\end{minipage}
		\caption{Extra Trees Classifier (ETC)}
		\label{table:etc}
	\end{table}
	\begin{table}[H]
		\centering
		\begin{minipage}[c]{\textwidth}
			\centering
			\scalebox{0.89}{
			\begin{tabular}{|l|c|}
				\hline
				Parameters&Values\\
				\hline
				\hline
				n\_estimators& $ 200 $\\
				\hline
				bootstrap& True\\
				\hline
				ccp\_alpha& $ 0.0 $\\
				\hline
				criterion& \enquote*{auto}\\
				\hline
				max\_depth& None\\
				\hline
				max\_features& \enquote*{auto}\\
				\hline
				max\_leaf\_nodes& None\\
				\hline
				max\_samples& None\\
				\hline
				min\_samples\_leaf& $ 1 $\\
				\hline
				min\_samples\_split& $ 2 $\\
				\hline
			\end{tabular}}
			\caption{Parameter values for RFC}
		\end{minipage}
		\medskip
		\begin{minipage}[c]{\textwidth}
			\scalebox{0.99}{
			\begin{tabular}{|l|l|l|l|l|l|l|l|}
					\hline
					Model Name & Fold 1 & Fold 2 & Fold 3 & Fold 4 & Fold 5 & Average & Score Type\\
					\hline\hline
					RFC & 0.861024 & 0.86142 & 0.86545 & 0.86012 & 0.86379 & \cellcolor[RGB]{255, 255, 0}0.86236 & F2 Score\\
					\hline
					&0.64332 & 0.64815 & 0.64527 & 0.64794 & 0.64844 & 0.64597 & BCE Loss\\
					\hline
			\end{tabular}}
			\caption{F$ _{2} $ Scores; BCE Loss}
		\end{minipage}
		\caption{Random Forest Classifier (RFC)}
		\label{table:RFC}
	\end{table}
	\begin{table}[H]
		\centering
		\begin{minipage}[c]{\textwidth}
			\centering
			\scalebox{0.83}{
				\begin{tabular}{|p{9.5cm}|p{4cm}|}
					\hline
					Parameters&Values\\
					\hline
					\hline
					criterion& \enquote*{friedman\_mse}\\
					\hline
					init & None\\
					\hline
					learning\_rate; validation\_fraction& $ 0.1 $\\
					\hline
					loss& \enquote*{deviance}\\
					\hline
					max\_depth& $ 3 $\\
					\hline
					max\_features; max\_leaf\_nodes; min\_impurity\_split& None\\
					\hline
					min\_impurity\_decrease; min\_weight\_fraction\_leaf; ccp\_alpha& $ 0.0 $\\
					\hline
					min\_samples\_leaf& $ 1 $\\
					\hline
					min\_samples\_split& $ 2 $\\
					\hline
					n\_estimators & $ 100 $\\
					\hline
					tol & $ 0.0001 $\\
					\hline
					subsample & $ 1.0 $\\
					\hline
			\end{tabular}}
			\caption{Parameter values for GBC}
		\end{minipage}
		\medskip
		\begin{minipage}[c]{\textwidth}
			\scalebox{0.89}{
				\begin{tabular}{|l|l|l|l|l|l|l|l|}
					\hline
					Model Name & Fold 1 & Fold 2 & Fold 3 & Fold 4 & Fold 5 & Average & Score Type\\
					\hline\hline
					GBC & 0.861024 & 0.86142 & 0.86545 & 0.86012 & 0.86379 & \cellcolor[RGB]{255, 255, 0}0.86236 & F2 Score\\
					\hline
					&0.64332 & 0.64815 & 0.64527 & 0.64794 & 0.64844 & 0.64597 & BCE Loss\\
					\hline
				\end{tabular}}
			\caption{F$ _{2} $ Scores; BCE Loss}
		\end{minipage}
		\caption{Gradient Boosting Classifier (GBC)}
		\label{table:gbc}
	\end{table}
	\subsubsection[Results - For Baseline Model]{\large Results - For Baseline Model}
	\begin{figure}[H]
		\centering
		\includegraphics[width=0.75\linewidth]{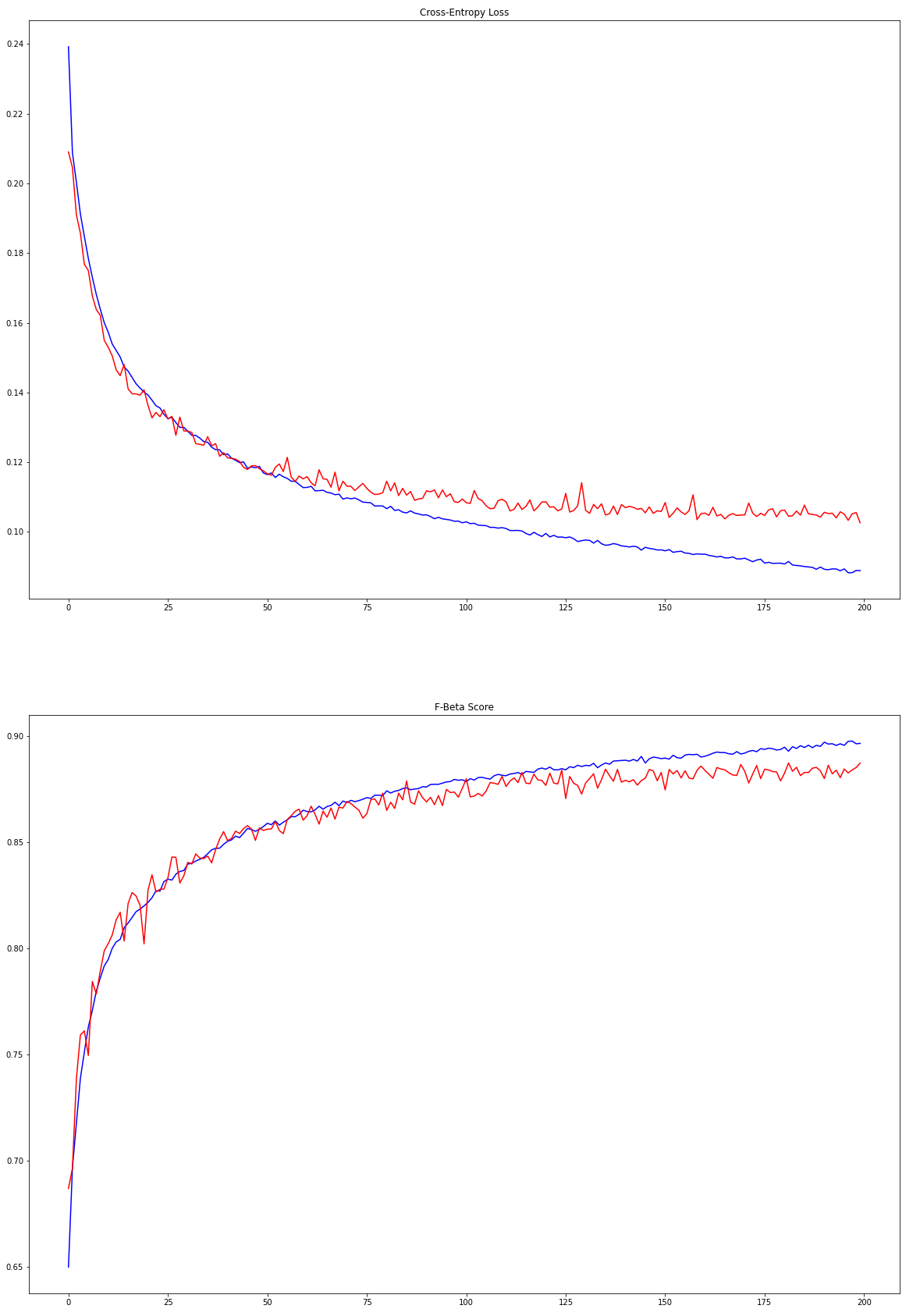}
		\caption{F$ _{2} $ score as well as hybrid cross-entropy loss plot for Baseline model}
		\label{fig:baseline_plot}
	\end{figure}
	\subsubsection[Results - For Deep Learning Model]{\large Results - For Deep Learning Model}
	We present our performance on the holdout validation set for each of the eleven architectures only for the first fold in the upcoming subsection. Accompanying the F$ _{2} $ score for each architecture, we have showcased class-wise precision, recall,  F$ _{1} $ score, accuracy score, model architecture diagram, and F$ _{2} $ score as well as hybrid cross-entropy loss plot for both training and validation set.\\
	
	{\noindent \textbf{Full Architectures of different models}}
	\begin{figure}[H]
		\centering
		\begin{subfigure}[b]{0.3\textwidth}
			\includegraphics[width=\linewidth]{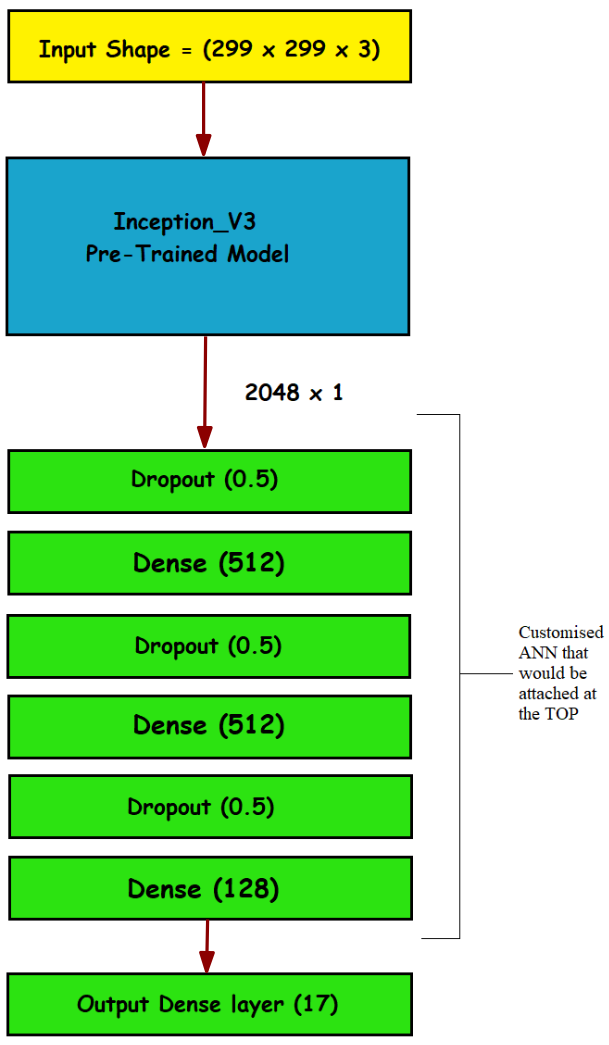}
			\caption{Inception-V3}
			\label{fig:inception_v3_full}
		\end{subfigure}
		\hfill
		\begin{subfigure}[b]{0.3\textwidth}
			\includegraphics[width=\linewidth]{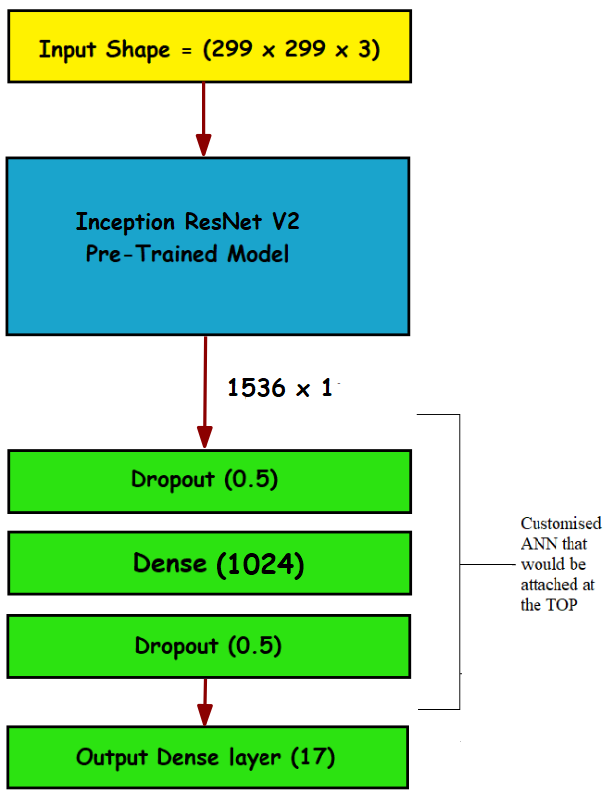}
			\caption{Inception ResNet-V2}
			\label{fig:inception_resnet_full}
		\end{subfigure}
		\hfill
		\begin{subfigure}[b]{0.3\textwidth}
			\includegraphics[width=\linewidth]{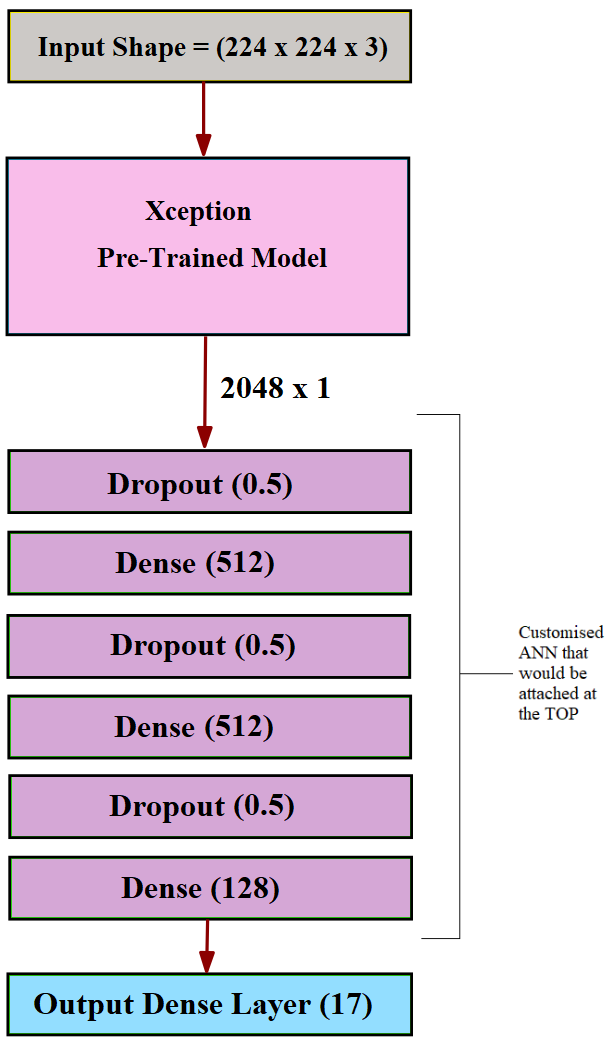}
			\caption{Xception}
			\label{fig:xception_full}
		\end{subfigure}
		\caption{Full Architecture of Inception-V3, Inception-ResNet-V2, and Xception}
		\label{fig:full_archs}
	\end{figure}
	\begin{figure}[H]
		\centering
		\begin{subfigure}[b]{0.3\textwidth}
			\includegraphics[width=\linewidth]{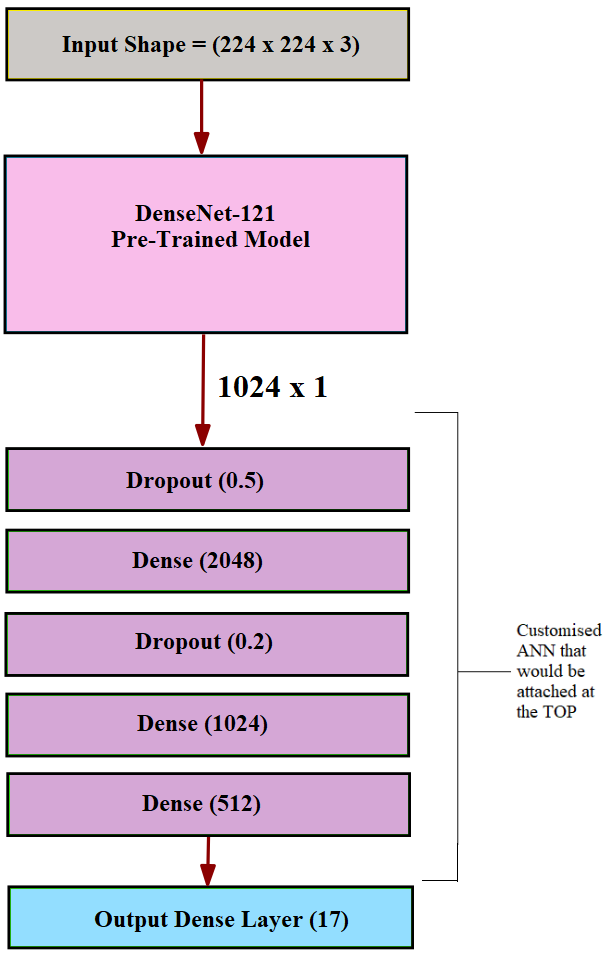}
			\caption{DenseNet-121}
			\label{fig:dense121_full}
		\end{subfigure}
		\hfill
		\begin{subfigure}[b]{0.3\textwidth}
			\includegraphics[width=\linewidth]{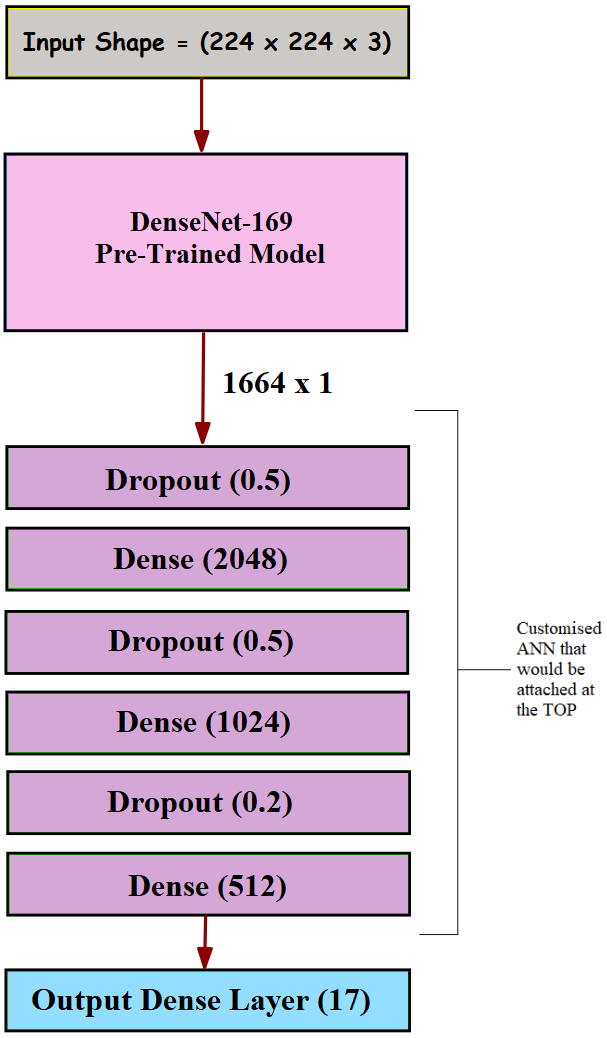}
			\caption{DenseNet-169}
			\label{fig:dense169_full}
		\end{subfigure}
		\hfill
		\begin{subfigure}[b]{0.3\textwidth}
			\includegraphics[width=\linewidth]{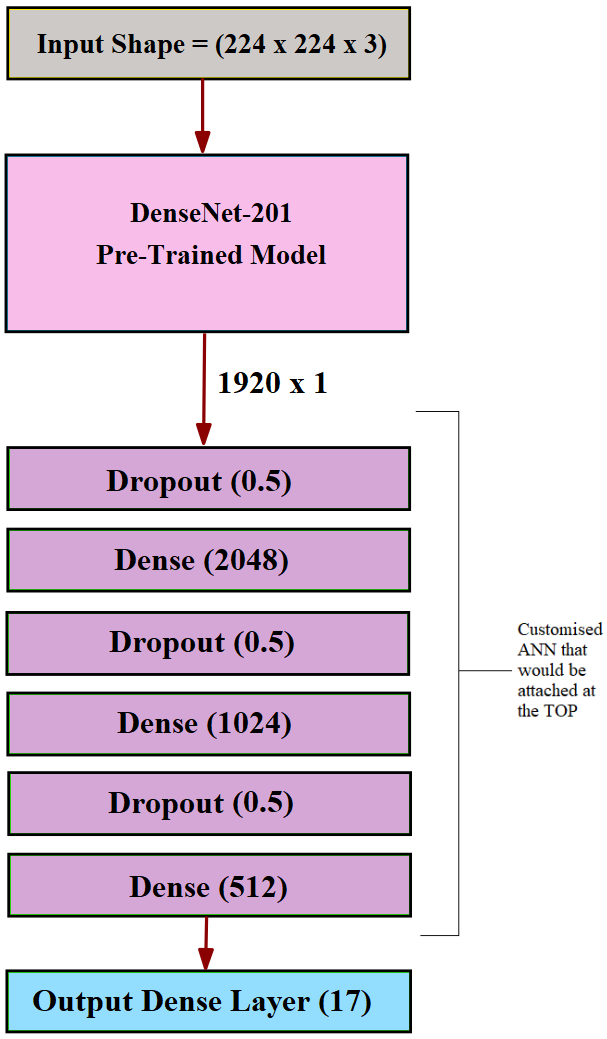}
			\caption{DenseNet-201}
			\label{fig:dense201_full}
		\end{subfigure}
		\caption{Full DenseNet Architectures}
		\label{fig:densenet_full_archs}
	\end{figure}
	\begin{figure}[H]
		\centering
		\begin{subfigure}[b]{0.3\textwidth}
			\includegraphics[width=\linewidth]{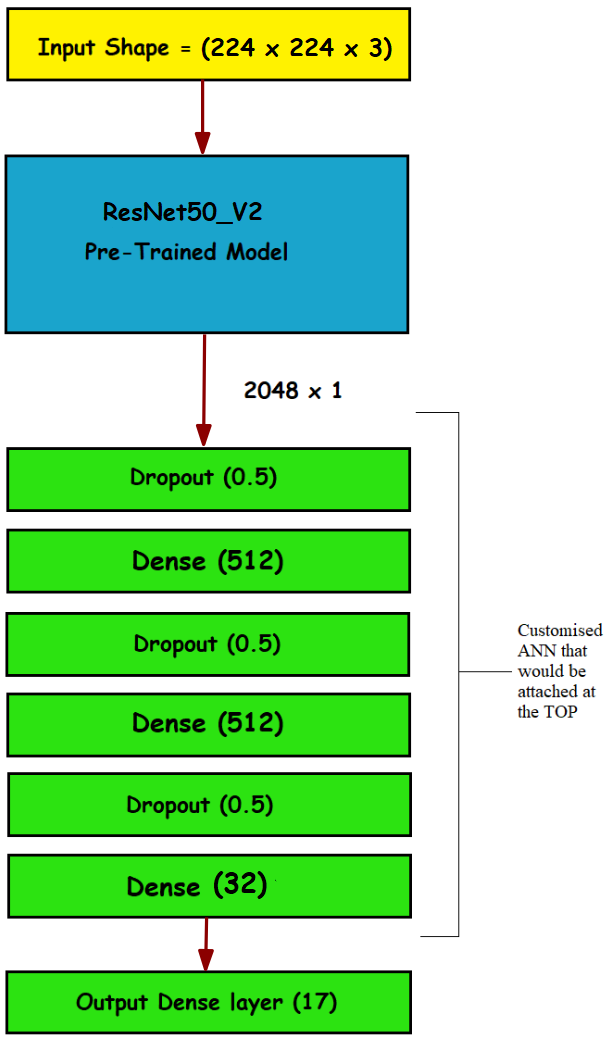}
			\caption{ResNet-50 V2}
			\label{fig:resnet50_v2_full}
		\end{subfigure}
		\hfill
		\begin{subfigure}[b]{0.3\textwidth}
			\includegraphics[width=\linewidth]{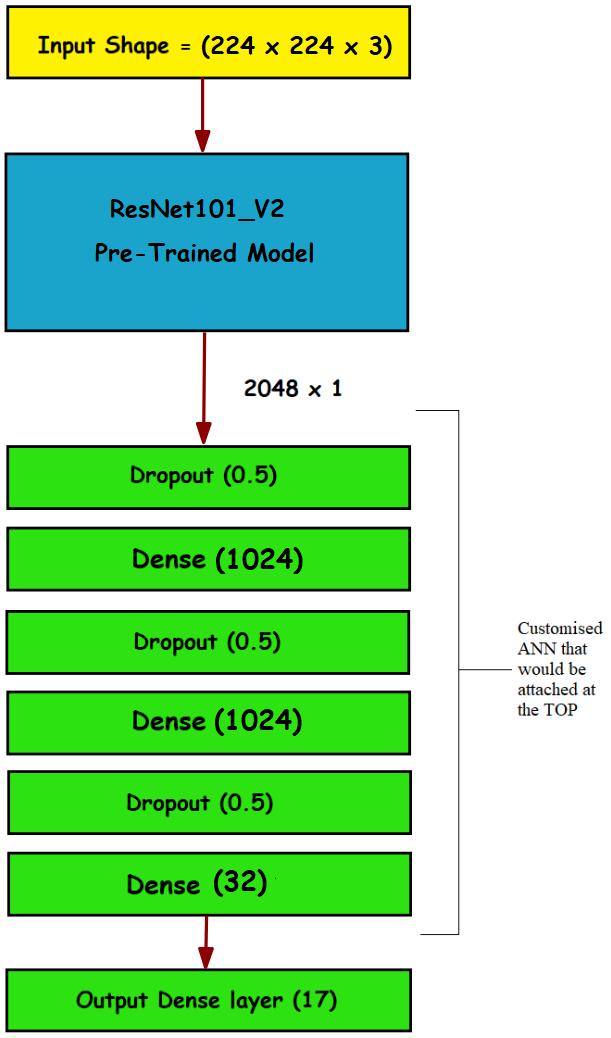}
			\caption{ResNet-101 V2}
			\label{fig:resnet101_v2_full}
		\end{subfigure}
		\hfill
		\begin{subfigure}[b]{0.3\textwidth}
			\includegraphics[width=\linewidth]{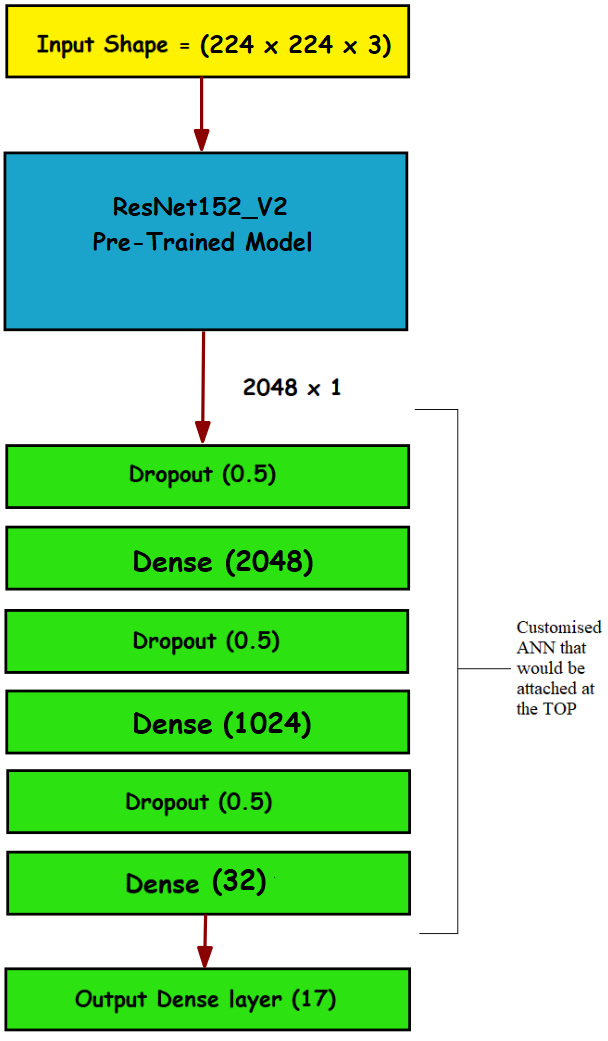}
			\caption{ResNet-152 V2}
			\label{fig:resnet152_v2_full}
		\end{subfigure}
		\caption{Full ResNet Architectures}
		\label{fig:resnet_full_archs}
	\end{figure}
	\begin{figure}[H]
		\centering
		\begin{subfigure}[b]{0.42\textwidth}
			\includegraphics[width=\linewidth]{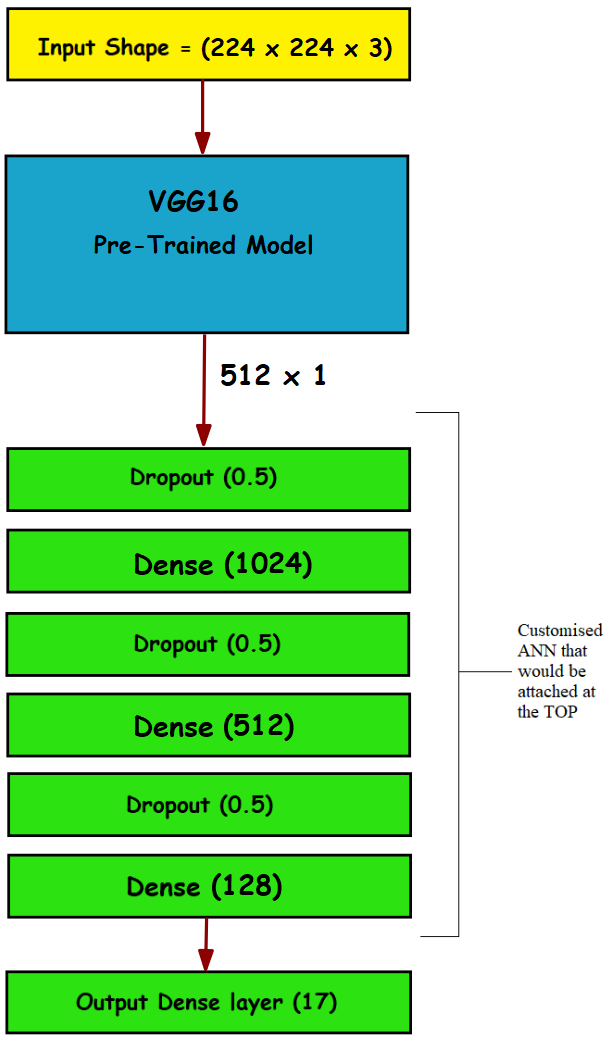}
			\caption{VGG16}
			\label{fig:vgg16_full_arch}
		\end{subfigure}
		\hfill
		\begin{subfigure}[b]{0.42\textwidth}
			\includegraphics[width=\linewidth]{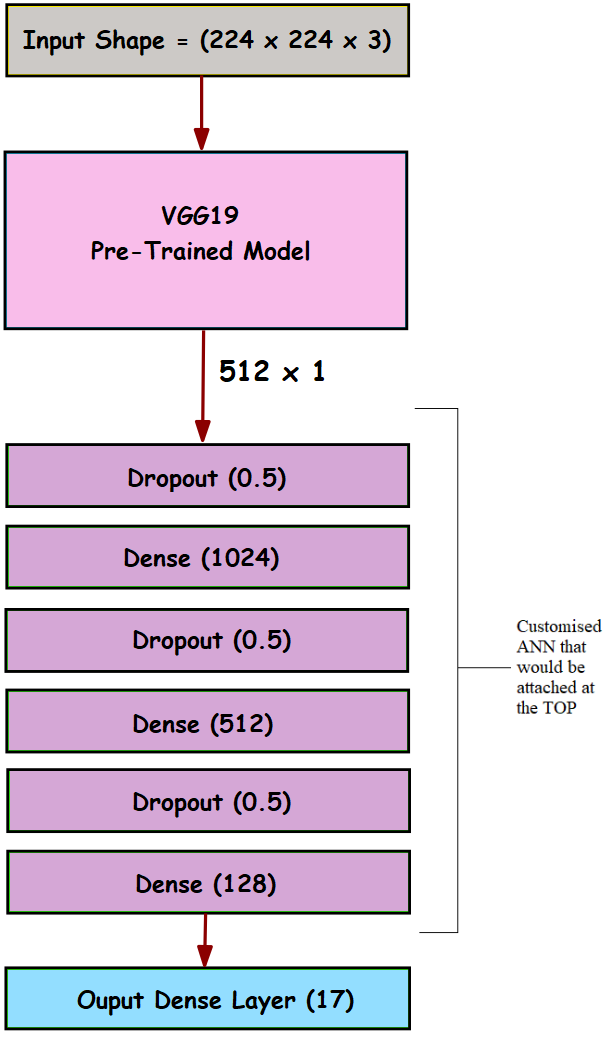}
			\caption{VGG19}
			\label{fig:vgg19_full_arch}
		\end{subfigure}
		\caption{Full VGG Architectures}
		\label{fig:vgg_full_archs}
	\end{figure}

	{\noindent\textbf{F2 Score \& BCE Loss Plot for Training and Validation Set}}\\
	\begin{figure}[H]
		\centering
		\begin{subfigure}[b]{0.49\textwidth}
			\includegraphics[width=\linewidth]{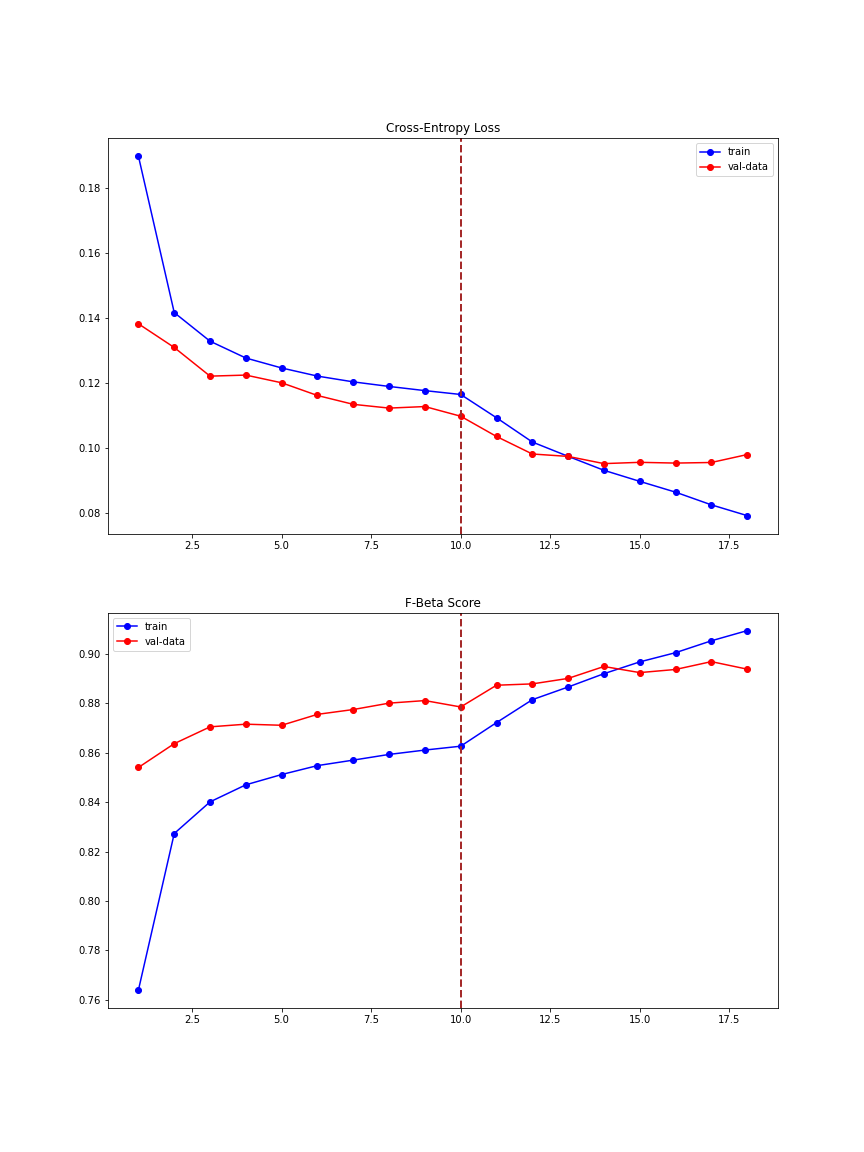}
			\caption{Inception-V3}
			\label{fig:inception-v3}
		\end{subfigure}
		\hfill
		\begin{subfigure}[b]{0.49\textwidth}
			\includegraphics[width=\linewidth]{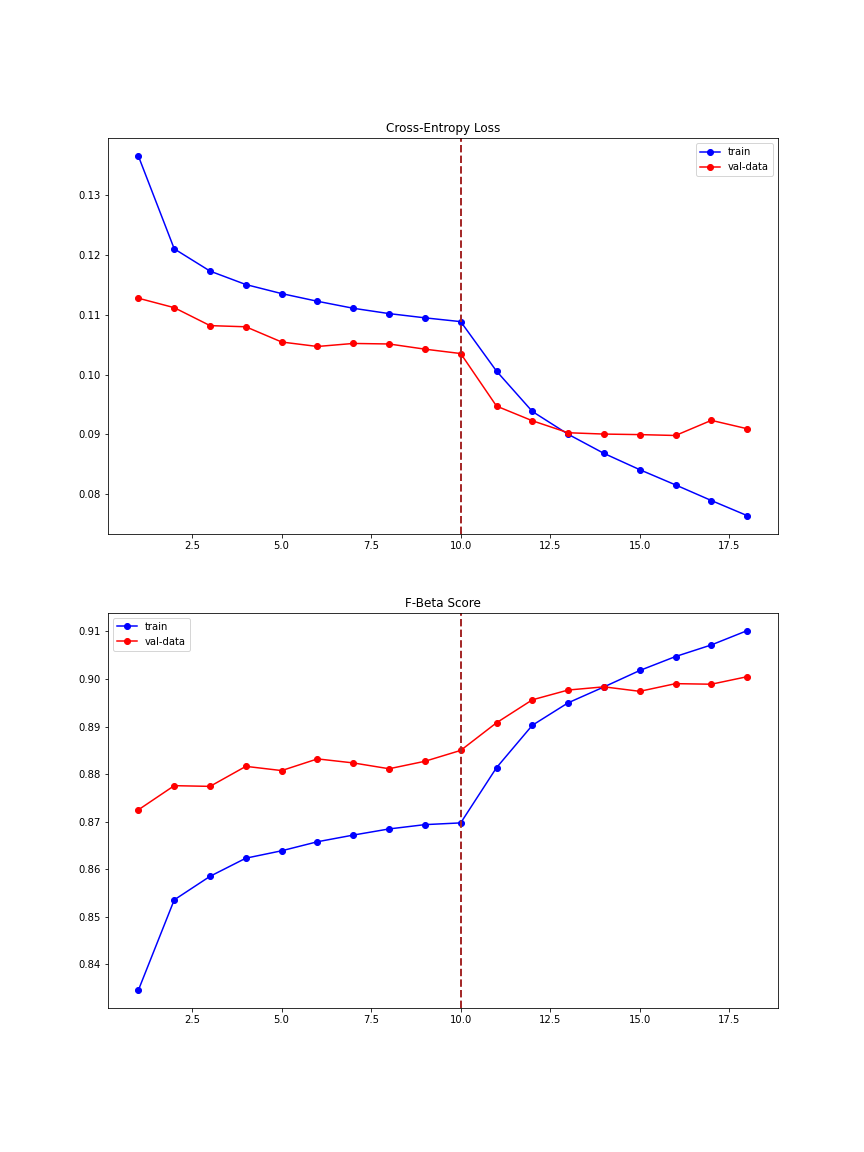}
			\caption{Inception-ResNet-V2}
			\label{fig:loss_score_inc_res_v2}
		\end{subfigure}
		\medskip
		\begin{subfigure}[b]{0.49\textwidth}
			\includegraphics[width=\linewidth]{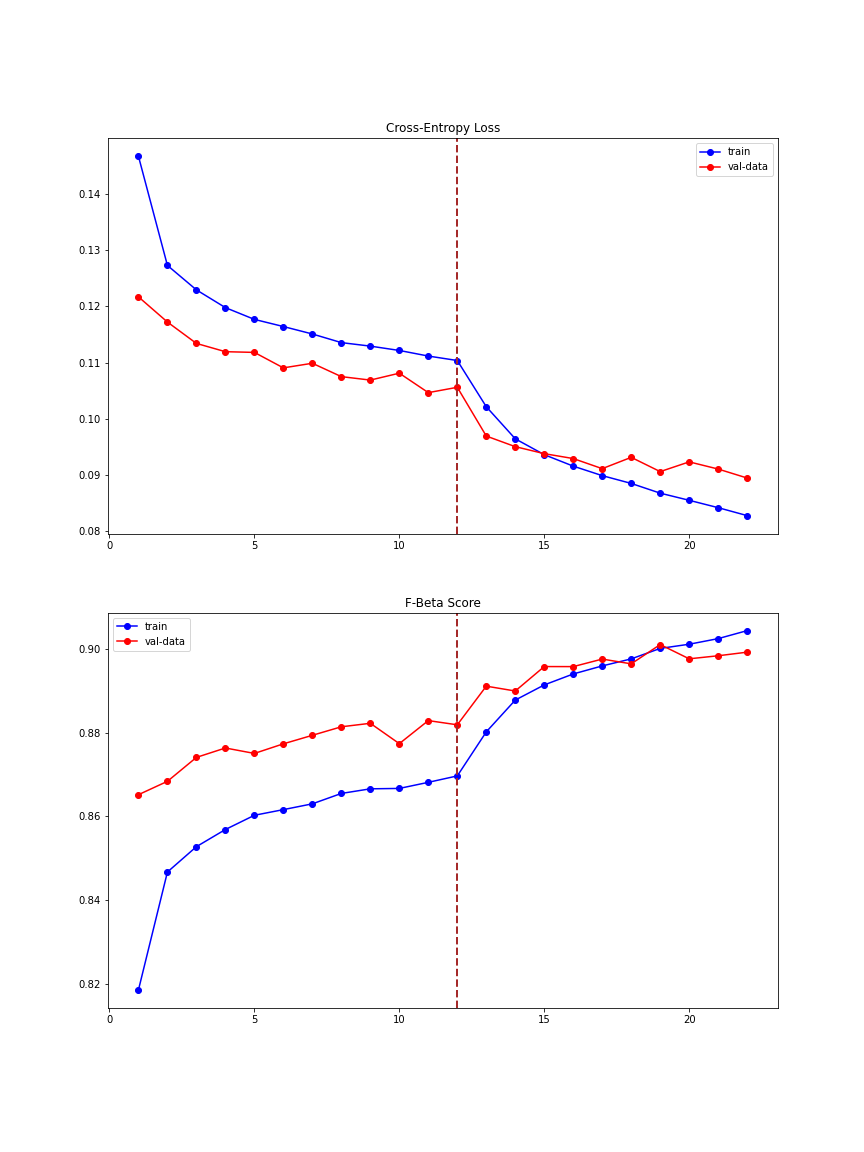}
			\caption{Xception}
			\label{fig:loss_score_xcpetion}
		\end{subfigure}
		\hfill
		\begin{subfigure}[b]{0.49\textwidth}
			\includegraphics[width=\linewidth]{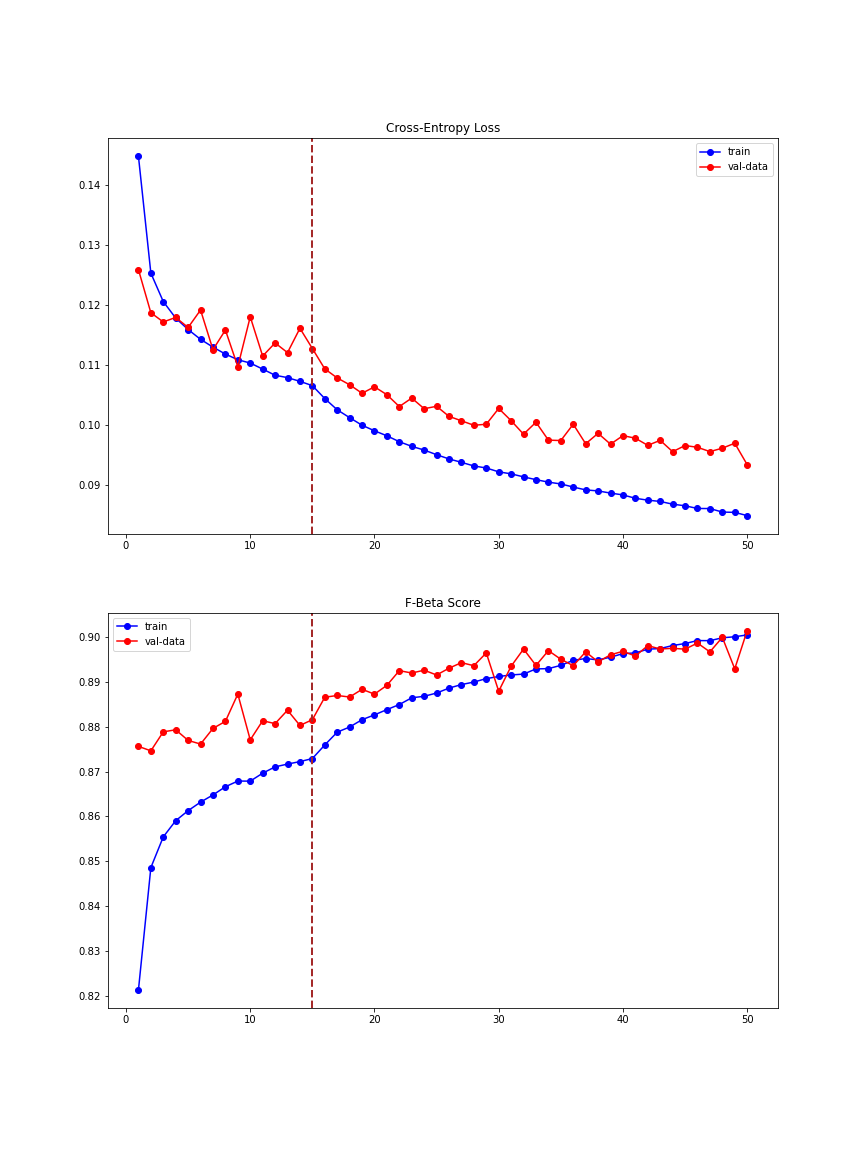}
			\caption{DenseNet-121}
			\label{fig:loss_score_dense121}
		\end{subfigure}
	\end{figure}
	\begin{figure}[H]
		\centering
		\begin{subfigure}[b]{0.49\textwidth}
			\includegraphics[width=\linewidth]{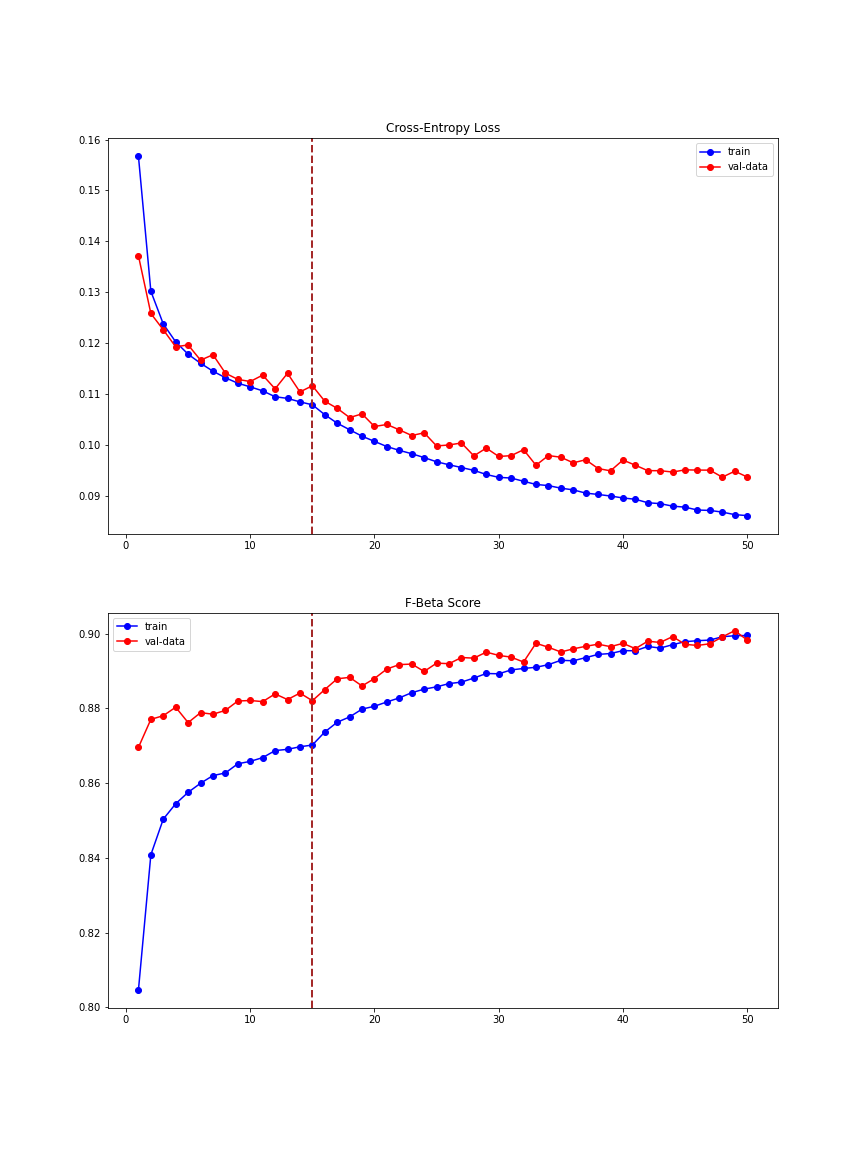}
			\caption{DenseNet-169}
			\label{fig:loss_score_dense169}
		\end{subfigure}
		\hfill
		\begin{subfigure}[b]{0.49\textwidth}
			\includegraphics[width=\linewidth]{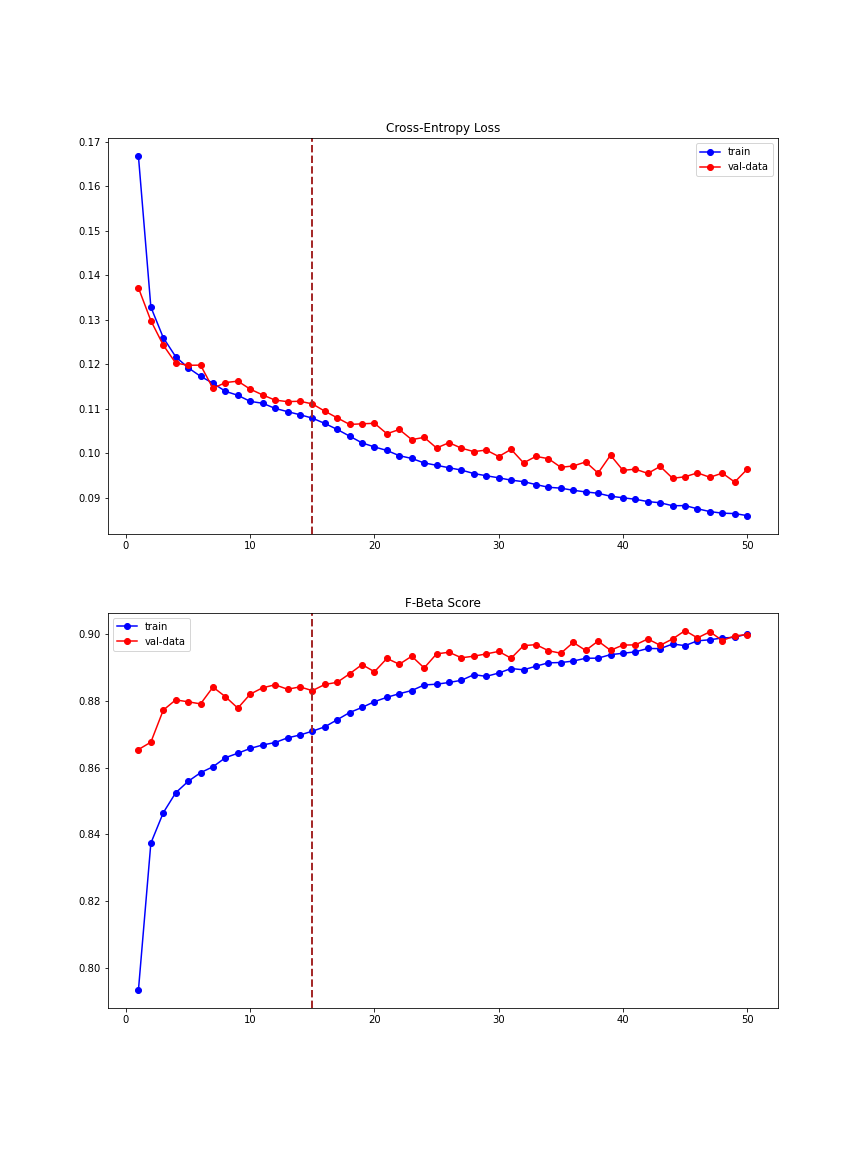}
			\caption{DenseNet-201}
			\label{fig:loss_score_dense201}
		\end{subfigure}
		\medskip
		\begin{subfigure}[b]{0.49\textwidth}
			\includegraphics[width=\linewidth]{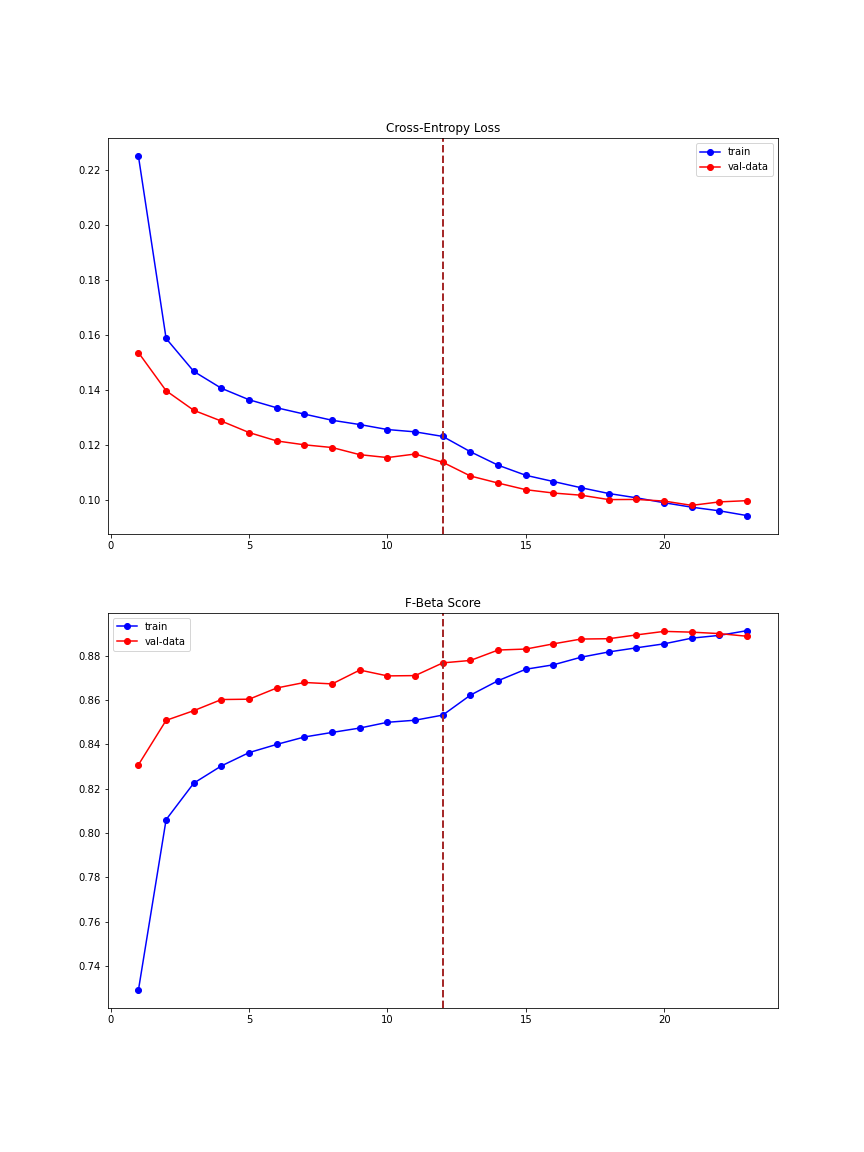}
			\caption{ResNet-50 V2}
			\label{fig:loss_score_resnet50}
		\end{subfigure}
		\hfill
		\begin{subfigure}[b]{0.49\textwidth}
			\includegraphics[width=\linewidth]{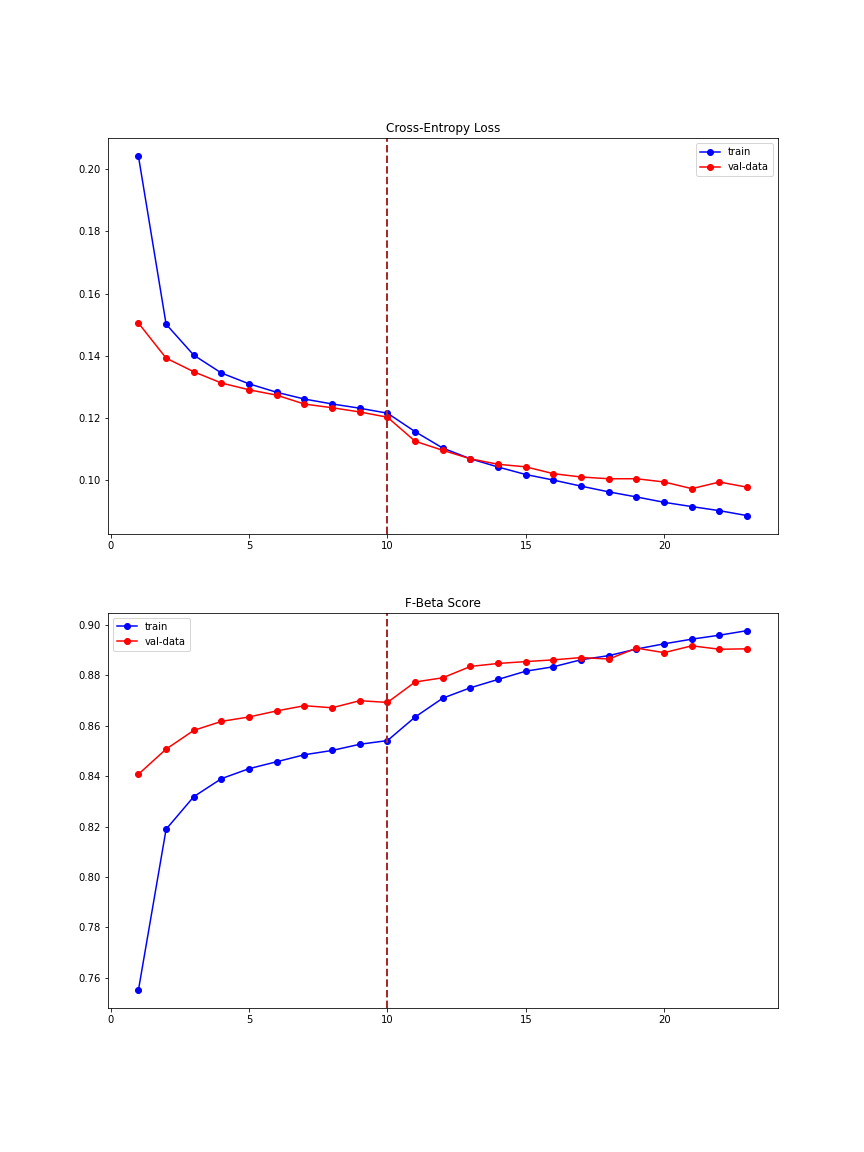}
			\caption{ResNet-101 V2}
			\label{fig:loss_score_resnet101}
		\end{subfigure}
	\end{figure}
	\begin{figure}[H]
		\centering
		\begin{subfigure}[b]{0.49\textwidth}
			\includegraphics[width=\linewidth]{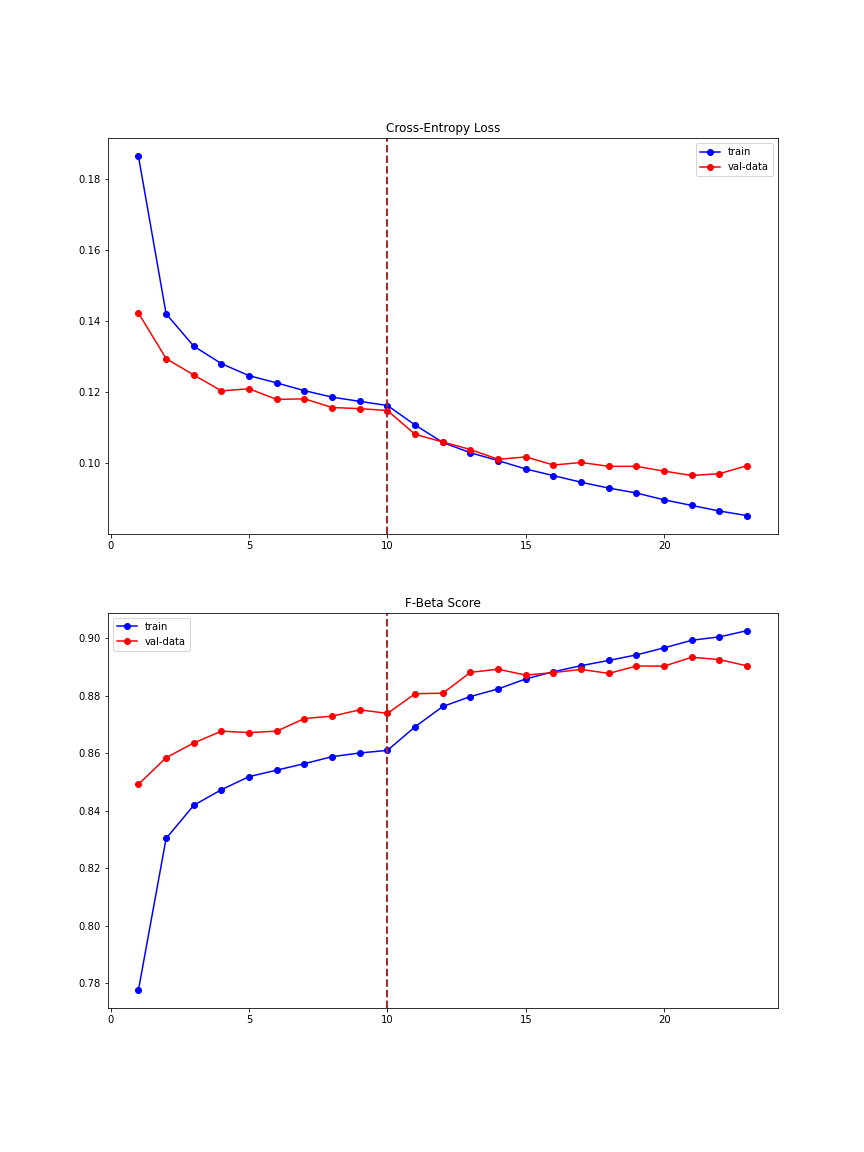}
			\caption{ResNet-152 V2}
			\label{fig:loss_score_resnet152}
		\end{subfigure}
		\hfill
		\begin{subfigure}[b]{0.49\textwidth}
			\includegraphics[width=\linewidth]{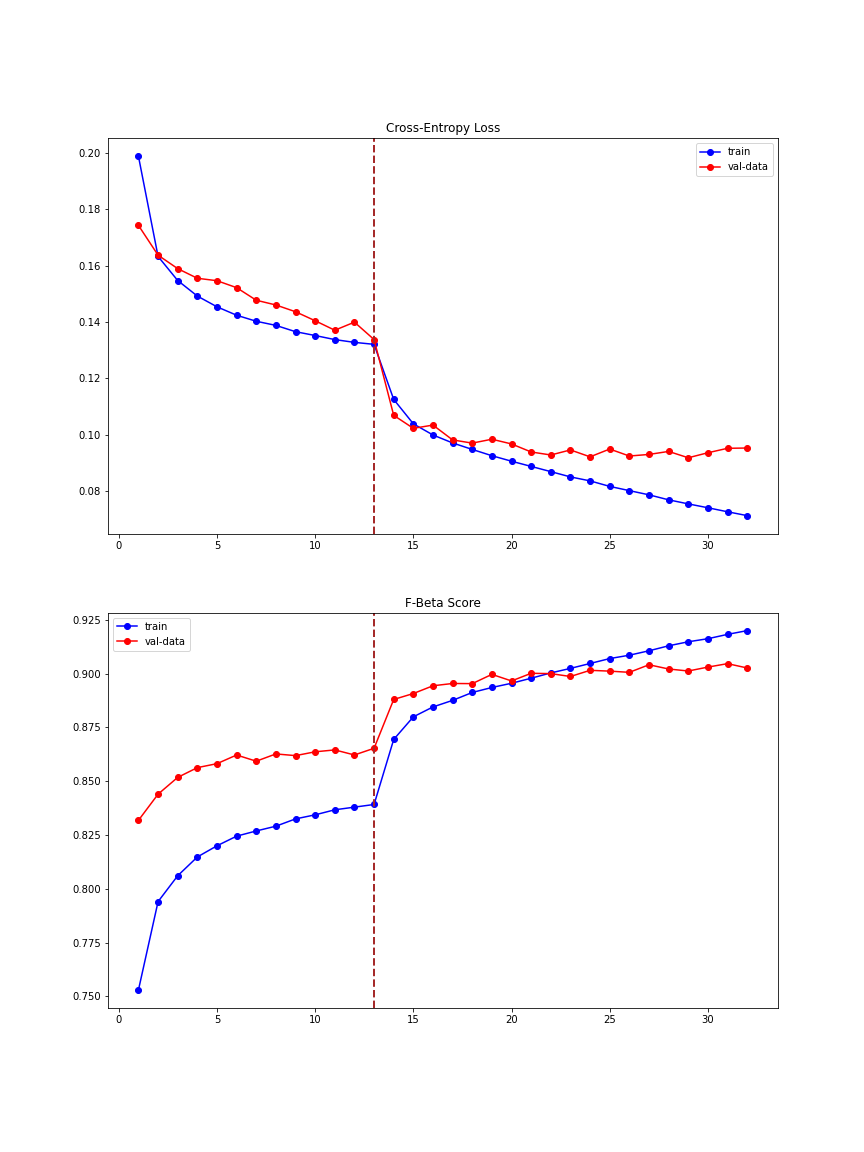}
			\caption{VGG16}
			\label{fig:loss_score_vgg16}
		\end{subfigure}
		\medskip
		\begin{subfigure}[b]{0.49\textwidth}
			\includegraphics[width=\linewidth]{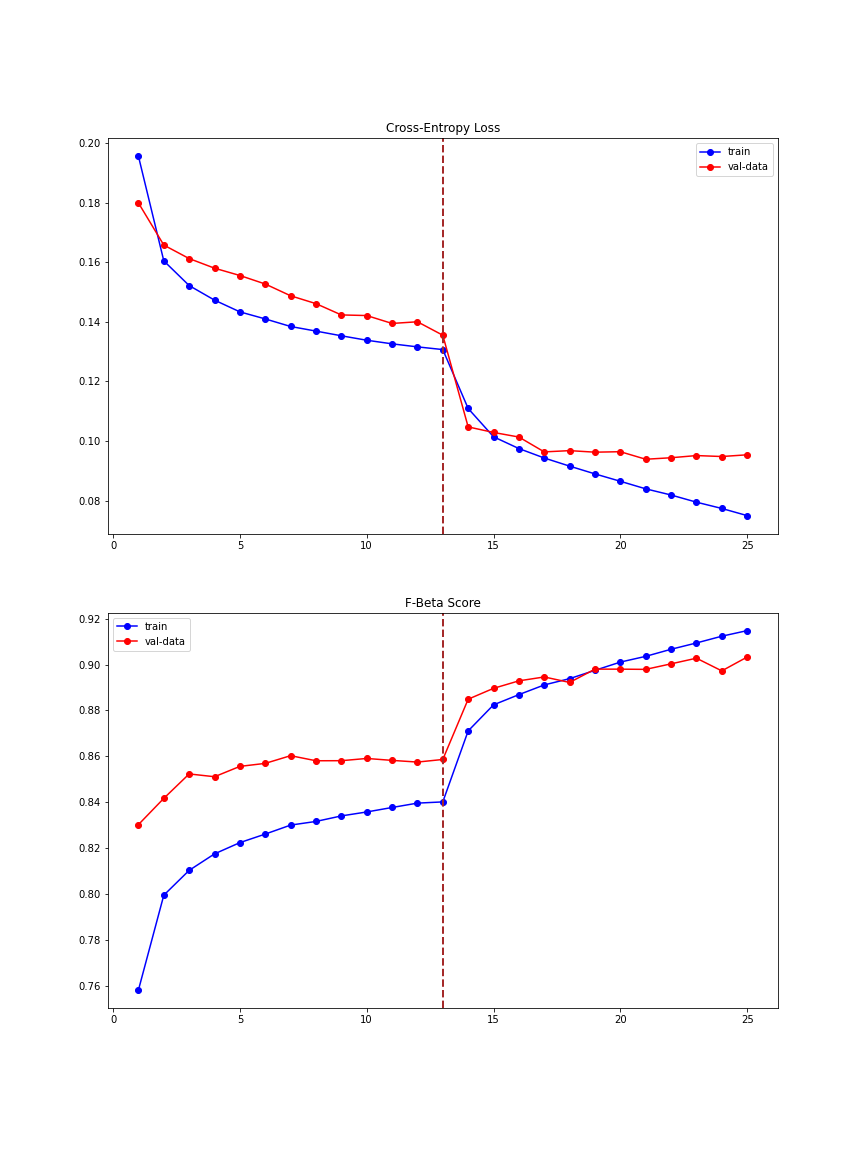}
			\caption{VGG19}
			\label{fig:loss_score_vgg19}
		\end{subfigure}
	\end{figure}
	\newpage
	{\noindent\textbf{Class-wise and Overall Performance of all the Above Models}}
	\begin{table}[H]
		\centering
		\begin{minipage}[c]{0.49\textwidth}
			\scalebox{0.55}{\begin{tabular}{lrrrrr}
				\toprule
				Class &  Precision &    Recall &  Accuracy &  F1 Score &  F2 Score \\
				\midrule
				clear &   0.960836 &  0.964115 &  0.947194 &  0.962473 &  0.963457 \\
				cloudy &   0.803984 &  0.850575 &  0.981591 &  0.826623 &  0.840829 \\
				haze &   0.750219 &  0.635015 &  0.961600 &  0.687826 &  0.655135 \\
				partly\_cloudy &   0.910699 &  0.893388 &  0.965159 &  0.901961 &  0.896798 \\
				agriculture &   0.817633 &  0.858605 &  0.898713 &  0.837618 &  0.850085 \\
				artisinal\_mine &   0.750000 &  0.772189 &  0.995948 &  0.760933 &  0.767647 \\
				bare\_ground &   0.527675 &  0.166473 &  0.979145 &  0.253097 &  0.192878 \\
				blooming &   0.447368 &  0.051360 &  0.991722 &  0.092141 &  0.062408 \\
				blow\_down &   0.000000 &  0.000000 &  0.997628 &  0.000000 &  0.000000 \\
				conventional\_mine &   0.000000 &  0.000000 &  0.997554 &  0.000000 &  0.000000 \\
				cultivation &   0.638688 &  0.474307 &  0.912180 &  0.544359 &  0.500047 \\
				habitation &   0.765143 &  0.590325 &  0.946576 &  0.666461 &  0.618592 \\
				primary &   0.982086 &  0.982269 &  0.966962 &  0.982177 &  0.982232 \\
				road &   0.834239 &  0.817596 &  0.931231 &  0.825834 &  0.820871 \\
				selective\_logging &   0.525547 &  0.212389 &  0.991796 &  0.302521 &  0.241125 \\
				slash\_burn &   0.000000 &  0.000000 &  0.994910 &  0.000000 &  0.000000 \\
				water &   0.848363 &  0.713688 &  0.924238 &  0.775220 &  0.737090 \\
				Total &   0.918523 &  0.896616 &  0.641775 &  0.900668 &  \cellcolor[RGB]{255, 255, 0} 0.896753 \\
				\bottomrule
			\end{tabular}}
			\caption{Inception-V3}
			\label{table:inception-v3}
		\end{minipage}
		\hfill
		\begin{minipage}[c]{0.49\textwidth}
			\scalebox{0.55}{\begin{tabular}{lrrrrr}
				\toprule
				Class &  Precision &    Recall &  Accuracy &  F1 Score &  F2 Score \\
				\midrule
				clear &   0.962373 &  0.965522 &  0.949270 &  0.963945 &  0.964891 \\
				cloudy &   0.830460 &  0.830460 &  0.982505 &  0.830460 &  0.830460 \\
				haze &   0.759235 &  0.647997 &  0.962860 &  0.699219 &  0.667558 \\
				partly\_cloudy &   0.907083 &  0.904959 &  0.966320 &  0.906019 &  0.905383 \\
				agriculture &   0.834559 &  0.838626 &  0.900319 &  0.836587 &  0.837809 \\
				artisinal\_mine &   0.774481 &  0.772189 &  0.996219 &  0.773333 &  0.772647 \\
				bare\_ground &   0.553191 &  0.151339 &  0.979392 &  0.237660 &  0.177063 \\
				blooming &   0.375000 &  0.027190 &  0.991673 &  0.050704 &  0.033383 \\
				blow\_down &   0.000000 &  0.000000 &  0.997628 &  0.000000 &  0.000000 \\
				conventional\_mine &   0.618182 &  0.343434 &  0.997875 &  0.441558 &  0.376940 \\
				cultivation &   0.692202 &  0.450179 &  0.917048 &  0.545553 &  0.484026 \\
				habitation &   0.776918 &  0.603444 &  0.948479 &  0.679280 &  0.631651 \\
				primary &   0.980308 &  0.986242 &  0.968890 &  0.983266 &  0.985049 \\
				road &   0.842540 &  0.822181 &  0.933900 &  0.832236 &  0.826174 \\
				selective\_logging &   0.496183 &  0.191740 &  0.991599 &  0.276596 &  0.218561 \\
				slash\_burn &   0.000000 &  0.000000 &  0.994910 &  0.000000 &  0.000000 \\
				water &   0.832806 &  0.746355 &  0.926141 &  0.787214 &  0.762179 \\
				Total &   0.923614 &  0.898184 &  0.648299 &  0.903931 &  \cellcolor[RGB]{255, 255, 0} 0.899026 \\
				\bottomrule
			\end{tabular}}
			\caption{Inception-ResNet-V2}
			\label{table:loss_score_inc_res_v2}
		\end{minipage}
		\medskip
		\begin{minipage}[c]{0.49\textwidth}
			\scalebox{0.55}{\begin{tabular}{lrrrrr}
				\toprule
				Class &  Precision &    Recall &  Accuracy &  F1 Score &  F2 Score \\
				\midrule
				clear &   0.960271 &  0.966859 &  0.948627 &  0.963554 &  0.965534 \\
				cloudy &   0.811667 &  0.846264 &  0.981937 &  0.828605 &  0.839111 \\
				haze &   0.766085 &  0.618323 &  0.961996 &  0.684319 &  0.643133 \\
				partly\_cloudy &   0.914956 &  0.896556 &  0.966493 &  0.905663 &  0.900177 \\
				agriculture &   0.825409 &  0.855843 &  0.901060 &  0.840351 &  0.849578 \\
				artisinal\_mine &   0.752778 &  0.801775 &  0.996145 &  0.776504 &  0.791472 \\
				bare\_ground &   0.556485 &  0.154831 &  0.979441 &  0.242259 &  0.180952 \\
				blooming &   0.428571 &  0.009063 &  0.991796 &  0.017751 &  0.011270 \\
				blow\_down &   0.000000 &  0.000000 &  0.997628 &  0.000000 &  0.000000 \\
				conventional\_mine &   0.736842 &  0.282828 &  0.997998 &  0.408759 &  0.322581 \\
				cultivation &   0.685229 &  0.450849 &  0.916356 &  0.543862 &  0.483956 \\
				habitation &   0.769178 &  0.616562 &  0.948603 &  0.684466 &  0.642040 \\
				primary &   0.981795 &  0.983549 &  0.967852 &  0.982671 &  0.983198 \\
				road &   0.844492 &  0.823668 &  0.934592 &  0.833950 &  0.827750 \\
				selective\_logging &   0.506579 &  0.227139 &  0.991673 &  0.313646 &  0.255305 \\
				slash\_burn &   0.000000 &  0.000000 &  0.994910 &  0.000000 &  0.000000 \\
				water &   0.852178 &  0.731506 &  0.927624 &  0.787245 &  0.752827 \\
				Total &   0.922641 &  0.897986 &  0.648743 &  0.903472 &  \cellcolor[RGB]{255, 255, 0} 0.898745 \\
				\bottomrule
			\end{tabular}}
			\caption{Xception}
			\label{table:loss_score_xcpetion}
		\end{minipage}
		\hfill
		\begin{minipage}[c]{0.49\textwidth}
			\scalebox{0.55}{\begin{tabular}{lrrrrr}
				\toprule
				Class &  Precision &    Recall &  Accuracy &  F1 Score &  F2 Score \\
				\midrule
				clear &   0.964731 &  0.961371 &  0.948183 &  0.963048 &  0.962041 \\
				cloudy &   0.865426 &  0.779215 &  0.982357 &  0.820060 &  0.795055 \\
				haze &   0.796509 &  0.592359 &  0.962762 &  0.679430 &  0.624365 \\
				partly\_cloudy &   0.875080 &  0.940771 &  0.965282 &  0.906737 &  0.926856 \\
				agriculture &   0.821027 &  0.854300 &  0.899009 &  0.837333 &  0.847432 \\
				artisinal\_mine &   0.745257 &  0.813609 &  0.996120 &  0.777935 &  0.798954 \\
				bare\_ground &   0.600000 &  0.167637 &  0.979960 &  0.262056 &  0.195865 \\
				blooming &   0.448980 &  0.066465 &  0.991697 &  0.115789 &  0.080117 \\
				blow\_down &   0.533333 &  0.083333 &  0.997653 &  0.144144 &  0.100251 \\
				conventional\_mine &   0.655172 &  0.575758 &  0.998221 &  0.612903 &  0.590062 \\
				cultivation &   0.641011 &  0.498660 &  0.913662 &  0.560945 &  0.521837 \\
				habitation &   0.762712 &  0.528833 &  0.942524 &  0.624596 &  0.563384 \\
				primary &   0.975533 &  0.990801 &  0.968445 &  0.983108 &  0.987709 \\
				road &   0.828992 &  0.824164 &  0.931034 &  0.826571 &  0.825125 \\
				selective\_logging &   0.507692 &  0.194690 &  0.991673 &  0.281450 &  0.222073 \\
				slash\_burn &   0.375000 &  0.014563 &  0.994860 &  0.028037 &  0.018029 \\
				water &   0.852382 &  0.738931 &  0.928785 &  0.791612 &  0.759139 \\
				Total &   0.919862 &  0.899134 &  0.641429 &  0.902398 &  \cellcolor[RGB]{255, 255, 0} 0.898872 \\
				\bottomrule
			\end{tabular}}
			\caption{DenseNet-121}
			\label{table:loss_score_dense121}
		\end{minipage}
		\medskip
		\begin{minipage}[c]{0.49\textwidth}
			\scalebox{0.55}{\begin{tabular}{lrrrrr}
					\toprule
					Class &  Precision &    Recall &  Accuracy &  F1 Score &  F2 Score \\
					\midrule
					clear &   0.965403 &  0.964044 &  0.950481 &  0.964723 &  0.964316 \\
					cloudy &   0.822980 &  0.843870 &  0.982579 &  0.833294 &  0.839607 \\
					haze &   0.796768 &  0.603487 &  0.963330 &  0.686788 &  0.634259 \\
					partly\_cloudy &   0.901990 &  0.917769 &  0.967358 &  0.909811 &  0.914569 \\
					agriculture &   0.831258 &  0.845773 &  0.900838 &  0.838453 &  0.842829 \\
					artisinal\_mine &   0.698297 &  0.849112 &  0.995676 &  0.766355 &  0.813953 \\
					bare\_ground &   0.528481 &  0.194412 &  0.979219 &  0.284255 &  0.222548 \\
					blooming &   0.366667 &  0.066465 &  0.991426 &  0.112532 &  0.079480 \\
					blow\_down &   0.400000 &  0.083333 &  0.997529 &  0.137931 &  0.099010 \\
					conventional\_mine &   0.606742 &  0.545455 &  0.998023 &  0.574468 &  0.556701 \\
					cultivation &   0.652910 &  0.491287 &  0.914848 &  0.560683 &  0.516877 \\
					habitation &   0.769288 &  0.564089 &  0.945291 &  0.650899 &  0.595877 \\
					primary &   0.980143 &  0.985735 &  0.968272 &  0.982931 &  0.984612 \\
					road &   0.830511 &  0.826394 &  0.931750 &  0.828447 &  0.827214 \\
					selective\_logging &   0.458101 &  0.241888 &  0.991253 &  0.316602 &  0.267101 \\
					slash\_burn &   0.181818 &  0.009709 &  0.994737 &  0.018433 &  0.011976 \\
					water &   0.860295 &  0.747300 &  0.931528 &  0.799827 &  0.767461 \\
					Total &   0.922211 &  0.900318 &  0.646495 &  0.904277 &  \cellcolor[RGB]{255, 255, 0} 0.900371 \\
					\bottomrule
			\end{tabular}}
			\caption{DenseNet-169}
			\label{table:loss_score_dense169}
		\end{minipage}
		\hfill
		\begin{minipage}[c]{0.49\textwidth}
			\scalebox{0.55}{\begin{tabular}{lrrrrr}
					\toprule
					Class &  Precision &    Recall &  Accuracy &  F1 Score &  F2 Score \\
					\midrule
					clear &   0.969307 &  0.958838 &  0.949764 &  0.964044 &  0.960913 \\
					cloudy &   0.842566 &  0.830460 &  0.983246 &  0.836469 &  0.832853 \\
					haze &   0.780984 &  0.612389 &  0.962737 &  0.686486 &  0.640022 \\
					partly\_cloudy &   0.896813 &  0.930165 &  0.968272 &  0.913185 &  0.923298 \\
					agriculture &   0.831265 &  0.854219 &  0.902889 &  0.842586 &  0.849527 \\
					artisinal\_mine &   0.726131 &  0.855030 &  0.996096 &  0.785326 &  0.825714 \\
					bare\_ground &   0.554598 &  0.224680 &  0.979713 &  0.319801 &  0.255021 \\
					blooming &   0.365854 &  0.090634 &  0.991277 &  0.145278 &  0.106686 \\
					blow\_down &   0.600000 &  0.125000 &  0.997727 &  0.206897 &  0.148515 \\
					conventional\_mine &   0.626374 &  0.575758 &  0.998122 &  0.600000 &  0.585216 \\
					cultivation &   0.639029 &  0.482127 &  0.912600 &  0.549599 &  0.507025 \\
					habitation &   0.768512 &  0.584313 &  0.946502 &  0.663872 &  0.613733 \\
					primary &   0.982350 &  0.983895 &  0.968692 &  0.983122 &  0.983586 \\
					road &   0.829756 &  0.828625 &  0.931923 &  0.829190 &  0.828850 \\
					selective\_logging &   0.467742 &  0.256637 &  0.991327 &  0.331429 &  0.282101 \\
					slash\_burn &   0.000000 &  0.000000 &  0.994786 &  0.000000 &  0.000000 \\
					water &   0.867519 &  0.723947 &  0.929230 &  0.789257 &  0.748730 \\
					Total &   0.922594 &  0.898991 &  0.645754 &  0.903701 &  \cellcolor[RGB]{255, 255, 0} 0.899356 \\
					\bottomrule
			\end{tabular}}
			\caption{DenseNet-201}
			\label{table:loss_score_dense201}
		\end{minipage}
	\end{table}
	\begin{table}[H]
		\centering
		\begin{minipage}[c]{0.49\textwidth}
			\scalebox{0.55}{\begin{tabular}{lrrrrr}
				\toprule
				Class &  Precision &    Recall &  Accuracy &  F1 Score &  F2 Score \\
				\midrule
				clear &   0.958495 &  0.966824 &  0.947293 &  0.962641 &  0.965146 \\
				cloudy &   0.841919 &  0.798372 &  0.981863 &  0.819567 &  0.806717 \\
				haze &   0.765843 &  0.632047 &  0.962613 &  0.692542 &  0.654931 \\
				partly\_cloudy &   0.916703 &  0.879201 &  0.963997 &  0.897560 &  0.886454 \\
				agriculture &   0.824296 &  0.837083 &  0.896143 &  0.830640 &  0.834494 \\
				artisinal\_mine &   0.778846 &  0.718935 &  0.995948 &  0.747692 &  0.730168 \\
				bare\_ground &   0.533742 &  0.101281 &  0.979046 &  0.170254 &  0.120867 \\
				blooming &   0.500000 &  0.003021 &  0.991821 &  0.006006 &  0.003771 \\
				blow\_down &   0.000000 &  0.000000 &  0.997628 &  0.000000 &  0.000000 \\
				conventional\_mine &   0.250000 &  0.010101 &  0.997504 &  0.019417 &  0.012500 \\
				cultivation &   0.671530 &  0.421582 &  0.913218 &  0.517980 &  0.455489 \\
				habitation &   0.768332 &  0.521181 &  0.942499 &  0.621071 &  0.557016 \\
				primary &   0.979565 &  0.985442 &  0.967457 &  0.982495 &  0.984261 \\
				road &   0.830567 &  0.781165 &  0.924584 &  0.805109 &  0.790569 \\
				selective\_logging &   0.571429 &  0.023599 &  0.991673 &  0.045326 &  0.029197 \\
				slash\_burn &   0.000000 &  0.000000 &  0.994910 &  0.000000 &  0.000000 \\
				water &   0.835431 &  0.707883 &  0.921001 &  0.766387 &  0.730179 \\
				Total &   0.921310 &  0.887516 &  0.635425 &  0.896764 &  \cellcolor[RGB]{255, 255, 0} 0.889705 \\
				\bottomrule
			\end{tabular}}
			\caption{ResNet-50-V2}
			\label{table:loss_score_resnet50}
		\end{minipage}
		\hfill
		\begin{minipage}[c]{0.49\textwidth}
			\scalebox{0.55}{\begin{tabular}{lrrrrr}
				\toprule
				Class &  Precision &    Recall &  Accuracy &  F1 Score &  F2 Score \\
				\midrule
				clear &   0.960460 &  0.963130 &  0.946255 &  0.961793 &  0.962595 \\
				cloudy &   0.842843 &  0.806513 &  0.982258 &  0.824278 &  0.813527 \\
				haze &   0.751209 &  0.633902 &  0.961625 &  0.687588 &  0.654338 \\
				partly\_cloudy &   0.908694 &  0.889669 &  0.964170 &  0.899081 &  0.893410 \\
				agriculture &   0.825873 &  0.843580 &  0.898293 &  0.834632 &  0.839978 \\
				artisinal\_mine &   0.753709 &  0.751479 &  0.995873 &  0.752593 &  0.751924 \\
				bare\_ground &   0.565854 &  0.135041 &  0.979441 &  0.218045 &  0.159297 \\
				blooming &   0.500000 &  0.009063 &  0.991821 &  0.017804 &  0.011278 \\
				blow\_down &   0.000000 &  0.000000 &  0.997628 &  0.000000 &  0.000000 \\
				conventional\_mine &   0.647059 &  0.222222 &  0.997801 &  0.330827 &  0.255814 \\
				cultivation &   0.651937 &  0.454870 &  0.912847 &  0.535860 &  0.484139 \\
				habitation &   0.744489 &  0.562995 &  0.943018 &  0.641145 &  0.591852 \\
				primary &   0.979895 &  0.986348 &  0.968593 &  0.983111 &  0.985051 \\
				road &   0.834305 &  0.798017 &  0.928118 &  0.815758 &  0.805020 \\
				selective\_logging &   0.481481 &  0.115044 &  0.991549 &  0.185714 &  0.135699 \\
				slash\_burn &   0.000000 &  0.000000 &  0.994910 &  0.000000 &  0.000000 \\
				water &   0.842937 &  0.712878 &  0.923126 &  0.772471 &  0.735577 \\
				Total &   0.921228 &  0.892100 &  0.637574 &  0.899371 &  \cellcolor[RGB]{255, 255, 0} 0.893509 \\
				\bottomrule
			\end{tabular}}
			\caption{ResNet-101-V2}
			\label{table:loss_score_resnet101}
		\end{minipage}
		\medskip
		\begin{minipage}[c]{0.49\textwidth}
			\scalebox{0.55}{\begin{tabular}{lrrrrr}
				\toprule
				Class &  Precision &    Recall &  Accuracy &  F1 Score &  F2 Score \\
				\midrule
				clear &   0.960045 &  0.963693 &  0.946329 &  0.961865 &  0.962961 \\
				cloudy &   0.833743 &  0.811782 &  0.981937 &  0.822616 &  0.816081 \\
				haze &   0.744196 &  0.618323 &  0.960414 &  0.675446 &  0.639972 \\
				partly\_cloudy &   0.904610 &  0.897383 &  0.964615 &  0.900982 &  0.898819 \\
				agriculture &   0.831808 &  0.834240 &  0.898243 &  0.833022 &  0.833753 \\
				artisinal\_mine &   0.768769 &  0.757396 &  0.996071 &  0.763040 &  0.759644 \\
				bare\_ground &   0.549550 &  0.142026 &  0.979318 &  0.225717 &  0.166758 \\
				blooming &   0.333333 &  0.021148 &  0.991648 &  0.039773 &  0.026022 \\
				blow\_down &   0.000000 &  0.000000 &  0.997628 &  0.000000 &  0.000000 \\
				conventional\_mine &   0.739130 &  0.171717 &  0.997825 &  0.278688 &  0.202864 \\
				cultivation &   0.683523 &  0.455987 &  0.916479 &  0.547038 &  0.488511 \\
				habitation &   0.738501 &  0.557256 &  0.942129 &  0.635202 &  0.586021 \\
				primary &   0.979398 &  0.984855 &  0.966765 &  0.982119 &  0.983759 \\
				road &   0.834194 &  0.800496 &  0.928488 &  0.816998 &  0.807016 \\
				selective\_logging &   0.580247 &  0.138643 &  0.991944 &  0.223809 &  0.163535 \\
				slash\_burn &   0.000000 &  0.000000 &  0.994910 &  0.000000 &  0.000000 \\
				water &   0.837923 &  0.714633 &  0.922459 &  0.771383 &  0.736300 \\
				Total &   0.920990 &  0.891597 &  0.637822 &  0.898950 &  \cellcolor[RGB]{255, 255, 0} 0.893035 \\
				\bottomrule
			\end{tabular}}
			\caption{ResNet-152-V2}
			\label{table:loss_score_resnet152}
		\end{minipage}
		\hfill
		\begin{minipage}[c]{0.49\textwidth}
			\scalebox{0.55}{\begin{tabular}{lrrrrr}
				\toprule
				Class &  Precision &    Recall &  Accuracy &  F1 Score &  F2 Score \\
				\midrule
				clear &   0.962832 &  0.966965 &  0.950579 &  0.964894 &  0.966135 \\
				cloudy &   0.849628 &  0.819923 &  0.983222 &  0.834511 &  0.825697 \\
				haze &   0.763594 &  0.614614 &  0.961650 &  0.681052 &  0.639571 \\
				partly\_cloudy &   0.906339 &  0.915702 &  0.967901 &  0.910997 &  0.913814 \\
				agriculture &   0.817520 &  0.867051 &  0.900665 &  0.841558 &  0.856671 \\
				artisinal\_mine &   0.800000 &  0.804734 &  0.996689 &  0.802360 &  0.803782 \\
				bare\_ground &   0.632479 &  0.172293 &  0.980306 &  0.270814 &  0.201635 \\
				blooming &   0.500000 &  0.027190 &  0.991821 &  0.051576 &  0.033532 \\
				blow\_down &   0.000000 &  0.000000 &  0.997628 &  0.000000 &  0.000000 \\
				conventional\_mine &   0.676471 &  0.464646 &  0.998147 &  0.550898 &  0.495690 \\
				cultivation &   0.669299 &  0.454424 &  0.914824 &  0.541317 &  0.485604 \\
				habitation &   0.782172 &  0.637879 &  0.951197 &  0.702695 &  0.662316 \\
				primary &   0.979044 &  0.987815 &  0.969112 &  0.983410 &  0.986048 \\
				road &   0.828069 &  0.828377 &  0.931478 &  0.828223 &  0.828315 \\
				selective\_logging &   0.458647 &  0.179941 &  0.991351 &  0.258475 &  0.204835 \\
				slash\_burn &   0.000000 &  0.000000 &  0.994910 &  0.000000 &  0.000000 \\
				water &   0.843976 &  0.756479 &  0.929823 &  0.797836 &  0.772497 \\
				Total &   0.921666 &  0.902721 &  0.649930 &  0.905495 &  \cellcolor[RGB]{255, 255, 0} 0.902357 \\
				\bottomrule
			\end{tabular}}
			\caption{VGG16}
			\label{table:loss_score_vgg16}
		\end{minipage}
		\medskip
		\begin{minipage}[c]{0.49\textwidth}
			\scalebox{0.55}{\begin{tabular}{lrrrrr}
				\toprule
				Class &  Precision &    Recall &  Accuracy &  F1 Score &  F2 Score \\
				\midrule
				clear &   0.963395 &  0.963904 &  0.948924 &  0.963649 &  0.963802 \\
				cloudy &   0.850421 &  0.822318 &  0.983370 &  0.836133 &  0.827789 \\
				haze &   0.755585 &  0.627226 &  0.961650 &  0.685448 &  0.649286 \\
				partly\_cloudy &   0.904458 &  0.916667 &  0.967679 &  0.910521 &  0.914199 \\
				agriculture &   0.812391 &  0.870056 &  0.899330 &  0.840235 &  0.857877 \\
				artisinal\_mine &   0.773743 &  0.819527 &  0.996491 &  0.795977 &  0.809941 \\
				bare\_ground &   0.587838 &  0.202561 &  0.980059 &  0.301299 &  0.233119 \\
				blooming &   0.000000 &  0.000000 &  0.991821 &  0.000000 &  0.000000 \\
				blow\_down &   0.000000 &  0.000000 &  0.997628 &  0.000000 &  0.000000 \\
				conventional\_mine &   0.767442 &  0.333333 &  0.998122 &  0.464789 &  0.375854 \\
				cultivation &   0.670736 &  0.454200 &  0.914972 &  0.541628 &  0.485550 \\
				habitation &   0.796134 &  0.596611 &  0.949715 &  0.682081 &  0.628093 \\
				primary &   0.979958 &  0.986908 &  0.969162 &  0.983421 &  0.985510 \\
				road &   0.835270 &  0.819331 &  0.931750 &  0.827224 &  0.822470 \\
				selective\_logging &   0.514286 &  0.106195 &  0.991673 &  0.176039 &  0.126227 \\
				slash\_burn &   0.000000 &  0.000000 &  0.994910 &  0.000000 &  0.000000 \\
				water &   0.842760 &  0.750270 &  0.928661 &  0.793830 &  0.767107 \\
				Total &   0.921178 &  0.900734 &  0.647681 &  0.904192 &  \cellcolor[RGB]{255, 255, 0} 0.900631 \\
				\bottomrule
			\end{tabular}}
			\caption{VGG19}
			\label{table:loss_score_vgg19}
		\end{minipage}
	\end{table}

	{\noindent\textbf{Can we improve further?}}\\
	
	{\noindent Yes,} we can because we still have some trump cards that we haven't played yet. \textit{Ensembling} and \textit{Class-wise thresholding} are still here to push our score a bit farther. Let's see how our result responds to both these techniques.
	
	\begin{enumerate}[leftmargin=*]
		\item \emph{Weighted Ensemble Technique:} With $ k = 11 $ and 
		\begin{table}[H]
			\centering
			\scalebox{0.9}{
				\begin{tabularx}{0.7\textwidth}{
						| >{\raggedright\arraybackslash}X
						| >{\centering\arraybackslash}X|}
					\hline
					Models Name & Weight \\
					\hline
					VGG16 & 3\\
					VGG19 & 3\\
					ResNet50-V2 & 2\\
					ResNet101-V2 & 2\\
					ResNet152-V2 & 2\\
					DenseNet-169 & 3\\
					DenseNet-121 & 2\\
					DenseNet-201 & 3\\
					Xception & 2\\
					Inception-V3 & 1\\
					Inception-ResNet-V2 & 1\\
					\hline
				\end{tabularx}}
			\caption{Integral Weights for Different ImageNet Models}
			\label{table:weights}
		\end{table}
		
		\begin{table}[H]
			\centering
			\caption{Class-wise precision, recall, accuracy, $ F_1 $, and $ F_2 $ scores before thresholding for WSE}
			\scalebox{0.9}{
				\begin{tabular}{lrrrrr}
					\toprule
					Class &  Precision &    Recall &  Accuracy &  F1 Score &  F2 Score \\
					\midrule
					clear &   0.966252 &  0.970025 &  0.955151 &  0.968135 &  0.969268 \\
					cloudy &   0.860570 &  0.824713 &  0.984062 &  0.842260 &  0.831643 \\
					haze &   0.804959 &  0.626113 &  0.964986 &  0.704360 &  0.655229 \\
					partly\_cloudy &   0.919496 &  0.925069 &  0.972028 &  0.922274 &  0.923949 \\
					agriculture &   0.836436 &  0.864696 &  0.907386 &  0.850331 &  0.858892 \\
					artisinal\_mine &   0.814371 &  0.804734 &  0.996837 &  0.809524 &  0.806643 \\
					bare\_ground &   0.671569 &  0.159488 &  0.980504 &  0.257761 &  0.188187 \\
					blooming &   0.500000 &  0.003021 &  0.991821 &  0.006006 &  0.003771 \\
					blow\_down &   0.000000 &  0.000000 &  0.997628 &  0.000000 &  0.000000 \\
					conventional\_mine &   0.823529 &  0.424242 &  0.998369 &  0.560000 &  0.469799 \\
					cultivation &   0.701635 &  0.460232 &  0.918654 &  0.555855 &  0.494242 \\
					habitation &   0.820669 &  0.590325 &  0.951296 &  0.686695 &  0.625434 \\
					primary &   0.979813 &  0.988748 &  0.970694 &  0.984260 &  0.986948 \\
					road &   0.858333 &  0.829616 &  0.938719 &  0.843730 &  0.835205 \\
					selective\_logging &   0.597561 &  0.144543 &  0.992019 &  0.232779 &  0.170376 \\
					slash\_burn &   0.000000 &  0.000000 &  0.994910 &  0.000000 &  0.000000 \\
					water &   0.875552 &  0.749325 &  0.934617 &  0.807536 &  0.771572 \\
					Total &   0.931345 &  0.903205 &  0.664286 &  0.910640 &  \cellcolor[RGB]{255, 255, 0} 0.904825 \\
					\bottomrule
				\end{tabular}}
			\label{table:wse_table}
		\end{table}
		
		Seeing this outcome, we're more excited to implement some other techniques, which in some way assures us of better results. Let's try \textit{Integrated Stacking Model} and see what other surprises we have in there.
		\item \textit{Integrated Stacking Model:} We stacked probabilistic predictions of all the eleven models whose volume became $ 40,469 \times (17\times11) = 40,469 \times 187 $ where each instance belongs to 187 dimensional feature space ($ \mathbb{R}^{187} $). We chose a simple ANN whose number of hidden layers as well as its architecture (like no. of neurons and dropout) would be decided through GridSearchCV.
		
		\textbf{Remarks:} \textit{Grid Search} is an efficient method for tuning the parameters in supervised learning and improving a model's generalization performance. It helps to loop through predefined hyperparameters provided by the user and fit the model on a given training set. So, in the end, we can select the most desirable parameters from the listed hyperparameters. With Grid Search, we try all possible combinations of the parameters of interest and find the fittest ones. Note that there is no way to know the best values for hyperparameters in advance, so ideally, we need to try all possible combinations to know the optimal values. Doing this manually could take a significant amount of time and resources, and thus we use GridSearchCV to automate the tuning of hyperparameters. The GridSearchCV class is provided by \textit{Scikit-learn} \cite{sklearn} package.
		
		
		\begin{table}[H]
			\centering
			\caption{Integrated Stacking Ensemble scores (precision, recall, accuracy, and F$ _{1} $ and F$ _{2} $) before Class-wise Thresholding -- Averaged across all 5 Folds from Stratified 5-fold CV}
			\scalebox{0.92}{
				\begin{tabular}{lrrrrr}
					\toprule
					Class & Precision &    Recall &  Accuracy &  F1 Score &  F2 Score \\
					\midrule
					clear & 0.965877 &  0.970623 &  0.955274 &  0.968238 &  0.969666 \\
					cloudy & 0.846843 &  0.834287 &  0.983642 &  0.840320 &  0.836642 \\
					haze & 0.783312 &  0.650969 &  0.964714 &  0.710745 &  0.673576 \\
					partly\_cloudy & 0.924712 &  0.917355 &  0.971756 &  0.920978 &  0.918792 \\
					agriculture & 0.829312 &  0.874606 &  0.907015 &  0.851274 &  0.865100 \\
					artisinal\_mine & 0.836880 &  0.772169 &  0.996837 &  0.802627 &  0.783943 \\
					bare\_ground & 0.636123 &  0.171134 &  0.980306 &  0.267848 &  0.199947 \\
					blooming & 0.366667 &  0.030258 &  0.991796 &  0.055714 &  0.037023 \\
					blow\_down & 0.333333 &  0.052632 &  0.997727 &  0.090909 &  0.063291 \\
					conventional\_mine & 0.809310 &  0.402632 &  0.998295 &  0.527532 &  0.443835 \\
					cultivation & 0.715945 &  0.464034 &  0.920235 &  0.562544 &  0.498912 \\
					habitation & 0.798276 &  0.639523 &  0.952729 &  0.709766 &  0.665817 \\
					primary & 0.979656 &  0.988563 &  0.970372 &  0.984088 &  0.986768 \\
					road & 0.850229 &  0.840149 &  0.938595 &  0.845129 &  0.842127 \\
					selective\_logging & 0.572346 &  0.194732 &  0.991944 &  0.286605 &  0.223122 \\
					slash\_burn & 0.000000 &  0.000000 &  0.994910 &  0.000000 &  0.000000 \\
					water & 0.858508 &  0.769031 &  0.934518 &  0.811273 &  0.785381 \\
					Total & 0.928411 &  0.906973 &  0.664311 &  0.911286 &  \cellcolor[RGB]{255, 255, 0} 0.907333 \\
					\bottomrule
				\end{tabular}}
			\label{table:ise_wt}
		\end{table}
		
		
		\begin{table}[H]
			\centering
			\caption{Integrated Stacking Ensemble Results after thresholding}			
			\begin{tabular}{lrrrrrr}
				\toprule
				Class & Precision &    Recall &  Accuracy &  F1 Score &  F2 Score &  Threshold \\
				\midrule
				clear & 0.936573 &  0.990079 &  0.955843 &  0.962561 &  0.978879 &   0.157288 \\
				cloudy & 0.723739 &  0.949231 &  0.983518 &  0.820139 &  0.892568 &   0.177699 \\
				haze & 0.552833 &  0.866836 &  0.965035 &  0.674645 &  0.777944 &   0.112663 \\
				partly\_cloudy & 0.838621 &  0.976860 &  0.972152 &  0.902433 &  0.945650 &   0.098355 \\
				agriculture & 0.729898 &  0.955413 &  0.907163 &  0.827495 &  0.899751 &   0.187907 \\
				artisinal\_mine & 0.690210 &  0.925988 &  0.997010 &  0.789633 &  0.865688 &   0.091134 \\
				bare\_ground & 0.255198 &  0.609935 &  0.980454 &  0.357506 &  0.473734 &   0.081878 \\
				blooming & 0.193623 &  0.513433 &  0.991697 &  0.280868 &  0.385380 &   0.070172 \\
				blow\_down & 0.211442 &  0.416316 &  0.997727 &  0.242707 &  0.293688 &   0.042642 \\
				conventional\_mine & 0.611695 &  0.646316 &  0.998394 &  0.613610 &  0.629195 &   0.176618 \\
				cultivation & 0.373189 &  0.868636 &  0.920186 &  0.521717 &  0.685865 &   0.129796 \\
				habitation & 0.550576 &  0.858155 &  0.952606 &  0.670084 &  0.771214 &   0.151551 \\
				primary & 0.964243 &  0.997760 &  0.970595 &  0.980713 &  0.990870 &   0.141257 \\
				road & 0.726990 &  0.925031 &  0.938990 &  0.813846 &  0.877003 &   0.176890 \\
				selective\_logging & 0.275238 &  0.589816 &  0.992043 &  0.365849 &  0.466691 &   0.090467 \\
				slash\_burn & 0.091369 &  0.470848 &  0.994910 &  0.151958 &  0.254132 &   0.050307 \\
				water & 0.683765 &  0.890658 &  0.934197 &  0.773307 &  0.839582 &   0.173569 \\
				\rowcolor{yellow}Total & 0.841126 &  0.965827 &  0.515654 &  0.885617 & 0.926868 &   -- \\
				\bottomrule
			\end{tabular}
			\label{table:ise_t}
		\end{table}
	\end{enumerate}
	Note that the accuracy is very high for the rarer labels because most of the images do not have these labels. However, this increased accuracy does not translate to high precision and recall measures for these classes. Both precision and recall get better for classes with more examples. For most classes, the recall is higher than the precision because the thresholds are chosen to maximize the F$ _{2} $ score, which favors higher recall.
	
	\newpage
	\thispagestyle{empty}
	\section[Chapter 5]{\Large Chapter 5}\label{conclusion}
	{\Huge Improvements \& Conclusion}
	\vspace*{0.8cm}
	\hrule
	\subsection[Possible Improvements]{\Large Possible Improvements}
	
	After extensive testing and trying all possible approaches and configuration of all the models we have deployed, we believe we have reached the limits of every ImageNet architecture, which represents the components of our current best model. However, several other CNN architectures may hold the potential to further improve the performance on this classification task. For example, NasNet Large \cite{nasnet} uses a Recurrent Neural Network (RNN) as a controller to sample the building blocks like identity mapping, average and max pooling, depth-wise separable convolution, and dilated convolution and put them together to create some end-to-end architecture. This new network architecture thus formed is then trained to convergence on a held-out validation set. The resulting accuracies help update the controller according to some policy gradient so that the controller will generate better architectures over time, perhaps by selecting better blocks or making better connections or configurations. It’s reasonably an intuitive approach! In naive terms:
	\begin{enumerate}[leftmargin=*]
		\item Have an algorithm grab different blocks and put them together to create a network.
		\item Train and test out that network.
		\item Based on the results, adjust those blocks you used in the network.
	\end{enumerate}
	
	Another possible improvement can be achieved by increasing (by additional augmentation) the number of less frequently observed training images that contain  rare labels like \enquote{conventional\_mine.} In the present training regime, the model sees very few such images compared to those with frequently occurring labels, so it is possible that it is not able to generalize related patterns to test images. Increasing the frequency of rare labels in the dataset and giving the model more exposure to those labels should improve the situation.\\
	
	One more approach for data imbalance is constructing an additional model that only trains on the rare labels. First, we train our primary model on the whole dataset where we represent the cluster of rare labels in the alias of "rare\_label," which transforms our problem into an 11-class multi-label issue. Next, we train our secondary model on the dataset corresponding to rare labels and use its output only if the score for the "rare\_label" in the primary network is above some threshold.\\
	
	Another way to exploit the label ordering in the dataset (i.e., basically to capture the correlation among labels), as well as the fixed \enquote*{vocabulary} (only 17 possible tags), is through making use of \textit{LSTM} architecture as our label captioner that generates a predicted image caption one word at a time using the inputs from ImageNet models.\\
	
	Further, we can transform our multi-label classification problem into a multi-class problem with one multi-class classifier trained on all unique label combinations found in the training data. This approach is regarded as \emph{Label Powerset} (LP) transformation. Here each combination is mapped to a unique combination id number and performs multi-class classification using a multi-class classifier and combinations ids as classes. Along with the advantage that this algorithm takes into label correlations, there are some downsides of this algorithm like the complexity is exponential with the number of labels, imbalance in the dataset as many labels are associated with a few examples, and the inability to predict a labelset which does not appear in the training dataset. RAndom k-labELsets (RAkEL) \cite{rakel} comes to the rescue for all the above problems as it randomly breaks the set of labels into several small-sized labelsets. RAkEL selects n random k-labelsets and learns n LP classiﬁers, each one focusing on its k-labelset. Each model provides binary predictions for each label in its corresponding k-labelset. These outputs are combined for a multi-label prediction following a majority voting process for each label. In the process, RAkEL mitigates all the above issues \cite{multi-label-ensemble}. 
	\begin{enumerate}[leftmargin=*]
		\item First, the LP tasks of each classiﬁer are much simpler since they only consider a small subset of the labels. 
		\item Also, the base classiﬁers include a much more balanced distribution of classes than using LP with the complete set of labels. 
		\item Further, RAkEL allows predicting a labelset that does not appear in the original training set.
	\end{enumerate}
	
\begin{figure}[H]
		\centering
		\copyrightbox[b]{\includegraphics[width=0.83\linewidth]{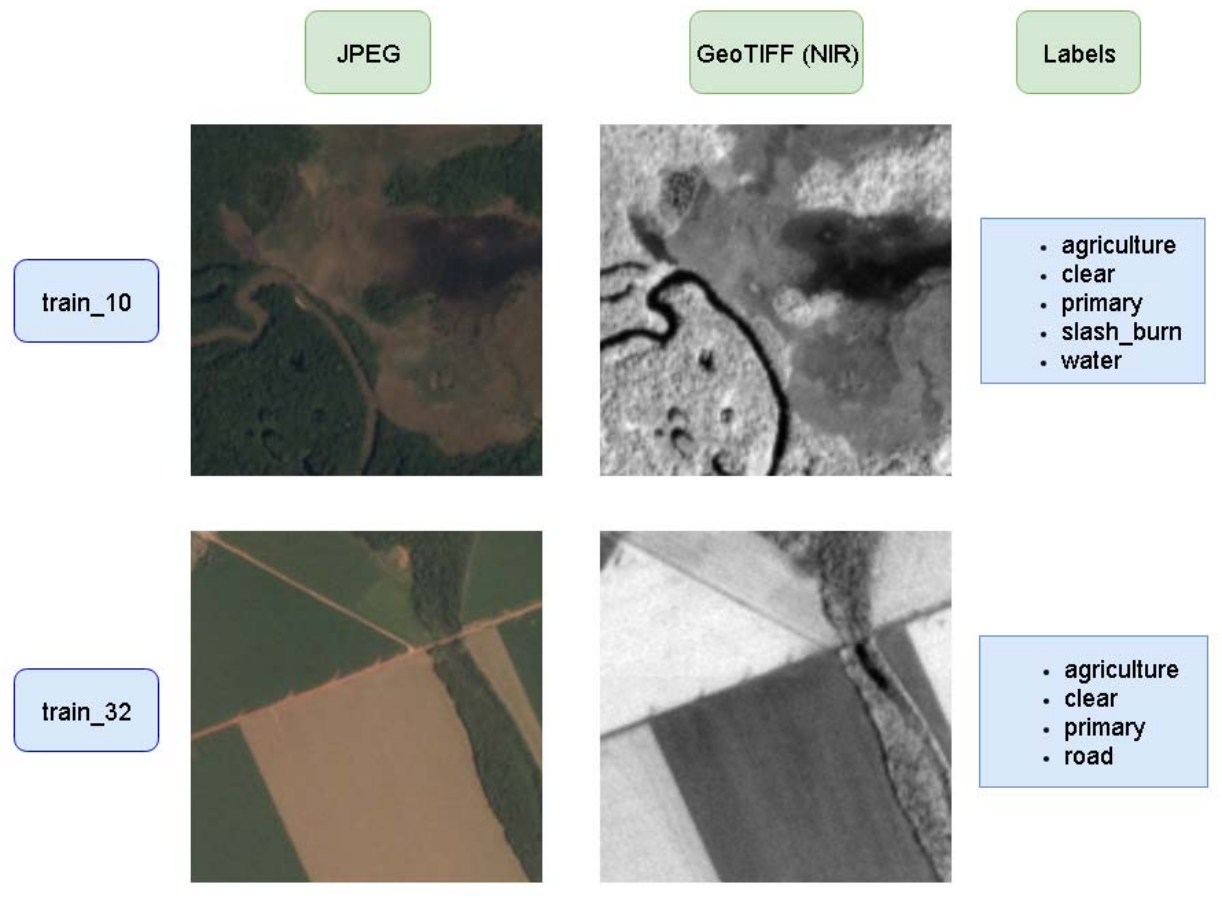}}
		{Source: \cite{comp-4}}
		\caption{JPEG vs. GeoTIFF NIR Band}
		\label{fig:tiff}
	\end{figure}

	Lastly, we aim to re-train our models on 4-band GeoTIFF images rather than 3-band JPEG images. The JPEG images are attractive for training because one can reuse the previously developed architectures for computer vision on natural images, such as the models that came out of the ImageNet Large Scale Visual Recognition Challenge (ILSVRC) \cite{22-4}. However, these images do not contain the near-infrared (NIR) band, so they miss some potentially useful information. Fig. \ref{fig:tiff} shows two JPEG images along with the NIR band (shown grayscale) from the corresponding GeoTIFF images.
	
	One can see that some of the crucial features, such as rivers, are more clearly visible in the NIR band than in the JPEG images. Hence, we believe that developing CNN architectures to process 4-band GeoTIFF images would be very useful for our task and would lead to further performance improvements.
	\subsection[Conclusion]{\Large Conclusion}
	
	This report represents several other pioneers who have consolidated their novel approaches into this task and took part in the classification contest created by \textit{Planet Labs} and  \textit{SCCON}. Our uniqueness lies in the fact that we have integrated all the models into a final ensemble model and tuned the class-wise threshold to have maximum output. We have discussed our shortcomings and several unprecedented ways to improve on them and hence on this project. With the current boom in satellite earth-imaging companies, the obvious challenge lies in the accurate and automated interpretation of the massive datasets of accumulated images. Here, we tried to tackle the challenge of understanding one subset of satellite images -- those capturing images of the Amazon rainforest -- with the particular goal of aiding in characterization and quantification in the area.\\
	
	Using pre-trained deep CNN state-of-the-art models designed for the ImageNet challenge, we were able to create architectures that exploit the structure of our dataset in multiple ways and achieved strong performance. Still moving forward, there are various milestones we wish to pursue. One interesting improvement would be to experiment with more exotic refining layers.\\
	
	Anthropological action may sometimes lead to ruination at a local, regional, and global scale. Conventional mining is one such deed that results in loss of biodiversity, or the contamination of groundwater, soil and surface water, erosion, and sinkholes due to the chemical emitted from the mining process. A mining company usually handles these operations and, while undoubtedly destructive, are generally sanctioned by the local and federal government in which they operate. Artisinal mines are small-scale and, by-in-large, illegal. Since there is no knowledge or oversight about where they are occurring, there is little to no ability for the local government to manage their impact on deforestation in the area. Having a model that can find these mines is a definite win for those concerned about destroying the rainforest habitat. Overall experimenting with and optimizing our suite model frameworks served to be an illuminating and exciting final year MS project, mainly when applied to a topical and impactful real-world challenge.

	\newpage
	\thispagestyle{empty}
	\rhead{References}
	\addcontentsline{toc}{section}{\protect\numberline{}References}
	\section*{References}\label{reference}
	\printbibliography[heading=none]
	
	\newpage
	\thispagestyle{empty}
	\rhead{Appendix}
	\addcontentsline{toc}{section}{\protect\numberline{}Appendix}
	\section*{Appendix}\label{appendix}
	{\noindent \emph{For all the following images we have referred to \cite{kaggle-planet}.}}
	\begin{enumerate}[wide = 0pt]
		\item \textbf{Atmospheric Labels}: Clouds are a significant challenge for passive satellite imaging, and daily cloud cover and rain showers in the Amazon basin can significantly hamper monitoring in the area. For this reason, a cloud cover label is included for each chip. Labels like: clear, partly cloudy, cloudy, and haze closely mirror what one would observe in a local weather forecast. In our case, atmospheric clouds are visible in haze chips, but they are not so opaque to obscure the ground. Clear chips show no evidence of clouds, and partly cloudy scenes can include opaque cloud cover over any portion of the image. Cloudy images have 90\% of the chip obscured by opaque cloud cover.
		\begin{enumerate}[wide = 0pt]
			\item \textbf{Cloudy}
			\begin{figure}[H]
				\centering
				\begin{subfigure}[b]{\textwidth}
					\includegraphics[width=\linewidth]{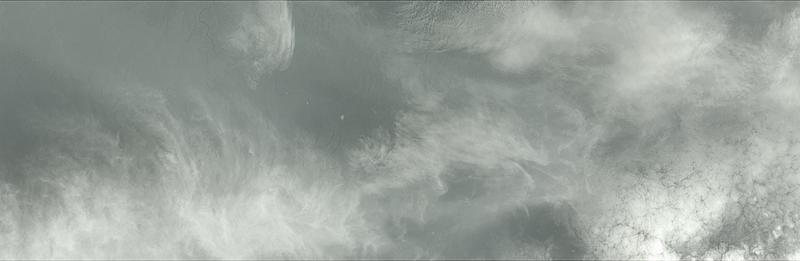}
				\end{subfigure}
				\medskip
				\begin{subfigure}[b]{\textwidth}
					\includegraphics[width=\linewidth]{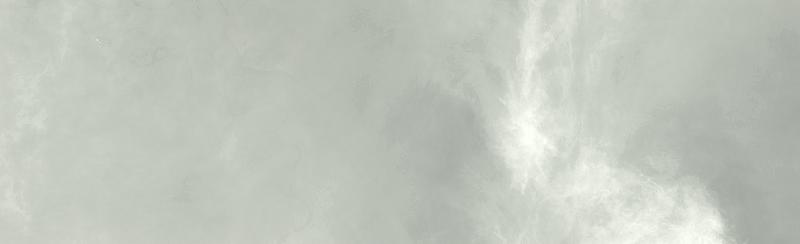}
				\end{subfigure}
				\caption{Cloudy Cover}
				\label{fi:cloud}
			\end{figure}
			\pagebreak
			\item \textbf{Partly Cloudy}
			\begin{figure}[H]
				\centering
				\begin{subfigure}[b]{\textwidth}
					\includegraphics[width=\linewidth]{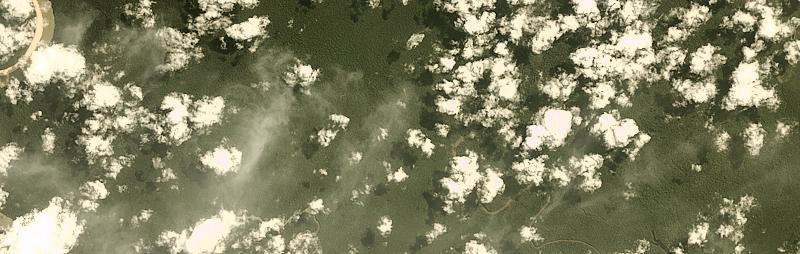}
				\end{subfigure}
				\medskip
				\begin{subfigure}[b]{\textwidth}
					\includegraphics[width=\linewidth]{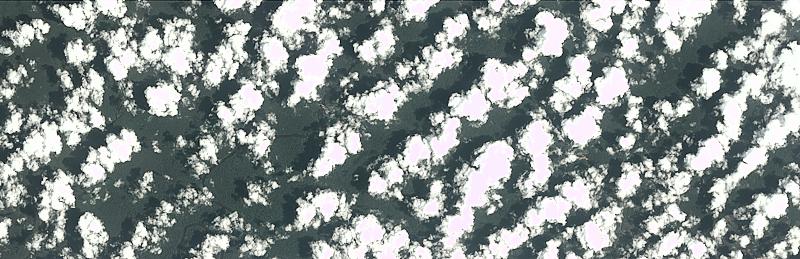}
				\end{subfigure}
				\caption{Partly Cloudy Scenes}
				\label{fig:pc}
			\end{figure}
			\item \textbf{Hazy Scenes}
			\begin{figure}[H]
				\centering
				\begin{subfigure}[b]{\textwidth}
					\centering
					\includegraphics[width=0.87\linewidth]{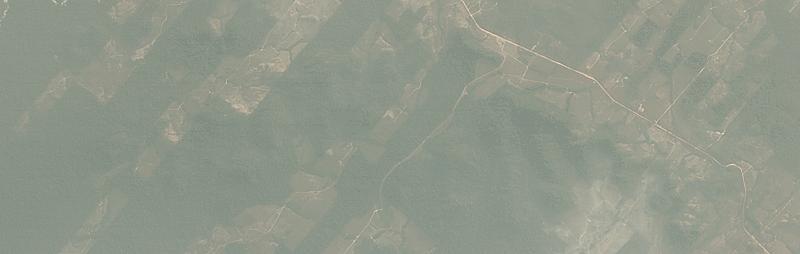}
				\end{subfigure}
				\medskip
				\begin{subfigure}[b]{\textwidth}
					\centering
					\includegraphics[width=0.87\linewidth]{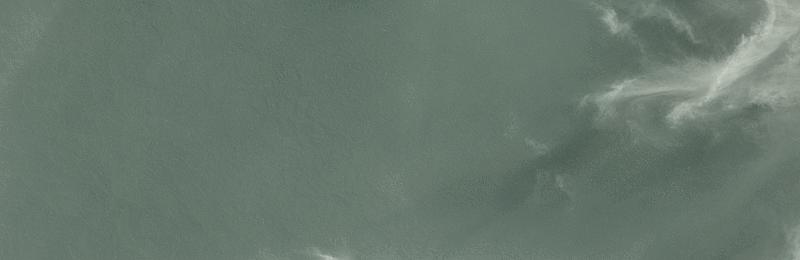}
				\end{subfigure}
				\caption{Hazy Scenes}
				\label{fig:haze}
			\end{figure}
		\end{enumerate}
		\item Common Labels
		\begin{enumerate}[wide = 0pt]
			\item \textbf{Primary Rain Forest}: The substantial majority of the data set is labeled as \enquote{primary,} which is shorthand for the primary rainforest, or what is known colloquially as virgin forest. Generally speaking, areas exhibiting dense tree cover were labeled as \enquote{primary.} The Mongobay article \cite{primary} briefly explains the difference between primary and secondary rainforest, but differentiating between the two is difficult entirely using satellite imagery. This is particularly true in older \enquote{secondary} forests.
			\begin{figure}[H]
				\centering
				\includegraphics[width=\linewidth]{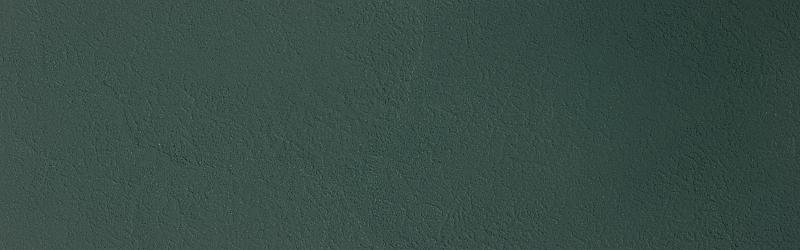}
				\caption{Approximately 25,000 acres of untouched primary rainforest}
				\label{fig:primary}
			\end{figure}
			\item \textbf{Water (Rivers \& Lakes)}:
			\begin{figure}[H]
				\centering
				\begin{subfigure}[b]{\textwidth}
					\includegraphics[width=\linewidth]{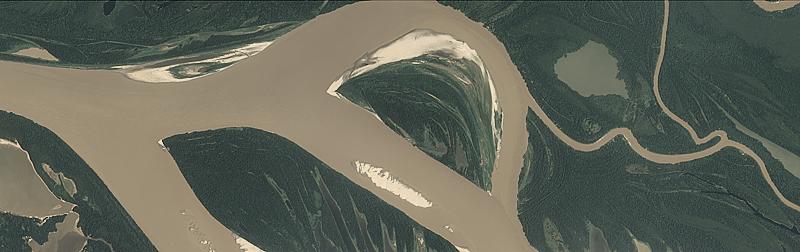}
					\caption{A larger and slower river with significant sand bars and brown silt deposits.}
					\label{fig:river1}
				\end{subfigure}
				\medskip
				\begin{subfigure}[b]{\textwidth}
					\includegraphics[width=\linewidth]{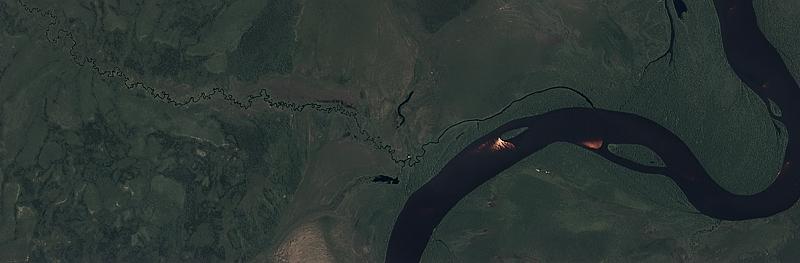}
					\caption{A small tributary joins a larger river system evident from the bright sand bars.}
					\label{fig:river2}
				\end{subfigure}
			\end{figure}
			\item \textbf{Habitation}: The chips that contain human homes or buildings were tagged with habitation class. This covers anything from dense metropolitan centers to rural villages along the banks of rivers. Small, single-dwelling habitations are often tricky to spot but usually seem clusters of a few bright white pixels.
			\begin{figure}[H]
				\centering
				\begin{subfigure}[b]{\textwidth}
					\includegraphics[width=\linewidth]{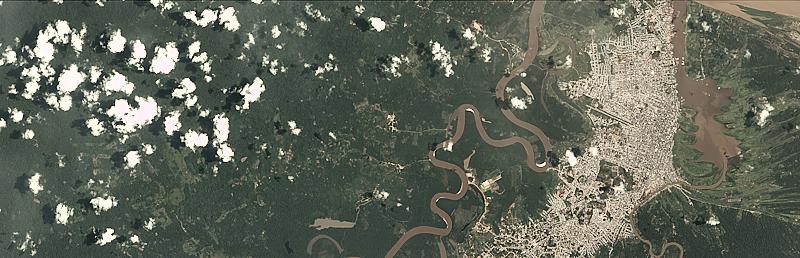}
					\caption{A larger city in the Amazon basin.}
					\label{fig:habitation1}
				\end{subfigure}
				\medskip
				\begin{subfigure}[b]{\textwidth}
					\includegraphics[width=\linewidth]{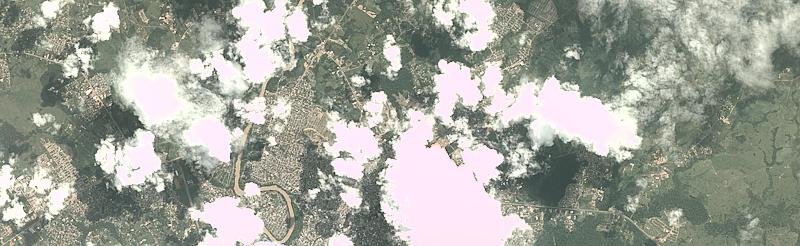}
					\caption{A large city.}
					\label{fig:habitation2}
				\end{subfigure}
			\end{figure}
			\item \textbf{Agriculture}: Commercial agriculture, while an influential industry, is also a major foundation of deforestation in the Amazon. In our dataset, the agriculture tag refers to any land cleared of trees that are being used for agriculture or rangeland.
			\begin{figure}[H]
				\centering
				\subfloat[][An agricultural area that showing the end state of \enquote{fishbone} deforestation.]{\includegraphics[width=\linewidth]{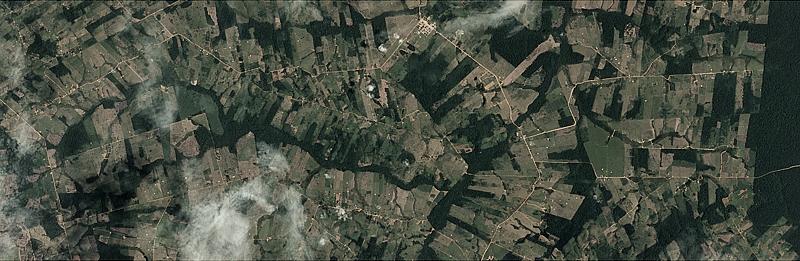}\label{fig:agg1}}
			\end{figure}
			\begin{figure}[H]
				\centering
				\subfloat[][A newer agricultural area showing \enquote{fishbone} deforestation.]{\includegraphics[width=\linewidth]{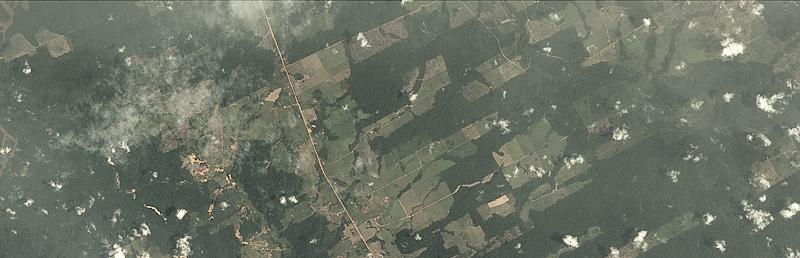}\label{fig:agg2}}
			\end{figure}
			\item \textbf{Road}: Roads are vital for transportation in the Amazon but also laid the basis for deforestation. In particular, \enquote{fishbone} deforestation often follows new road construction, while more minor logging roads run selective logging operations. In our dataset, all types of roads are labeled with a single \enquote{road} label. Sometimes rivers are confused with more minor logging roads, and consequently, there may be some noise in this label. Analysis of the near-infrared band image may prove helpful in disambiguating the two classes.
			\begin{figure}[H]
				\centering
				\subfloat[][classic \enquote{Fishbone} deforestation following a road.]{\includegraphics[width=\linewidth]{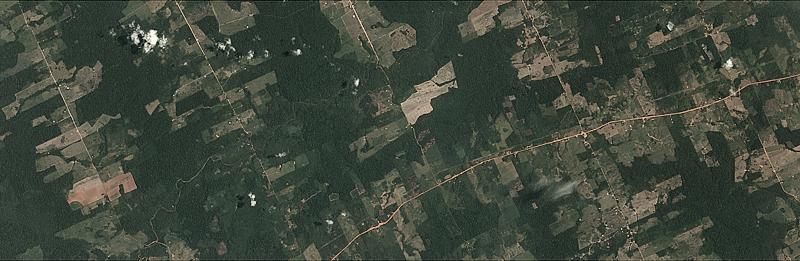}\label{fig:road1}}
				\newline
				\subfloat[][roads snake out of a small town in the Amazon.]{\includegraphics[width=\linewidth]{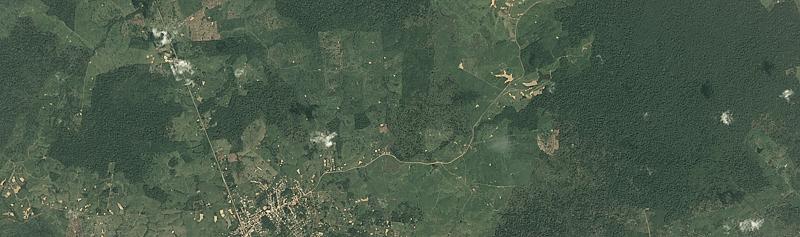}\label{fig:road2}}
			\end{figure}
			\item \textbf{Cultivation}: Shifting cultivation is a subset of agriculture that can be easily seen from space and occurs in rural areas where people maintain farm plots for subsistence. This article \cite{cultivation} by MongaBay gives a comprehensive overview of the practice. This type of agriculture is usually found near smaller villages along main rivers and on the outskirts of agricultural lands. It typically relies on non-mechanized labor and covers relatively small areas.
			\begin{figure}[H]
				\centering
				\begin{subfigure}[b]{\textwidth}
					\includegraphics[width=\linewidth]{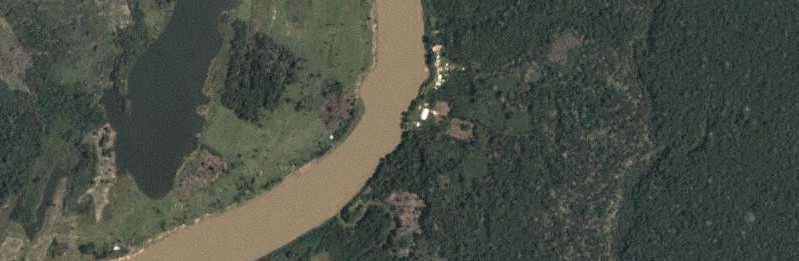}
					\caption{A zoomed-in area showing cultivation (right side of river)}
					\label{fig:cult1}
				\end{subfigure}
				\medskip
				\begin{subfigure}[b]{\textwidth}
					\includegraphics[width=\linewidth]{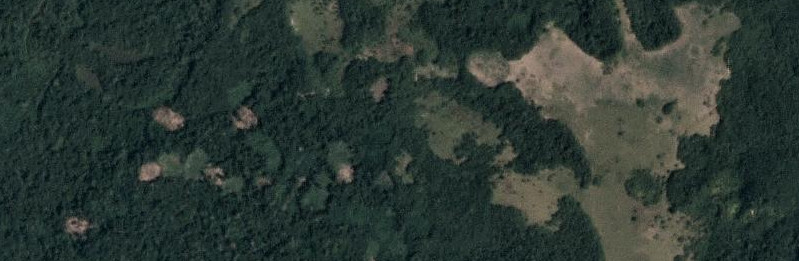}
					\caption{A zoomed-in area showing cultivation and some selective logging. Dark areas indicate recent slash/burn activity}
					\label{fig:cult2}
				\end{subfigure}
			\end{figure}
			\item \textbf{Bare Ground}: Bare ground is a catch-all term for naturally occurring tree-free areas. Some of these areas occur naturally in the Amazon, while others result from the source scenes containing small biome regions similar to the Pantanal \cite{pantanal} or cerrado \cite{cerrado}.
			\begin{figure}[H]
				\centering
				\includegraphics[width=\linewidth]{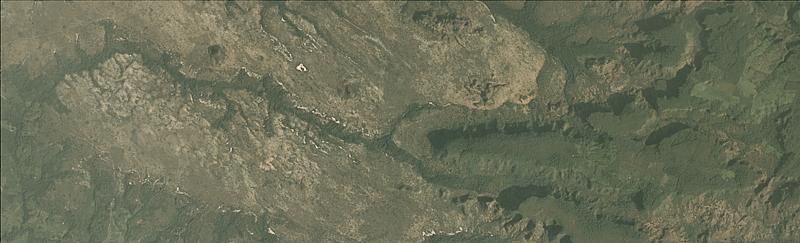}
				\caption{a naturally occuring bare area in the cerrado.}
				\label{fig:bare1}
			\end{figure}
			\begin{figure}[H]
				\centering
				\includegraphics[width=\linewidth]{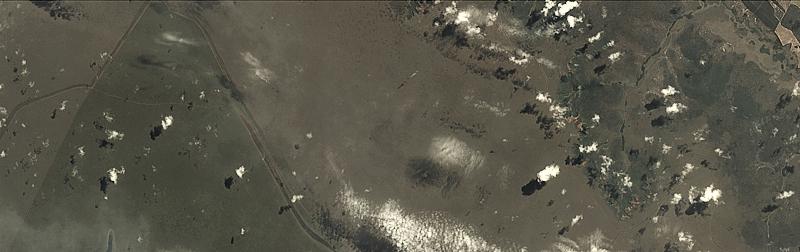}
				\caption{a naturally occuring bare area in the cerrado.}
				\label{fig:bare2}
			\end{figure}
		\end{enumerate}
		\item \textbf{Rare Labels}:
		\begin{enumerate}[wide = 0pt]
			\item \textbf{Slash and Burn}: It is considered a subset of the shifting cultivation used for areas that demonstrate recent burn events. Further, the shifting cultivation patches appear to have dark brown or black spots consistent with recent burning.
			\begin{figure}[H]
				\centering
				\begin{subfigure}[b]{\textwidth}
					\includegraphics[width=\linewidth]{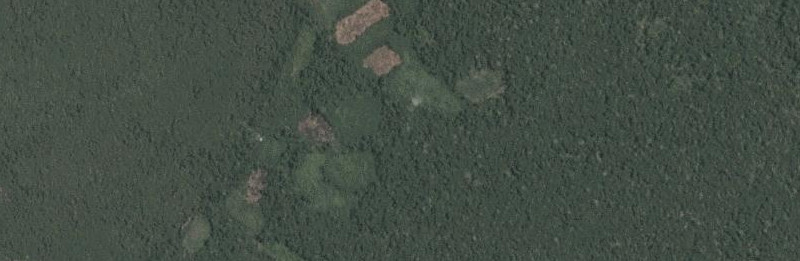}
					\caption{A zoomed-in view of an area with shifting cultivation with evidence of a recent fire.}
					\label{fig:sb1}
				\end{subfigure}
				\medskip
				\begin{subfigure}[b]{\textwidth}
					\includegraphics[width=\linewidth]{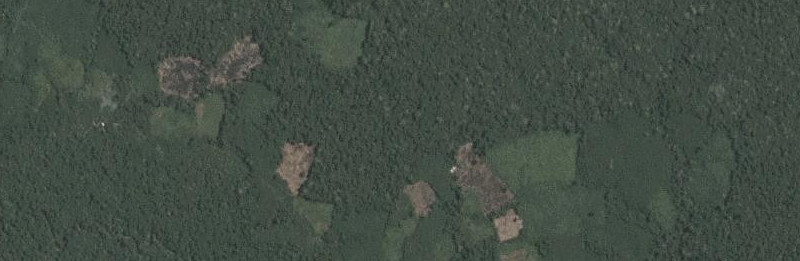}
					\caption{A zoomed-in view of an area with shifting cultivation and evidence of a recent fire.}
					\label{fig:sb2}
				\end{subfigure}
			\end{figure}
			\item \textbf{Selective Logging}: This label encompasses the practice of selectively removing high-value tree species from the rainforest (such as teak and mahogany). This appears as winding dirt roads adjacent to bare brown patches in the otherwise primary rain forest from space. The Mongabay Article \cite{sel-logging-1} covers the details of this process. Global Forest Watch \cite{sel-logging-2} is another excellent resource for learning about deforestation and logging.
			\begin{figure}[H]
				\centering
				\begin{subfigure}[b]{\textwidth}
					\includegraphics[width=\linewidth]{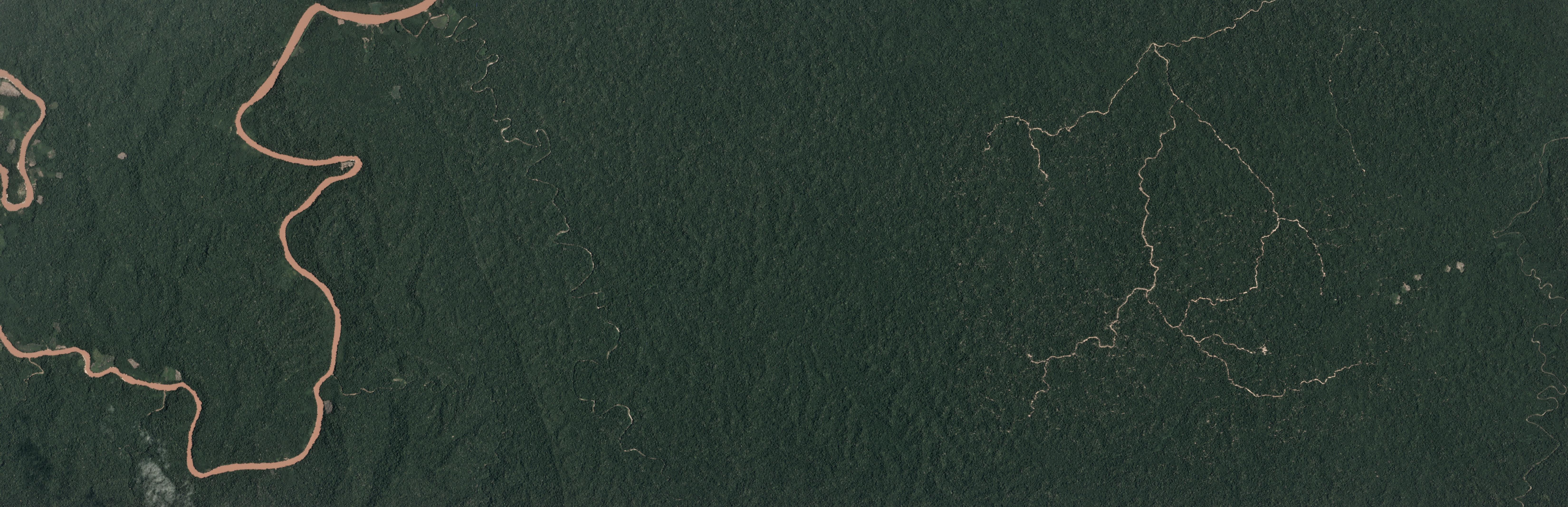}
					\caption{The brown lines on the right of this scene are a logging road. Note the small brown dots in the area around the road.}
					\label{fig:log1}
				\end{subfigure}
				\medskip
				\begin{subfigure}[b]{\textwidth}
					\includegraphics[width=\linewidth]{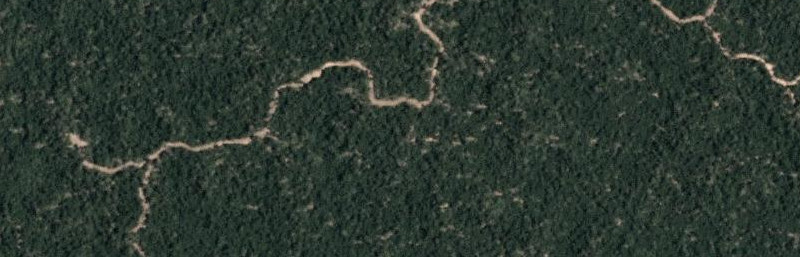}
					\caption{A zoomed image of logging roads and selective logging.}
					\label{fig:log2}
				\end{subfigure}
				\medskip
				\begin{subfigure}[b]{\textwidth}
					\includegraphics[width=\linewidth]{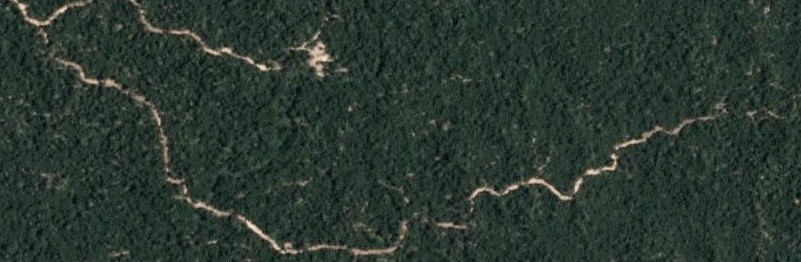}
					\caption{A zoomed image of logging roads and selective logging.}
					\label{fig:log3}
				\end{subfigure}
			\end{figure}
			\item \textbf{Blooming}: It is a natural phenomenon found in the Amazon where particular species of flowering trees bloom, fruit, and flower simultaneously to maximize the possibilities of cross-pollination. These trees are pretty large, and these events can be seen from space.
			\begin{figure}[H]
				\centering
				\includegraphics[width=\linewidth]{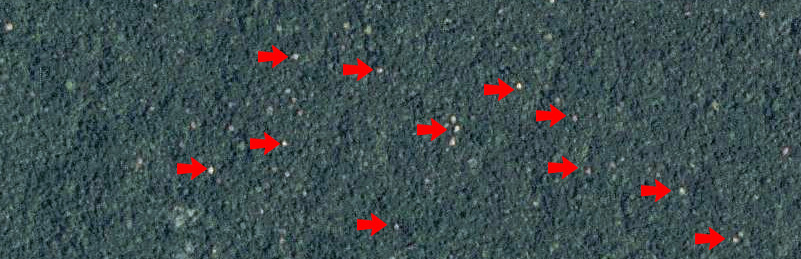}
				\caption{A zoomed and contrast enhanced of a bloom event in the Amazon basin. The red arrows point to a few specific trees. The canopies of these trees can be over 30m across ($ \sim $100ft).}
				\label{fig:bloom}
			\end{figure}
			\item \textbf{Conventional Mining}:
			\begin{figure}[H]
				\centering
				\includegraphics[width=\linewidth]{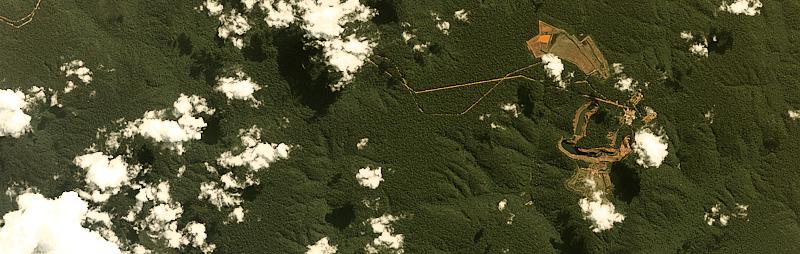}
				\caption{A conventional mine in the Amazon.}
				\label{fig:conv-mine}
			\end{figure}
			\item \textbf{Artisinal Mining}: This term is for small-scale mining operations. Throughout the Amazon, especially at the foothills of the Andes, gold deposits lace the deep, clay soils. Artisanal miners, sometimes working illegally in land designated for conservation, cut through the forest and excavate deep pits near rivers. They pump a mud-water slurry into the river banks, blasting them away so that they can be processed further with mercury - which is used to separate the gold. The denuded moonscape left behind takes centuries to recover.
			\begin{figure}[H]
				\centering
				\includegraphics[width=0.85\linewidth]{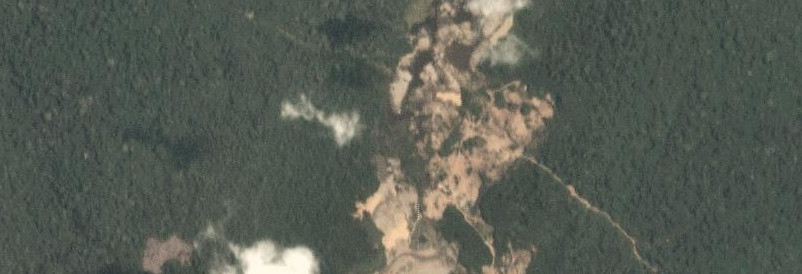}
				\caption{An image of an artisanal mine in Peru.}
				\label{fig:artmine1}
			\end{figure}
			\begin{figure}[H]
				\centering
				\includegraphics[width=\linewidth]{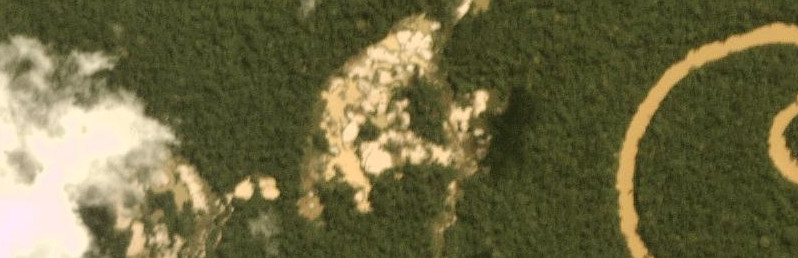}
				\caption{Zoomed part of an artisanal mine in Peru.}
				\label{fig:artmine2}
			\end{figure}
			\item \textbf{Blow Down}: Blow down, also called windthrow \cite{blow-down-1}, is a naturally occurring phenomenon in the Amazon that occurs during microbursts \cite{blow-down-2} where cold, dry air from the Andes settles on top of warm moist air in the rainforest. The colder air punches a hole in the moist warm layer and sinks with tremendous force and high speed (over 100MPH). These high winds overthrow the more giant rainforest trees resulting in large open areas visible from space. These areas soon get covered with plants to take advantage of the sunlight.
			\begin{figure}[H]
				\centering
				\includegraphics[width=\linewidth]{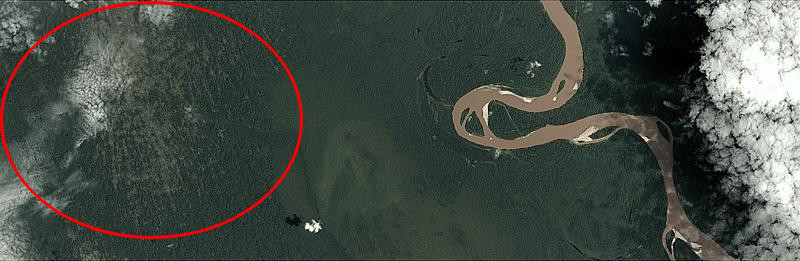}
				\caption{A recent blow down event in the Amazon circled in red. Note the light green of the forest understory and the pattern of tree loss.}
				\label{fig:blowdown}
			\end{figure}
		\end{enumerate}
	\end{enumerate}

\end{document}